\documentclass[12pt]{article}

\usepackage[T1]{fontenc}
\usepackage{jmlr2e}
\usepackage[margin=.75cm,skip=4pt]{caption}
\usepackage{mathtools}
\usepackage{longtable}
\usepackage[utf8]{inputenc}
\usepackage{array}
\usepackage{float}
\usepackage{booktabs}
\usepackage{url}
\usepackage{bm}
\usepackage{multirow}
\usepackage{amsmath}
\usepackage{amsbsy}
\usepackage{amssymb}
\usepackage{graphicx}

\makeatletter

\providecommand{\tabularnewline}{\\}

\usepackage{hyperref}
\usepackage{adjustbox}
\usepackage{enumitem}
\usepackage[section]{placeins}
\usepackage{setspace}
\usepackage{breakcites}
\usepackage{breqn}
\setlist[itemize]{leftmargin=2pc, rightmargin=2pc}
\setlist[enumerate]{leftmargin=2pc, rightmargin=2pc}

\usepackage{url}
\usepackage{color}
\usepackage[ruled,linesnumbered,resetcount]{algorithm2e}

\makeatletter \providecommand{\@LN}[2]{} \makeatother
\usepackage{algcompatible}
\newcounter{algoline}
\newcommand\Numberline{\refstepcounter{algoline}\nlset{\thealgoline}}
\usepackage[english]{babel}

\ShortHeadings{Non-smooth Bayesian Optimization in Tuning Problems}{Authors}

\makeatother
\numberwithin{figure}{section}
\numberwithin{table}{section}
\begin{document}

    \title{Non-smooth Bayesian Optimization in Tuning Problems}
    \author{\name{Hengrui Luo} \email hrluo@lbl.gov \\
    	\addr Lawrence Berkeley National Laboratory\\
    	Computational Research Department\\
    	1 Cyclotron Rd, Berkeley, CA 94720, USA 
    	\AND
    	\name{James W. Demmel} \email demmel@berkeley.edu \\
    	\name{Younghyun Cho}  \email younghyun@berkeley.edu \\
    	\addr University of California, Berkeley\\
    	Department of EECS, \\
    	253 Cory Hall, Berkeley, CA 94720, USA
        \AND
    	\name{Xiaoye S. Li} \email xsli@lbl.gov \\
        \name{Yang Liu} \email liuyangzhuan@lbl.gov \\
    	\addr Lawrence Berkeley National Laboratory\\
    	Computational Research Department\\
    	1 Cyclotron Rd, Berkeley, CA 94720, USA
    	$ $}
	
	\editor{}
	
\maketitle

\date{\today}
\begin{abstract}
Building surrogate models is one common approach when we attempt to
learn unknown black-box functions. Bayesian optimization provides
a framework which allows us to build surrogate models based on sequential
samples drawn from the function and find the optimum. Tuning algorithmic
parameters to optimize the performance of large, complicated `` black-box''
application codes is a specific important application, which aims
at finding the optima of black-box functions. Within the Bayesian optimization
framework, the Gaussian process model produces smooth or continuous
sample paths. However, the black-box function in the tuning problem
is often non-smooth. This difficult tuning problem is worsened
by the fact that we usually have limited sequential samples from the
black-box function. Motivated by these issues encountered in tuning,
we propose a novel additive Gaussian process model called clustered
Gaussian process (cGP), where the additive components are induced
by clustering. In the examples we studied, the performance can be improved by as much as 90\% among repetitive experiments.
By using this surrogate model, we want to capture the
non-smoothness of the black-box function.
In addition to an algorithm for constructing this model, we also apply
the model to several artificial and real applications to evaluate
it.
\end{abstract}

\keywords{Bayesian optimization, surrogate modeling, additive Gaussian
models.}
\maketitle

\section{Introduction}
As an introduction, we briefly review the recent literature on Bayesian
optimization and identify the problem caused by non-smoothness in
tuning scenarios in subsection \ref{subsec:Problem-and-Challenges}.
The follow-up subsections \ref{subsec:Non-smoothness-in-the-domain}
and \ref{subsec:Non-smoothness-in-the-black-box-function} describe
the non-smoothness phenomena and focus on one specific kind of non-smoothness. %

\subsection{\label{subsec:Problem-and-Challenges}Problems and Challenges}

Suppose that we have a scalar function $f$ whose expression we do not know,
but we can draw noisy samples 
in the form of pairs $(\bm{x},y)=(\bm{x},f(\bm{x}))$
from function $f$, i.e., we consider a noisy optimization model. The methodology we present in the current paper focuses on univariate functions and could be extended to multivariate functions, but this remains a possible future work. 
\begin{itemize}
\item When it is easy (or cheap) to evaluate this function at any point
of its domain, the default strategy is to use an exhaustive grid search
of $\bm{x}$ to find its optimum.
\item When it is difficult (or expensive) to evaluate this function at some
points of its domain, one strategy is to use a relatively cheap \emph{surrogate
model} to approximate the black-box function.
\end{itemize}
The general tuning problem is to find the maximum or minimum of such
a function $f$ based on (noisy) samples sequentially drawn from the
function. This function with an unknown expression is usually called
a \emph{black-box function}. An analytic solution for the optimum
of $f$ is usually impossible; even if the analytic expression of
the function is known, the analytic optimum could be not tractable
or expensive to evaluate. In various scenarios, we need to solve this
kind of black-box optimization problem \citep{muller_surrogate_2019,gramacy_surrogates_2020}. 

Following common strategies, a \emph{surrogate model} $g$ is proposed
to model the underlying black-box function $f$. The chosen surrogate
model $g$ needs to be easier to evaluate and approximate the black-box function $f$ reasonably well. This strategy usually works well for smooth functions. 

Bayesian optimization provides a surrogate modeling optimization framework
that allows us to model and optimize these black-box functions using
sequential sampling \citep{booker_rigorous_1998,snoek_practical_2012,shahriari_taking_2016,gramacy_surrogates_2020}.
In practice, we sequentially draw expensive samples from the black-box
function $f$ and update the surrogate model $g$ with the
samples drawn using Bayes' theorem. If the surrogate model $g$ approximates the black-box function $f$ sufficiently well, we will
expect that the optimum obtained from the surrogate model is sufficiently
close the true optimum $\max_{\bm{x}}f(\bm{x})$ of the black-box function. That is, if we can
ensure $g\approx f$, then their optima can be off by a max error at
most $\|f-g\|_{\infty}$. 
\emph{Gaussian process} (GP) modeling is a predominantly popular choice for
building surrogate models $g$ in Bayesian optimization. The most
obvious reason for this choice of GP surrogate model is its convenient
conjugacy in Bayes' theorem and its well-studied covariance dependence
structures \citep{rasmussen_gaussian_2006,shah_studentt_2014}. And
the sequential sampling scheme can be guided by maximizing the acquisition
function within the Bayesian framework \citep{gramacy_particle_2010,gramacy_sequential_2015}.
We summarize the recent progress of the existing problems and challenges
pointed out by \citet{shahriari_taking_2016}, where a comprehensive
introduction to Bayesian optimization is given.
\begin{itemize}
\item \emph{Constrained Bayesian optimization.} A constrained Bayesian optimization
problem considers a domain which is not a full space. Notably, \citet{muller_surrogate_2019}
propose grid-based searches for optimization with hidden constraints,
following a line of research tackling mixed integer \citep{holmstrom_adaptive_2008},
unknown \citep{gramacy_optimization_2010} and noisy constraints \citep{letham_constrained_2019}. 
\item \emph{Cost sensitivity.} The problem of cost sensitivity arises in
computer experiment applications and can be incorporated in the Bayesian
optimization context \citep{brochu_tutorial_2010}. \citet{lee_costaware_2020}
recently propose a cost-aware metric, in combination with cost-effective
pilot samples. 
\item \emph{High-dimensional problems.} Most Bayesian optimization models
focus on low-dimensional domains \citep{gramacy_particle_2010,snoek_practical_2012}.
\citet{chen_joint_2012} propose a two-stage strategy to fit high-dimensional
GP surrogate models by dropping some dimensions. However, when there
are insufficient samples, high-dimensional over-parameterized surrogate
models may suffer.
\item \emph{Multi-task.} Multi-task optimization aims at approximating
multiple responses simultaneously. Following \citet{bonilla_multi-task_2008},
\citet{swersky_multi-task_2013} provide a unified framework for multiple-output
GP surrogates. Additive GP surrogate models are designed for multi-fidelity
applications with approximations \citep{kandasamy_multifidelity_2017}
or generalized acquisition functions \citep{song_general_2018}.
\item \emph{Freeze-thaw.} The freeze-thaw strategy considers a subcategory of
GP surrogate models where inner loops of iterative optimization are
needed \citep{swersky_freezethaw_2014}. This approach has been incorporated
into the iterative learning framework \citep{nguyen_bayesian_2020}.
Meanwhile, the criticisms of freeze-thaw approach are mostly based on its complicated
implementation and highly application-dependent nature \citep{dai_bayesian_2019}.
\end{itemize}
The main application that motivates the current paper is the tuning
problem in computational science. In the tuning problem, an expensive
black-box cost function $f$, typically the running time of a complicated simulation, with a moderate dimensional variable $\bm{x}$, typically a set of algorithmic tuning parameters, needs to be optimized with a limited sequential sample size. In addition,
another important challenge is the potential non-smoothness in the
black-box cost function. Tuning becomes more crucial in large-scale computational applications, and surrogate modeling with GP becomes an effective approach \citep{sid-lakhdar_multi-task_2019}. The GP modeling will produce continuous sample
paths \citep{cramer_stationary_2013}, but has difficulty in capturing
potential non-smoothness (e.g., jumps in black-box performance functions)
in the black-box function. This feature of GP surrogates makes it difficult
to model cost functions in tuning problems, whose black-box
functions often come with potential discontinuities. 

As a motivation, we consider a univariate cost function that
arises in a specific tuning problem. In a matrix multiplication problem
$AB=C$, the practice of blocking large matrices $A,B,C$ and performing
block-wise multiplication is done to improve the computational performance
\citep{hong_complexity_1981,blackford2002updated,bilmes1997optimizing,whaley2001automated,zee2016blis}. 
The black-box function $f$
is the measured computational speed ($\mathtt{Mflops/s}$) of this
matrix multiplication operation (matmul), which is determined by the efficiency of
the block matrix multiplication algorithm along with the potential measurement noise. The tuning parameter in this problem is the block size for these blocked
matrices. On one hand, we can increase the block size in order to
reduce the amount of data movement %
between fast and slow caches to speed up the
multiplication. On the other hand, since the processor cache is finite,
when the matrix block is too large to fit into the cache, it would cause cache misses
and hence a sudden
drop in the computational performance. 

\citet{hong_complexity_1981} use a pebble game model on an acyclic directed
graph to derive an analytical optimal bound for cost  function
of this application. The optimal block size is machine dependent
since processor cache sizes can differ, and also depends on the structure of the innermost loops.
We want to ``tune'' the optimal block size to attain the
fastest matrix multiplication. This motivating example drives us to
consider an optimization problem where the black-box function $f:\mathbb{Z}_{+}\rightarrow\mathbb{R}^{1}$
has potential non-smoothness with respect to the block size to be tuned (and where there is no explicit way to compute the optimal block size).

This application of matmul will be our example showing why different partitions may arise naturally in the context of tuning. We also provide multiple synthetic and real-world tuning  examples like SuperLU \citep{yamazaki_new_2012} showing the improvement when partitioning is well utilized in the optimization.
We want to optimize a non-smooth black-box function with a variant
of GP surrogate, which leads to our proposal below.

\subsection{\label{subsec:Non-smoothness-in-the-domain}Non-smoothness caused
by the domain}

The black-box function $f$ may be defined on a discrete domain like
$\mathbb{Z}^{d}$ instead of $\mathbb{R}^{d}$. This kind of problem
is known as a problem with \emph{integer constraints }\citep{muller_surrogate_2019,garrido-merchan_dealing_2020}.
Furthermore, if only some of the input dimensions are discrete and
the black-box function is defined as $f:\mathbb{Z}^{d-r}\times\mathbb{R}^{r}\rightarrow\mathbb{R}^{1}$,
then the problem has \emph{mixed-integer constraints}.
This kind of problem could be addressed with different grid-based search methods
and generalized into multiple-objective situations within the Bayesian
optimization framework \citep{holmstrom_adaptive_2008,gramacy_optimization_2010}. 

Usually, a GP surrogate model will be assumed to be fitted on a continuous
input domain. In the situation where the input domain is (partially)
discrete, a common practice is to discretize the continuous input
variables of a (univariate) GP surrogate model $g: \mathbb{R}^{d}\rightarrow\mathbb{R}^{1}$ whose mean function approximates $f$ with uncertainty quantification,
while the handling of categorical or integer variables in Bayesian optimism is ongoing research \citep{ru_bayesian_2020}. However, if we have a (piecewise)
continuous black-box function $f:\mathbb{R}^{d}\rightarrow\mathbb{R}^{1}$,
this simple discretization without appropriate adjustment in
the covariance structure of GP surrogate model may yield incorrect
results \citep{garrido-merchan_dealing_2020}. We mainly focus on
another kind of non-smoothness in the next section -- the non-smoothness
in the black-box function $f$ on a continuous domain.

\subsection{\label{subsec:Non-smoothness-in-the-black-box-function}Non-smoothness in the black-box function}

In this paper, we consider the non-smoothness
of the function $f$ of the following two types, where we assume the function domain is $\mathbb{R}^{d}$.

\begin{example}
For example, the function $f_{1}:\mathbb{R}^{1}\rightarrow\mathbb{R}^{1}$
\begin{align}
f_{1}(x)=\begin{cases}
-x+1 & x<0\\
x^{2} & x\geq0
\end{cases},
\end{align}
 defined on $\mathbb{R}^{1}$ is a piecewise continuous
function. The function $f_{1}$ is not smooth and has a ``jump'' at the point $x=0$ from 1 to 0 with minimum $0$
reached at $0$. 
\end{example}

Other than the non-smoothness caused by ``jumps'',
another kind of non-smoothness in the black-box function is caused by 
discontinuous derivatives.

\begin{example}
For example, the function $f_{2}:\mathbb{R}^{2}\rightarrow\mathbb{R}^{1}$
\begin{align}
f_{2}(\bm{x})=\begin{cases}
(\|\bm{x}\|-1) & \|\bm{x}\|>1\\
(\|\bm{x}\|-1)^{4} & \|\bm{x}\|\leq1
\end{cases},
\end{align}
 has support $\mathbb{R}^{2}$ and a discontinuous derivative 
on the circle $S^1:\|\bm{x}\|=1$, where it attains its minimum value of 0.
In special cases, such non-smoothness can be captured asymptotically
with sufficient samples \citep{luo_asymptotics_2020}. 
\end{example}

Although different choices of covariance kernels would adjust the
smoothness of the surrogate model $g$, a GP surrogate would not 
accurately model ``jumps''. Here is one example of the effect
of non-smooth black-box functions. 
Since there could be more than one parameter that specifies the kernel,
we use $\bm{\kappa}$ to denote all kernel parameters for kernel $K=K_{\bm{\kappa}}$
below for brevity. 

\begin{figure}[!htb]
\begin{adjustbox}{center}

\begin{tabular}{cccc}
\toprule 
$\nu=1/2$ & $\nu=3/2$ & $\nu=5/2$ & $\nu=\infty$ \tabularnewline
\midrule
\midrule 
\multicolumn{4}{c}{GP}\tabularnewline
\midrule 
\includegraphics[width=3.5cm]{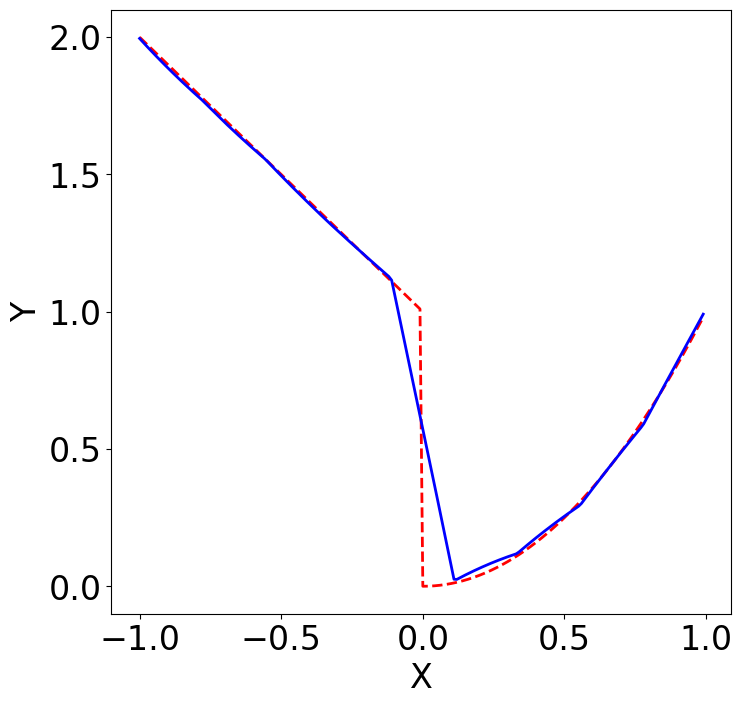} & \includegraphics[width=3.5cm]{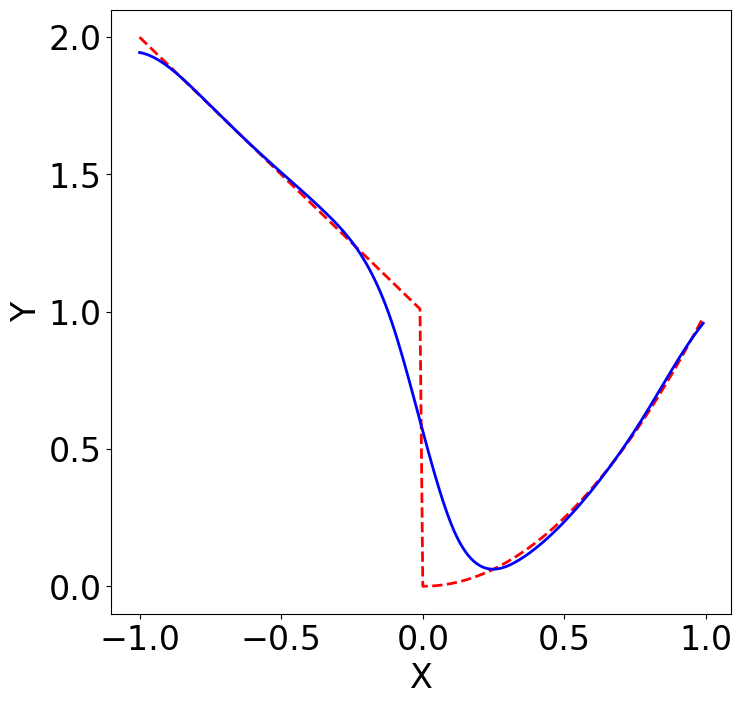} & \includegraphics[width=3.5cm]{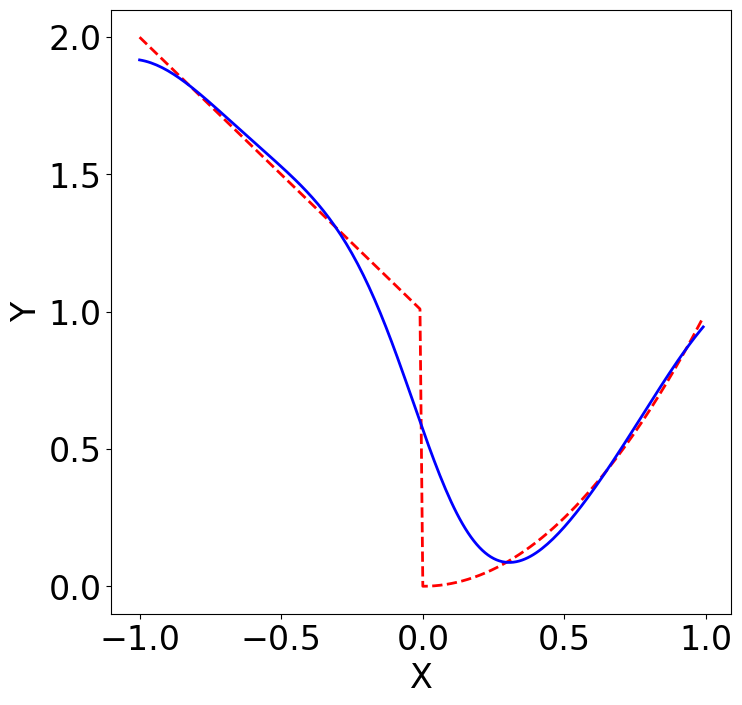} & \includegraphics[width=3.5cm]{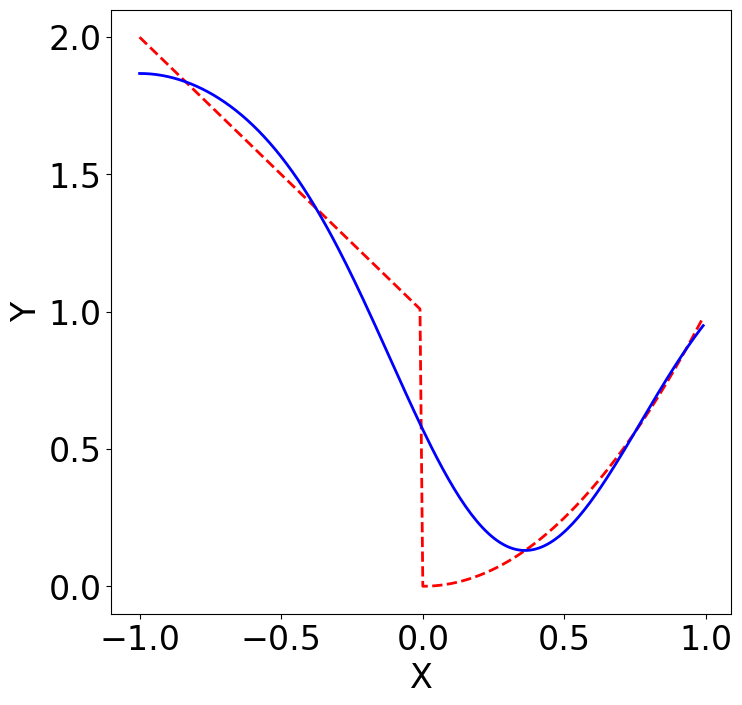}\tabularnewline
\midrule 
\multicolumn{4}{c}{additive GP (with 2 components supported on $(-\infty,0]$ and $[0,+\infty)$
respectively)}\tabularnewline
\midrule 
\includegraphics[width=3.5cm]{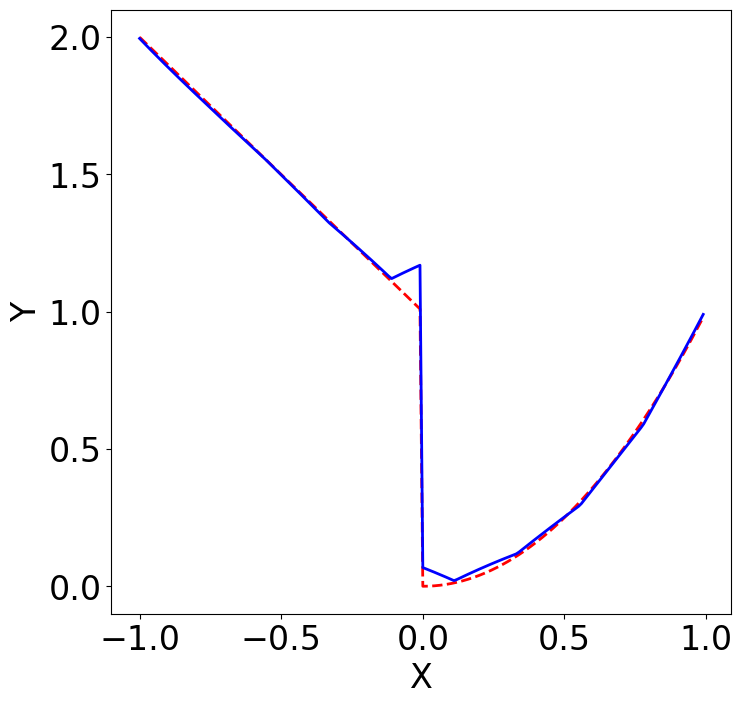} & \includegraphics[width=3.5cm]{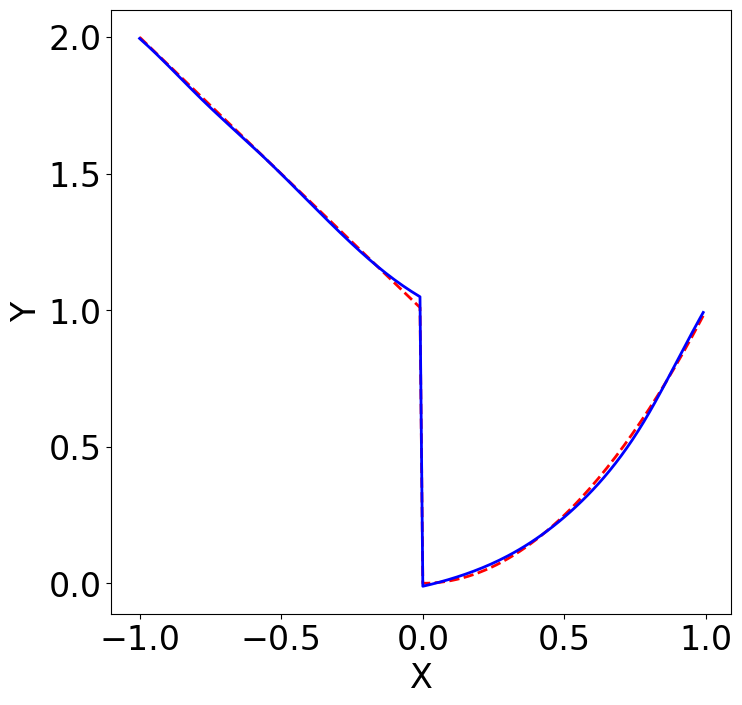} & \includegraphics[width=3.5cm]{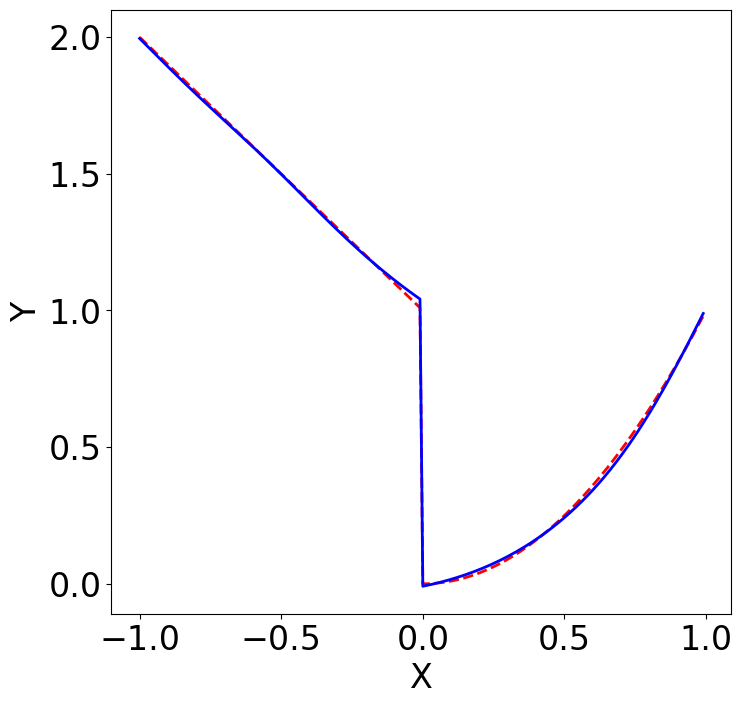} & \includegraphics[width=3.5cm]{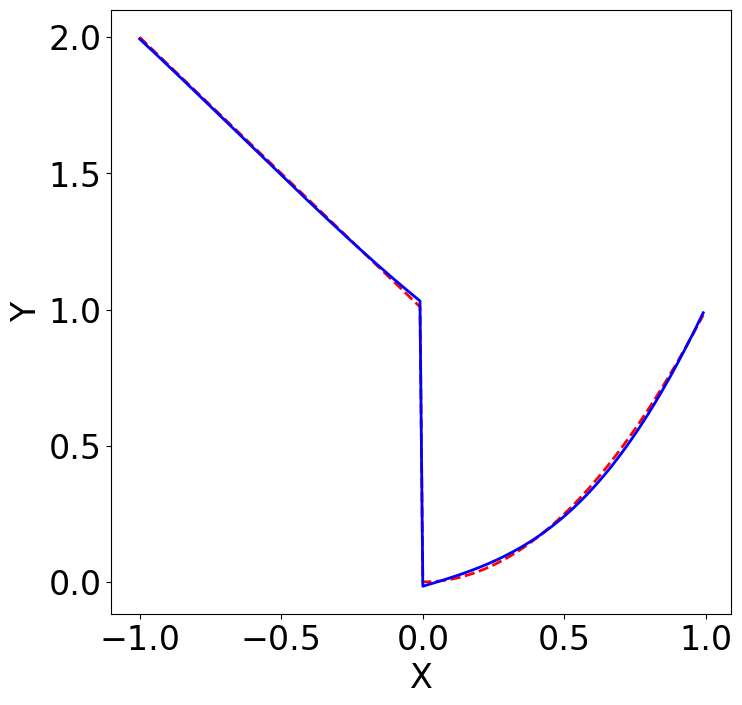}\tabularnewline
\bottomrule
\end{tabular}

\end{adjustbox}

\caption{\label{fig:Mean-function-extracted}Mean function extracted from the
fitted GP (first row) and piecewise GP (second row, with components
supported at $(-\infty,0]$ and $[0,+\infty)$ respectively) surrogate
model with $10$ equally spaced samples on $[-1,1]$ from $f_{1}$.
The blue solid line is the fitted surrogate mean function and the
red dashed line is the true function $f_{1}$.}
\end{figure}

In Figure \ref{fig:Mean-function-extracted}, we show the fitted mean
function of GP and piecewise GP surrogate models (with a deterministic
partition) in the unrealistic best-case situation where we know the exact location of the discontinuity. All models are based on the same 10 pilot 
samples drawn
from equally spaced locations on $[-1,1]$ from $f_{1}$ above, with Gaussian
noise with variance $0.01$. We provide surrogate fits with Mat\'ern
covariance kernels with parameters $\nu=1/2,3/2,5/2,\infty$, which
adjusts the smoothness of the GP surrogate.

For the simple GP surrogate models displayed in the first row, near
the non-smooth point $x=0$, we see a lack-of-fit between the black-box
function and the fitted mean function from the surrogate models. We
also fit a piecewise GP surrogate model $g=g_{1}+g_{2}$ where $g_{1}$ and $g_2$ are supported on $(-\infty,0]$ and $[0,+\infty)$ respectively. From
the second row of the figure, we can see that this additive approach captures
the non-smoothness that occurs at $x=0$, and the minimum obtained
from $g$ is closer to the true minimum at 0 for a single GP with
any kernel. An important observation is that an appropriate partition
of the input domain helps the surrogate model capture the non-smoothness,
compared to simple GP surrogate models.

The non-smoothness in $f_{2}$ is caused by discontinuous derivatives,
and the non-smoothness in $f_{1}$ is a ``jump''. This terminology
is more general than the ``change-point'' and ``change-surface'',
which are usually restricted to specific dimensions. The presence
of non-smoothness in the black-box function will cause GP surrogate
models $g$ to be mis-specified, and have a underlying  lack-of-fit to the black-box function
$f$. The existence of non-smoothness in $f$ presents challenges
to the optimization since the optima may no longer be close to the
true optimum when the black-box functions are non-smooth as shown
in Figure \ref{fig:Mean-function-extracted}. 

On one hand, mathematically, the non-smoothness should be considered in the context of continuity of functions. However, the mathematical definition requires arbitrarily many points to characterize discontinuities. 

On the other hand, in the surrogate modeling context, we only take finitely many sample points. Therefore,  we consider those points $x_2$ near a (fixed) point $x_1$ at which the finite difference $|f(x_1)-f(x_2)|/\|x_1-x_2\|$ exceeds a certain threshold to be ``non-smooth'' but those points where the finite difference ratio  falls below certain threshold to be ``smooth''. 
This is a proxy for the formal mathematical concept. Such a consideration motivates our later construction where we use $(x,\xi y)$ as the clustering criterion in our algorithm, with a factor $\xi$ for threshold adjustment. Detailed discussions are delayed to section \ref{subsec:Clustering-GP-(CGP)}.

This conflict between mathematical and modeled non-smoothness cannot be resolved without high-order derivative modeling (e.g., \citet{solak_derivative_2003}). One solution is to put yet another GP surrogate to model first-order derivatives and then use that as a criterion for clustering. This scheme can be generalized to higher order derivatives, but we leave this as future work. 

Formally, we have a definition of non-smoothness of a function $f$,
which captures both jumps and discontinuous derivatives:

\begin{definition}
\label{def: (Non-smoothness)-Let-}(Non-smoothness) Let $f$ be a
function of variable $\bm{x}\in\mathbb{R}^{d}$, we define a \emph{non-smooth
point} $\bm{x}_{0}$ of $f$ to be such a point that its first-order gradient  $\nabla_{\bm{x}}f$ is either unbounded or does not exist  at some point $\bm{x}$ in any open neighborhood of $\bm{x}_{0}$.
\end{definition}

The two examples above do not have gradients at certain points of their domains due to ``jumps'' or different left- and right- limits. Another example is the function $f_0(x) = x^{3/2}\cdot\sin(1/x)$, which  has a gradient 0 at $x=0$ but is not bounded near $x=0$. These kinds of non-smoothness can be observed in various black-box functions, as we shall see later.

The algorithm we introduce later does not use the definition, but we are motivating the choice of the clustering criterion $(\bm x,\xi y)$ (for the moment $\xi=1$, and its role will be clear later) using this specific definition, which will enable the surrogate fitting to take non-smoothness into account.

\subsection{Organization }
We have briefly reviewed  surrogate modeling and pointed out the problem
of non-smoothness in the tuning context. The organization of the rest
of this paper is as follows. In Section \ref{sec:Methodology}, we
describe our methodology of building a non-smooth GP surrogate model,
which we call a clustered GP (cGP) surrogate model, leveraging the partitioning scheme induced by the clustering to accommodate the potential non-smoothness. We first review
relevant methods in handling non-smoothness in subsection \ref{subsec:Existing-methods},
and \ref{subsec:A-unified-view}. Subsection \ref{subsec:Clustering-GP-(CGP)}
discusses the intuition and construction of the proposed model. Section
\ref{sec:Experiments} provides simulated experiments with benchmark
functions and analysis for the blocked matrix multiplication tuning
problem we mentioned in subsection \ref{subsec:Problem-and-Challenges}. Through these experiments, we observe that the surrogate-based optimization with cGP is better than GP while providing reasonable partitions. (For example, we gain around 5\% improvement with a limited number of evaluations when tuning the widely used LU factorization application in Section \ref{sec:high-D tuning}. )
Moderately high-dimensional experiments are also provided to evaluate
the proposed cGP surrogate model.
The paper concludes with a discussion
of the cGP surrogate model and future work in section \ref{sec:Discussion}.

\section{\label{sec:Methodology}Methodology}

From what we described above in Definition \ref{def: (Non-smoothness)-Let-}
and the example in Figure \ref{fig:Mean-function-extracted}, the
problem of tackling non-smoothness in the form of ``jumps'' or ``piecewise
continuity'' can be addressed by building a GP surrogate model with
appropriately chosen additive components. 

In Section \ref{subsec:Existing-methods}, we discuss the existing
GP methods that are relevant to consider non-smooth modeling. In the
context of the tuning problem, we further focus on modeling a potentially
non-smooth black-box function with limited sequential (and possibly
noisy) samples. Then, we present our proposed method that is suitable for
surrogate modeling in a high-dimensional domain with detailed algorithms
in Section \ref{subsec:Clustering-GP-(CGP)}. 

\subsection{\label{subsec:Existing-methods}Relevant methods for non-smooth modeling}

The first line of relevant research is the online change-point detection
problem with GP models \citep{adams_bayesian_2007,saatci_gaussian_2010}.
In a one-dimensional domain (e.g., a time-series), a \emph{change-point}
means that the function behavior is different before and after this
point. The observed response could be characterized using
different models. The problem involving change-point detection investigates
whether and when such a change of behavior occurs. The setup of sequential
sampling and inference is known as \emph{online} detection, in contrast
to \emph{offline} detection where the samples are fixed and the
inference is retrospective \citep{basseville_detection_1993,saatci_gaussian_2010}.
\citet{page_continuous_1954} first posed the problem of change-point
detection, but not until \citet{smith_bayesian_1975} was the change-point
problem studied with a Bayesian approach, and then \citet{barry_bayesian_1993}
put the offline inference into a fully Bayesian framework. However,
not until \citet{adams_bayesian_2007} did a popular Bayesian change-point
detection for online inference appear. This kind of online Bayesian
change-point model relies on the one-dimensional time domain,
making it hard to generalize to higher dimensions. 

The second line of relevant research are the non-stationary GP models.
In a broader (yet more vague) sense, another collection of research
that is related to non-smooth modeling is non-stationary GP modeling,
which is investigated in the context of spatial statistics (2-dimensional
domain). Since people use the term \emph{stationary} to refer to a
specific family of GP models with covariance kernels depending only
on pairwise point distance, GP models without such a covariance kernel could be referred as \emph{non-stationary}.
A stationary kernel (e.g., Mat\'ern kernels) takes the form of $K_{\bm{\kappa}}(\boldsymbol{x}_{i},\boldsymbol{x}_{j})=K_{\bm{\kappa}}(\|\boldsymbol{x}_{i}-\boldsymbol{x}_{j}\|)$,
while a non-stationary kernel could take the form that depends on
specific locations $\boldsymbol{x}_{i},\boldsymbol{x}_{j}$ \citep{paciorek_nonstationary_2004}.
In the context of offline surrogate models, \citet{krause_nonmyopic_2007}
propose a mixture model to address the non-stationarity based
on a 
partition of 
the input domain provided by the user. Furthermore,
\citet{martinez-cantin_locallybiased_2015} suggest that the partition
weight functions consist of a local part and a global part in order
to learn global local features including potential non-smoothness.
In a different direction, \citet{herlands_scalable_2016} propose
a model for the partition weight function to model general change-surfaces.
These non-stationary kernels are usually overly parameterized, which
might be a problem when there are only limited samples. 

The third line of relevant research are in the context of online
(non-GP) surrogate models. When the derivative information is not available, \citet{stoyanov_predicting_2017} model
piecewise constant functions for cracking patterns using hierarchical
grid-based methods, and generalize to piecewise polynomials in \citet{stoyanov_adaptive_2018, fuchs_simplex_2019}.
Although motivated by a similar problem, we do not see an immediate
adaptation of the grid-based method into Bayesian optimization framework,
which gives up the sequential sampling procedure. A simpler model that is often used in non-smoothness detection is MARS \citep{friedman_multivariate_1991}. 
When the derivative information is available, \citet{solak_derivative_2003} suggested an approach to model observation derivatives using GP directly in the specific situation of a dynamic system where the generative equations are known. However, this is not the usual situation in the tuning or data analysis context.

\subsection{\label{subsec:A-unified-view}Partitioning the input domain}

The observation we made in Figure \ref{fig:Mean-function-extracted}
is that, by fitting additive Gaussian components on appropriate partitions,
the non-smoothness can be captured in the fitted model. 

In the setting of limited samples, an overly parameterized surrogate
model would not provide a good fit. There may be more model parameters than the samples and no unique fit is
possible. Due to high dimensionality, grid search methods are no longer
efficient and the non-smoothness or a change of behavior would require
more samples to fit a surrogate model. These two challenges both urge
for a parsimonious surrogate model, which is able to model the non-smoothness
in an effective way. As a generalization of the piecewise GP shown
in Figure \ref{fig:Mean-function-extracted}, we believe a partition
based GP surrogate model is a good candidate for handling non-smoothness
in the tuning context, as will be explained below. With normalization of the inputs, we can assume the following without loss of generality. 
\begin{quote}
\textbf{Assumption.} Hereafter, we assume that our input domain is a $d$-dimensional unit hypercube $H^{d}=[0,1]^{d}\subset\mathbb{R}^{d}$
($d\geq1$) for the simplicity of discussion. Our methodology can
be modified to work for $\mathbb{R}^{d}$ and $\mathbb{Z}^{d}$.
\end{quote}
It is natural to
think that different components in such a surrogate model should be
fitted over different partition regions determined by the non-smooth
points. For instance, in Figure \ref{fig:Mean-function-extracted},
the only non-smooth point for $f_{1}$ is $\bm{x}_{*}=0$ and its
complement divides the domain $\mathbb{R}$ into $(-\infty,0]$ and
$[0,+\infty)$. 

As explained in Section \ref{subsec:Non-smoothness-in-the-black-box-function},
we are inspired by the partitioning method used for GP modeling in
general. For example, a tree-like partition is utilized in order to
handle non-stationarity in GP modeling \citep{chipman_bayesian_1998,chipman_bart_2010,luo_sparse_2020}.
Furthermore, overly parameterized change-surface models \citep{martinez-cantin_locallybiased_2015,herlands_scalable_2016}
can be perceived as partitioning the domain with the support of the
weight functions using the change-surfaces. The effect of partitioning
is restricting each surrogate model component to the corresponding
piece of the domain, and assembling these components to obtain the overall
surrogate. Existing partition-based GP models are usually for offline
situations, but in our context, the partition must be able to update according
to the sequential sampling. 

The tuning problem brings us the following challenges. Although our main
focus is to develop a novel modeling method to model non-smoothness,
we will show by experiments and examples that our model can address
the following challenges relatively well.
\begin{itemize}
\item \emph{Limited and sequential samples.} For tuning problems, we need
to fit a GP surrogate model with limited sequential samples. Limited
and sequential samples make fitting GP surrogate models more difficult
and dependent on the choice of samples in each sampling step. With
limited samples, it is impossible to fit (non-stationary) models with
over-parameterization.
\item \emph{High-dimensional generalization.} Bayesian optimization for
the tuning problem usually has a natural moderately high-dimensional domain
(e.g., $\mathbb{R}^{d}$ with $d>2$), but it is not immediately clear
how to generalize these partition-based GP methods into a high-dimensional
domain. Besides, the number  of parameters %
and the samples needed for fitting %
for surrogate models
also grows quickly in high dimensions.
\item \emph{Restrictive partition shape.} Most existing partition-based
GP methods have quite restrictive partition shapes (i.e., determined by a system of inequalities $t^-_i\leq x_i\leq t^+_i, i=1,\cdots,d$. \citep{gramacy2008bayesian, chipman_bart_2010}). When the true
non-smooth partition boundaries are not aligned with the coordinate
axes, rectangular partitions would not be appropriate for non-smoothness
modeling since we cannot expect non-smoothness along rectangular boundaries.
\end{itemize}

The clustered GP method we propose below could be considered as a
special partition-based GP, where the partition is induced by clusters
of the observed locations and their responses (i.e., the pairs $(\bm{x},y)$).
It not only has improved performance over classical GP surrogate models
when there are abundant samples, but also does not have the problem
of over-parameterization as the other existing partitioning models. With a reasonable computational cost, we design
a clustering induced partition to build a surrogate model. The idea
behind our method is to partition using decision boundaries of a clustering
method, which generalizes naturally to a domain of any dimension. A good
clustering would not only induce a partition adapting to non-smoothness,
but also guide the sequential sampling in the Bayesian optimization
procedure.

\subsection{\label{subsec:Clustering-GP-(CGP)}Clustered GP (cGP) surrogate model}

As we explained in Section \ref{subsec:A-unified-view}, partitioning
is central to our solution to the problem of non-smoothness in the black-box
function to be modeled. The intuition for our clustered
GP surrogate (cGP) model can be explained as follows. 

The proposed surrogate model uses the \emph{decision boundaries} (of
classifiers trained by cluster results on observations) to model
partition boundaries directly, with a modified acquisition function
weighted by the cluster sizes used in sequential sampling. We can
introduce clustering based on the joint input-response pair $(\bm{x},\xi y)$
in accordance with our sample-based definition of non-smoothness 
as follows. 

\begin{figure}[ht!]
\begin{adjustbox}{center}\includegraphics[height=12cm]{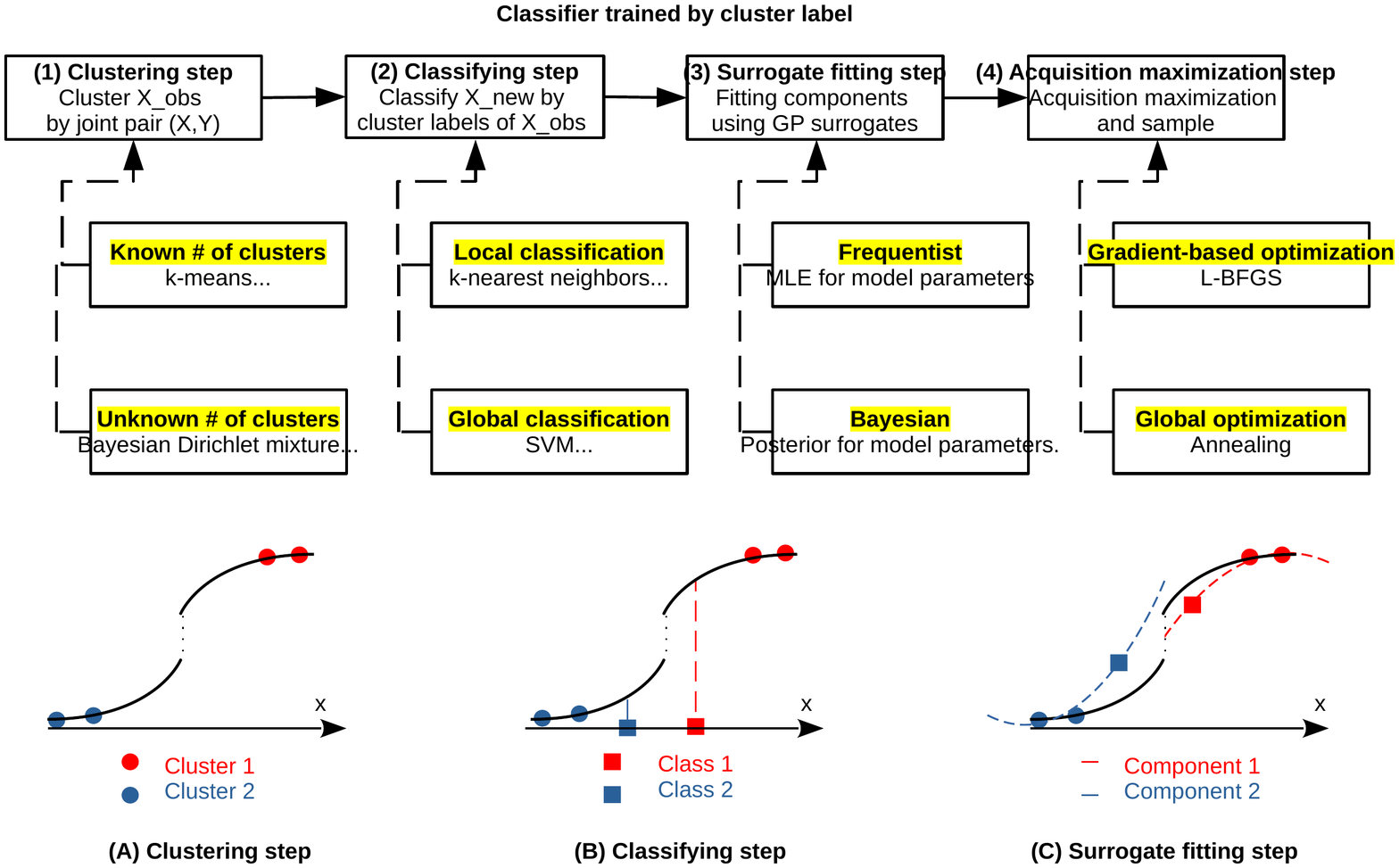}\end{adjustbox}
\caption{\label{fig:The-cluster-classify-scheme}The cluster-classify algorithm
for partitioning in the cGP surrogate model. In each step, we illustrate
options for the task and highlight these options. In the bottom
panel, we illustrate how the cluster-classify algorithm works for
a one-dimensional non-smooth function according to the bullet list
on page \pageref{enu:(Clustering-step)-Cluster}. In the illustration,
we use solid lines to indicate black-box function $f$ and dashed lines
for the fitted surrogate model $g$.\protect \\
(A) There are 4 pairs of observations $(\bm{x},y)$, and the clustering
algorithm clusters them into 2 clusters, represented by circles colored
in red and blue.\protect \\
(B) There are 2 new locations $\bm{x}_{1},\bm{x}_{2}$ represented
by squares. The classification algorithm classifies these two locations into different clusters, represented by different colors.\protect \\
(C) We use the corresponding surrogate component to predict at these
2 new locations $\bm{x}_{1},\bm{x}_{2}$. Using $(\bm{x}_{1},y_{1})$
and $(\bm{x}_{2},y_{2})$ we can update each component of the surrogate
model. 
}
\end{figure}

When $\bm{x}_{0}$ is not a non-smooth point, by our definition \ref{def: (Non-smoothness)-Let-}, we have a small $\epsilon$-neighborhood of $\bm{x}_{0}$ and in this neighborhood the gradient of $f$  
exists and is bounded.
Given our assumption that our domain is $H^d$ for continuous functions, we can find an $\bm{x}^*$ in this neighborhood such that 
\begin{align}
\left|f(\bm{x}_{1})-f(\bm{x}_{2})\right| & \leq \|\bm{x}_1-\bm{x}_2\|\cdot\|\nabla_{\bm{x}^*}f(\bm{x})\|,\label{eq:lipschitz} %
\end{align}
\label{enu:-is-continuous 4 cases}
The rationale
for using the joint input-response pair $(\bm{x},y),y=f(\bm{x})$
to determine whether $\bm{x}_{0}$ is a smooth point is that,
if we can find such a pair of $(\bm{x}_{1},y_{1})$ and $(\bm{x}_{2},y_{2})$
that $\|\bm{x}_{1}-\bm{x}_{2}\|$ is small but $|y_{1}-y_{2}|$ is
large then the necessary condition (\ref{eq:lipschitz}) for some
point $\bm{x}_{0}$ being a non-smooth point is violated. Therefore,
$(\bm{x}_{1},y_{1}),(\bm{x}_{2},y_{2})$ should belong to different
clusters. To summarize,
\begin{enumerate}
\item When  $\|\bm{x}_{1}-\bm{x}_{2}\|$ is small,
but $|y_{1}-y_{2}|$ large, then we put $(\bm{x}_{1},y_{1})$
and $(\bm{x}_{2},y_{2})$ into two distinct clusters.  
\item When $\|\bm{x}_{1}-\bm{x}_{2}\|$ is small,
and $|y_{1}-y_{2}|$ is small), then we expect 
$(\bm{x}_{1},y_{1})$ and $(\bm{x}_{2},y_{2})$ to belong to the
same cluster.
\item When  $\|\bm{x}_{1}-\bm{x}_{2}\|$ is
large, the value $|y_{1}-y_{2}|$ can be small or large, regardless
of smoothness of $f$. So we do not make a decision about whether
$(\bm{x}_1,y_1)$ and $(\bm{x}_2,y_2)$ are in the same cluster.
\end{enumerate}
Based on these considerations, we propose clustering
the $(d+1)$-dimensional pairs $(\bm{x},y),\bm{x}\in H^{d},y\in\mathbb{R}^{1}$ using one of the methods discussed below,
and then use these cluster labels to train a classifier. This classifier
would label points and induce a partition scheme on the input domain,
and hence on the new locations. One advantage of such clustering
is that we do not have to model the support of each additive component
nor the change-surface explicitly. This cluster-classify procedure
is described in Figure \ref{fig:The-cluster-classify-scheme}.

The clustering labels define classes and can be used for training
classifiers. The classifier partitions the domain $H^{d}$.
In addition,
we can maintain an interpretable model which indicates where the discontinuities
appear. Compared
to other partitioning-based methods \citep{gramacy_local_2014,luo_sparse_2020},
our cluster-classify algorithm creates a relatively parsimonious parameterization
for the partition-based surrogate GP model and generalizes to high-dimensional
domains well. 

With the additive Gaussian components induced by the clustering scheme, we can fit independent smaller Gaussian processes instead of one large one. This not only allows parallelization over additive components, but also normalization within each component across sequential samples. These features bring computational convenience.

The procedure
of sequential sampling with this cluster-classify surrogate model
can be summarized as below and also in Figure \ref{fig:The-cluster-classify-scheme}.

\begin{enumerate}
\item \label{enu:(Clustering-step)-Cluster}(Clustering step) Cluster pairs
$(\bm{x},y)$, with the clustering method you choose (e.g., if you
do not know the number of clusters, Bayesian Dirichlet GP mixture
(DGM) \citep{teh_dirichlet_2010} works\footnote{However, we still need to set the maximum number of clusters. Selection of hyper-parameters of clustering and classification are not avoidable, see also Section \ref{sec:Discussion}.};
if you know the number of clusters, $k$-means \citep{devroye2013probabilistic} works\footnote{By default, we use $k$-means with $k=3$ for the clustering step.}).
can attach cluster labels to samples $\bm{x}$ and form classes among
samples \footnote{The difference between clusters and classes is that clustering does
not need predetermined labels while classification does (which are called ``classes''). Our method uses the clustering labels as predetermined
labels for classifications.}. This step labels the observed locations and takes cluster labels as class labels for each location $\bm{x}$.
\item (Classifying step) Classify $\bm{x}_{new}$ by a classification method
trained by the labeled samples $\bm{x}$ (e.g., $k$-nearest neighbors (k-NN, \citep{devroye2013probabilistic})
provides a local classification\footnote{By default, we use $k$-nearest neighbors with $k=3$ for the classifying step.}; a support vector machine provides a
global classification.) The new predictive locations $\bm{x}_{new}$
(without knowledge of corresponding responses) can be assigned to one and only one cluster
class. In each sampling step, the clustering step (re-)labels the observations; and the classifying step is (re-)trained using the labels and predicts on domain. 
\item (Surrogate fitting step) Construct additive GP modeling over the partition
and fit it by either a frequentist or Bayesian approach.
\item (Acquisition maximization step) Maximize the modified acquisition
function (See details in Appendices \ref{sec:cGP-Algorithm} and \ref{sec:theory discussion}) for each component (e.g., expected improvement, which concerns maxima of the black-box function) using a selected
(gradient-based or global) optimization algorithm to determine the
next sequential sample. 
\end{enumerate}
To familiarize the novel notion of cluster-classify step, we use $k$-means and DGM (introduced below) in the cluster step of all examples; but use only k-NN in the classify step of all examples in the current paper. However, our implementation supports a wide variety of clustering and classification algorithms used in machine learning.
In the cluster-classify scheme,  the k-means use the proximity of pair $(x,\xi y)$ to cluster sampled observations; and the k-NN use only the  proximity of $x$ to classify \emph{any} location $x$ without the knowledge of $y$.

Following the Gaussian process modeling notation \citep{rasmussen_gaussian_2006},
the overall clustered GP surrogate model (with $k$ clusters) for
a sample \newline $\left\{ y_{1}(\bm{x}_{1}),\cdots,y_{n}(\bm{x}_{n})\right\} $
at $n$ different locations $\bm{X}=\{\bm{x}_1,\cdots,\bm{x}_n\}$ takes the following vector form: 
\begin{align}
\boldsymbol{y}(\bm{X}) & =(y_{1}(\bm{x}_{1}),\cdots,y_{n}(\bm{x}_{n}))^{T}=\sum_{j=1}^{k}\boldsymbol{f}_{j}(\bm{X})\cdot\bm{1}_{j}(\bm{X})+\boldsymbol{\epsilon},\label{eq:overall cGP model}
\end{align}
where the $\bm{f}_{j}(\bm{x})=(f_{j}(\bm{x}_1),\cdots, f_{j}(\bm{x}_n))^T$ %
denotes the vector of the $j$-th scalar function evaluated at $n$ different locations $\bm{X}=\{\bm{x}_1,\cdots,\bm{x}_n\}$ of all
$k$ Gaussian mean components; the overall Gaussian noise $\boldsymbol{\epsilon}$
has mean $0$ and variance $\sigma_{\epsilon}^{2}$. $\bm{1}_{j}(\bm{x})$
is an indicator function returning 1 if $\bm{x}$ is classified as
belonging to the $j$-th class and 0 otherwise. 

In contrast to a hierarchical partitioning model like the sparse additive Gaussian process (SAGP) 
model proposed by \citet{luo_sparse_2020,pmlr-v13-park10a}, the additive components of this model
are mutually independent due to the disjoint partition scheme induced
by our cluster-classify procedure. In the cGP model (\ref{eq:overall cGP model}),
each $\boldsymbol{f}_{j}$ is modeled by a zero-mean
Gaussian process prior, i.e., $\boldsymbol{f}_{j}(\bm{x})\sim N_{n}(\boldsymbol{0},\boldsymbol{K}_{j})$,
where the notation $N_{d}(\boldsymbol{m},\Sigma)$ denotes the $d$-dimensional
Gaussian distribution with mean vector $\boldsymbol{m}$ and covariance
matrix $\Sigma$, with covariance matrix $\boldsymbol{K}_{j}=\left[K_{\bm{\kappa}}(\boldsymbol{x}_{r},\boldsymbol{x}_{s})\cdot\bm{1}_{j}(\boldsymbol{x}_{r})\cdot\bm{1}_{j}(\boldsymbol{x}_{s})\right]_{r,s=1}^{n}$
and the covariance kernel $K_{\bm{\kappa}}$ modeling the dependence
between $\bm{x}$ within the same $j$-th class. 

To evaluate the acquisition functions and determine the next
sequential sample, we need to fit the surrogate model and draw predictions
from the surrogate model. The model parameters are the kernel parameter
$\bm{\kappa}$ that determines $K$ and the error variance $\sigma_{\epsilon}^{2}$.
One approach to fit the model is to estimate the parameters by maximizing
the likelihood function $L(\bm{\kappa},\boldsymbol{\sigma}_{\epsilon}^{2}\mid\bm{1}_{1}(\cdot),\cdots,\bm{1}_{k}(\cdot),\boldsymbol{y},\bm{x}_{1},\ldots,\bm{x}_{n})$
of (\ref{eq:overall cGP model}), which is usually known as the ``frequentist
approach''. Alternatively, we can impose priors on the parameters
$\bm{\kappa},\boldsymbol{\sigma}_{\epsilon}^{2}$ and use the modes
of their joint posterior 
\begin{align*}
& P(\bm{\kappa},\boldsymbol{\sigma}_{\epsilon}^{2}\mid\bm{1}_{1}(\cdot),\cdots,\bm{1}_{k}(\cdot),\boldsymbol{y},\bm{x}_{1},\ldots,\bm{x}_{n})\propto \\
& \underbrace{P(\boldsymbol{y}\mid\boldsymbol{\kappa},\boldsymbol{\sigma}_{\epsilon}^{2},\bm{f}_{1},\cdots,\bm{f}_{k})\prod_{j=1}^{k}P(\bm{f}_{j}\mid\bm{1}_{j}(\cdot),\bm{x}_{1},\ldots,\bm{x}_{n})}_{\text{Likelihood Function}}\cdot\\
 & \underbrace{P(\boldsymbol{\kappa}\mid\bm{1}_{1}(\cdot),\cdots,\bm{1}_{k}(\cdot))}_{\text{Kernel Prior}}\cdot\underbrace{P(\boldsymbol{\sigma}_{\epsilon}^{2})}_{\text{Error Prior}}.
\end{align*}
 as their estimates. This is usually known as the ``(fully) Bayesian
approach''. 

The Bayesian approach is preferred by multiple authors \citep{snoek_practical_2012,snoek_input_2014,gramacy_surrogates_2020}
for its flexibility of incorporating prior information and reproducibility.
However, the frequentist approach is still widely used due to its
computational efficiency and clear convergence criteria. 

\begin{figure}[t!]
\begin{center}\includegraphics[height=12cm]{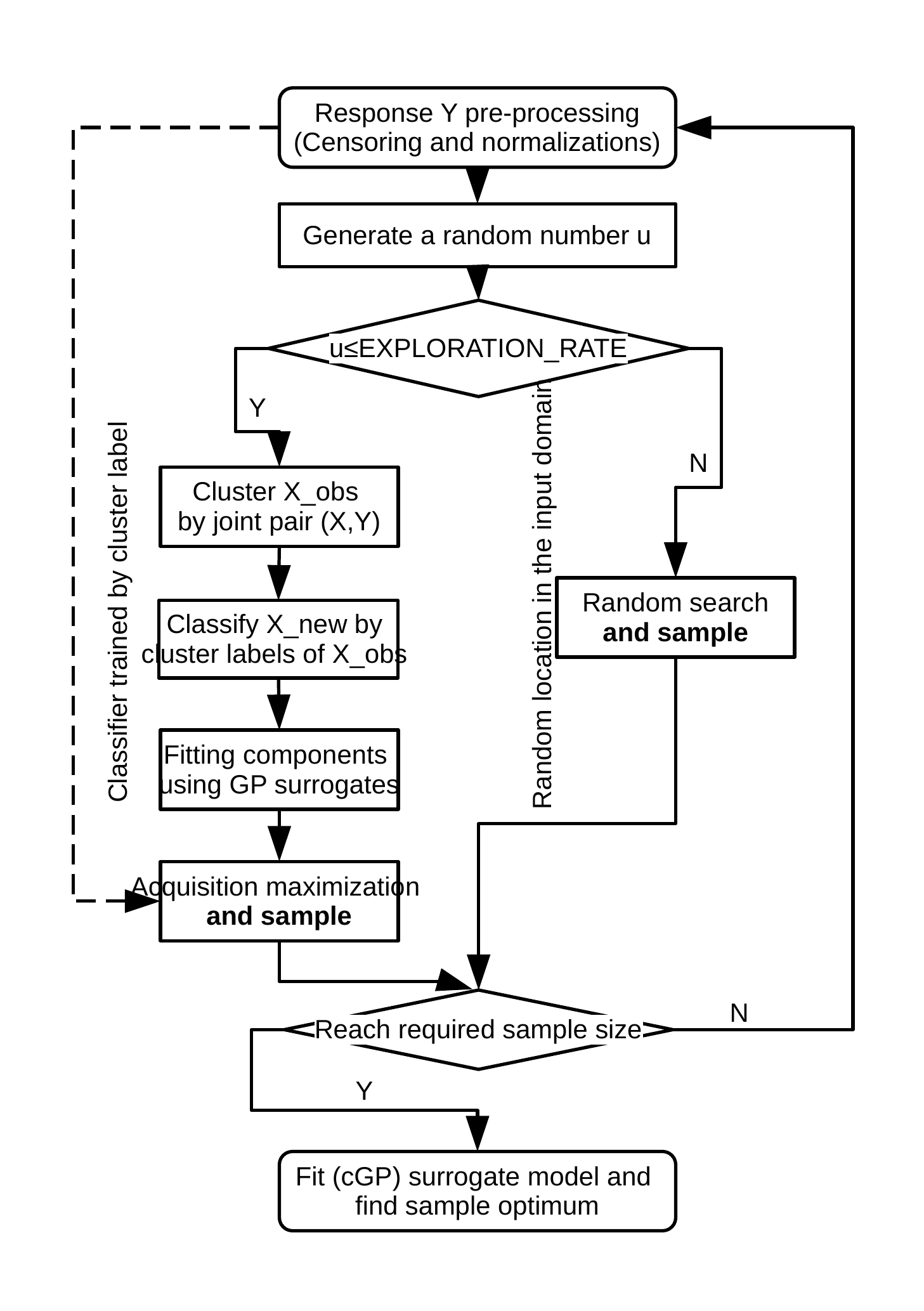}\end{center}
\caption{\label{fig:The-algorithm-flowchart}The algorithm flowchart for the cGP
surrogate model. The dashed line indicates the usual GP surrogate
model, which does not include a partition scheme. The detailed psuedo-code is available in Appendix \ref{sec:cGP-Algorithm}. Note that in each step, the clustering and the classifiers are re-trained with the newly sampled observation. Detailed pseudo-code for this algorithm is in Algorithm \ref{sec:cGP-Algorithm}. %
}
\end{figure}

Our main idea follows the scheme on page \pageref{enu:-is-continuous 4 cases}
and Figure \ref{fig:The-algorithm-flowchart}. We create labels for
locations in the input domain based on $(d+1)$-dimensional pairs $(\bm{x}_{obs},y_{obs})$
by clustering (unsupervised). Then, we compute the decision boundary
of a classifier trained by $\bm{x}_{obs}$ and its cluster labels,
to induce a partitioning of the input domain. We expect that sample points that belong to different regimes determined by the non-smoothness would be assigned to different partition components.

With this trained classifier (supervised), we can classify any point
$\bm{x}_{new}$ in the $H^{d}$ and form a partition over the domain.
Then the (weighted) acquisition function (by default we use the expected improvement (EI) as the acquisition function) based on the surrogate model is maximized over each component. This weighted acquisition function can be applied to different kinds of acquisition functions like UCB or PI \citep{hernandez-lobato_predictive_2014}, 
we provide a theoretic justification of this weighting scheme in the additive surrogate model in Appendix \ref{sec:theory discussion}. In short, such a weighting scheme and the exploration rate below would ensure convergence of the Bayesian optimization procedure based on cGP.

 In a broader view, this cluster-classify scheme also introduces a model selection problem in an online context. When the data accumulates, we want to determine the number of clusters dynamically. Currently, we retrain the clustering algorithm in every step, but to reduce the computational complexity, a more principled way might be to  derive an online version of selection criteria like the gap statistics developed by \citet{tibshirani2001estimating}. %

We have to point out again that we consider noisy observations, and we pick the expected improvement \citep{shahriari_taking_2016} as our default acquisition function. However, if a noiseless model is assumed, a different choice of acquisition function is needed.
Our algorithm is described in Algorithm \ref{clustered GP}, 
and we call
this model the \textbf{clustered Gaussian
process} (cGP) model. 

The cGP model is intrinsically related to at least two different kinds of (Bayesian) GP models. One is the treed 
GP model \citep{gramacy2008bayesian} that partitions the input domain using a dynamic tree (hence not clear how acquisition function can be elicited for an online setting), while the cGP uses a more flexible way of component creation via the cluster-classify step. The other is  a (two-layer) hierarchical model with input clusters \citep{pmlr-v13-park10a}, but the hierarchical model requires latent parameters and user-specific clusters, and it is unclear whether the existing acquisition functions are still suitable if used in an online setting, compared to the cGP model as a simple additive model without latent parameters.

In the implementation of the cGP model, we introduce the notion of exploration
rate, following the practice described by \citet{bull_convergence_2011} (in Definition 4) in the spirit of reinforcement learning \citep{sutton_reinforcement_2018}. The\emph{ exploration rate} is the probability that the next
sequential example would be sampled according to the acquisition maximization.
When the exploration rate is exactly 0, the sequential samples are all
randomly chosen without referring to the acquisition function at all.
When the exploration rate is exactly $1$, the sequential samples are
all sampled according to the maximizer of the acquisition function. 
In our specification, we default the exploration rate to be 0.8. It is
worth pointing out that adjusting the exploration rate would affect
the efficiency with which the sequential sampling explores the domain. In the algorithm described in 
Appendix \ref{sec:cGP-Algorithm}, we can see how this exploration
rate controls alternating between acquisition maximization and random
sampling schemes. We introduce this element to allow different sequential
sampling schemes and investigate its effect in Section \ref{subsec:Simple-illustrations-of}. 

Another minor adjustment we introduce in our implementation is to add a boundary penalty function to the acquisition function. In the problem of handling non-smoothness, it seems that it is of benefit to sample more near the boundary of the partition. In an offline context, \citet{park2011domain} had discussed thoroughly about the effect of partition boundaries in GP modeling. In the online optimization context, we implement a user-specified additive penalty function (normalized to the same scale) adding to the acquisition function, therefore, the maximization step would be biased towards the boundary of a component. For instance, the penalty function of a specific component can be configured to be proportional to the squared distance of a point to the boundary of this component. As the partition components determined by the cGP algorithm updates according to sequential sampling, this penalty function can also be updated to account for updated partitions as well. We recommend a careful penalty adjusted to the design and constraints on the domain or no penalty by default. 

Our cGP model can address the challenges in Section \ref{subsec:A-unified-view} while not introducing too much complexity in surrogate modeling:
\begin{itemize}
\item \emph{Limited and sequential samples.} We do not introduce more complex
parameterizations for the partitioning scheme induced by clustering,
except for the parameters required by the additive GP components. Even
with few samples, unlike more parameterized and complex  models \citep{damianou2013deep,herlands_scalable_2016}, cGP models can still be fitted in an online context.
\item \emph{High-dimensional generalizations.} We do not have a high computational
cost even in a high-dimensional domain, since the clustering and classification
algorithm are usually not affected by the curse of dimensionality.
The major computational cost is fitting the component GP surrogate
model.
\item \emph{Restrictive partition shape.} We introduce the cluster-classify
procedure to determine the partition scheme. Since the classification
boundary determines the partitions, we expect that the shape of partitions
would reflect non-smoothness better.
\end{itemize}
We also point out that cGP is unable to model non-smoothness
within a single component by its construction. In an extreme situation,
we pick only two points, $x_{1}=-0.5$ and $x_{2}=0.5$ for $f_{1}$
function with $f_{1}(x_{1})=1.5,f_{1}(x_{2})=0.25$ as shown in 
Figure \ref{fig:Mean-function-extracted}.
If we treat $x_{1},x_{2}$ as
one cluster, then we fail to capture the non-smooth point $x_{*}=0$
and will have a constant mean function at $0.375$. If we treat
$x_{1},x_{2}$ as two clusters, then we will have piecewise constant
functions of values $0.5$ and $0.25$ respectively with a non-smooth
change-point at $0$. This problem is intrinsic to any partition-based
method with a mis-specified partition scheme. Empirically, we observe
that this is not a too serious issue for cGP if we have enough pilot
samples. 

\section{\label{sec:Experiments}Experiments}

Subsection \ref{subsec:Simple-illustrations-of} starts 
with simple illustrations of our proposed cGP model. Then, subsection
\ref{subsec: Tuning} revisits the motivating tuning problem
we briefly mentioned in Section \ref{subsec:Problem-and-Challenges}
and qualitatively compares the performance of different surrogate models. We first showed that cGP is a competitive alternative for simple smooth and non-smooth functions. Then in matmul, piston and SuperLU applications, our cGP model shows considerable improvement in tuning results. In specific synthetic functions (i.e., Bukin.N6), the improvement is strict in 90\% of repeated experiments. 
The partitioned regime identified by the cGP model is also described
and discussed.

\subsection{\label{subsec:Simple-illustrations-of}Synthetic Studies}
To establish an intuition of how the cGP surrogate model and Algorithm \ref{clustered GP} behave, we show two synthetic examples in $H^{2}\subset\mathbb{R}^{2}$.
These low-dimensional examples allow us to examine the results visually
and qualitatively along with quantitative summaries. 
\subsubsection{Smooth functions}
First, we want to consider a smooth black-box function and see if
cGP is competitive against GP surrogates when there is no non-smoothness. In this situation, we expect a simple GP surrogate to perform the best, but expect that cGP models are also competative. Since when we choose the number of clusters $k=1$, the cGP model
reduces to the usual GP model, we also expect that GP and cGP would have
similar fits when $k$ is small. In the situation where the black-box
function has no non-smoothness, for example, we take the simple function $f_{3}(\bm{x})=f_{3}(x_{1},x_{2})=1/(1 + (x_1 - 0.25)^2 + (x_2 - 0.25)^2)$.
We examine one sample of cGP fits (with Mat\'{e}rn $3/2$ kernels) for $k=1,2,\cdots,5$
in Figure \ref{fig:2d smooth black-box function compare} and average performance in Table \ref{tab:average_Fig31_and_Fig32}.

When we try to maximize the function, the two key statistics we observe are $$\Delta\arg\max\coloneqq\|\arg\max_{\bm{x}}(f)-\text{sample maximum point }\bm{x}\|_2$$  the $L_2$ distance between sample maximum point $\bm{x}$ and the exact maximum point,  and the $$\Delta\max\coloneqq|f_\max-\text{sample maximal observed }y|$$ which is the difference between sample maximum function value $y$ and the exact maximum value $f_\max=f(\arg\max_{\bm{x}}f)$. 

In the optimization context, smaller statistics $\Delta\arg\max$ indicate better performance of the surrogate model, which leads to a sample maximum closer to the truth. And smaller statistics $\Delta\max$ indicates the surrogate model finds an (possibly noisy) observed value that is closer to the true function value. 

We can see that for this $f_3$ with the unique maximum 1 attained at $(1/4,1/4)$, 
only the case $k=3$ gives significantly worse performance than the simple GP models in one fit in Figure \ref{fig:2d smooth black-box function compare}, but reasonably close to simple GP on average. 
And we can see that cGP performs very closely to the simple GP surrogate as shown in Table \ref{tab:average_Fig31_and_Fig32}; as the number of $k$ increases, the surrogate fits improve. 
\begin{table}[ht!]
    \centering
\begin{tabular}{c|cc|cc}
\toprule 
Surrogate & \multicolumn{2}{c}{$f_{3}$ (smooth, Figure 3.1)} & \multicolumn{2}{c}{$f_{4}$ (non-smooth, Figure 3.2)}\tabularnewline
\midrule 
 & $\Delta\arg\max$ & $\Delta\max$ & $\Delta\arg\max$ & $\Delta\max$\tabularnewline
\midrule 
GP ($k=1$) & \textbf{0.018718} & \textbf{0.000349} & 0.082762 & 0.006721\tabularnewline
\midrule 
cGP ($k=2$, $k$-means) & 0.034503 & 0.001182 & 0.067821 & 0.004524\tabularnewline
\midrule 
cGP ($k=3$, $k$-means) & 0.041824 & 0.001730 & 0.085938 & 0.007180\tabularnewline
\midrule 
cGP ($k=4$, $k$-means) & 0.028472 & 0.000808 & 0.078971 & 0.006084\tabularnewline
\midrule 
cGP ($k=5$, $k$-means) & 0.026969 & 0.000724 & /%
& 
/%
\tabularnewline
\midrule 
cGP ($k=2$, DGM) & / & / & 0.082752 & 0.006623\tabularnewline
\midrule 
cGP ($k=3$, DGM) & / & / & 0.078508 & 0.006031\tabularnewline
\midrule 
cGP ($k=4$, DGM) & / & / & \textbf{0.067031} & \textbf{0.004412}\tabularnewline
\midrule 
cGP (partitioned) & / & / & 0.160008 & 0.022961\tabularnewline
\bottomrule
\end{tabular}
    \caption{Average statistics $\Delta\arg\max$ and  $\Delta\max$ for the repeated 50 experiments (different random pilot samples) for synthetic functions $f_3$ and $f_4$. The smallest statistics (indicating the best model in terms of this statistics) in each column are shown in bold fonts. %
    \label{tab:average_Fig31_and_Fig32}}
\end{table}

This set
of synthetic experimental results supports that the cGP method performs
reasonably when the black-box function is  smooth and would
not create a lot of non-smoothness in the fitted surrogate model.
Averaging over multiple runs with different pilot samples leads to similar results, therefore, we conclude that adopting the cGP model would not create comparable tuning results on average compared to GP surrogates even there is no non-smoothness at all.

\subsubsection{Non-smooth functions}
Second, we want to see how cGP performs when non-smoothness exists
in the black-box function, compared to GP. 
We create some non-smoothness in the function $f_3$ along the line $x_2=0$. We use the indicator function $\bm{1}(\cdot)$ to complete the construction. 
$f_{4}(\bm{x})=f_{4}(x_{1},x_{2})=\bm{1}(x_2>0)\cdot(1/(1 + (x_1 - 0.25)^2 + (x_2 - 0.25)^2)) + \bm{1}(x_2\leq0)\cdot(0.25/(1 + (x_1^2 + x_2^2)) )  $
as shown below with the same unique exact maximum point (1/4,1/4) as $f_3$. Therefore, if the non-smoothness introduced by the indicator functions does not affect the surrogate-based optimization at all, then we would be able to locate the maximum value of $f_4$ as easily as $f_3$. We examine the mean of cGP fits (with Mat\'ern $3/2$
kernels) for $k=1,2,\cdots,4$ in Figure \ref{fig:2d non-smooth black-box function compare} using both $k$-means and Dirichlet Gaussian mixture (DGM) as clustering algorithms.
In this case, $k=1$ (i.e., simple GP surrogate) is underfitting and cGP with $k=3,4$
are overfitting since 
we know from its expression that there are exactly two continuous regimes of this function, separated by the straight line $x_2=0$, 
in the true black-box function $f_4$.

Since the cGP model consists of two main parts, the cluster-classify step to get the partitions and modeling on the partitions, we want to explore how the cGP model works when the exact partitioning is known, and compare to the results when the partitions must be computed.
Following the practice in Figure \ref{fig:Mean-function-extracted}, we get a partitioned cGP assuming that we know the partition is given by $x_2=0$. The panel ``cGP (partitioned)'' in Figure \ref{fig:2d non-smooth black-box function compare} shows that when we know the partition exactly, an additive GP (i.e., cGP with known partition) produces the best surrogate fit among all models when compared to the truth of $f_4$. 
In short, if we know the exact partition, then we should not use cluster-classify step, but the model with the exact partition components  directly (i.e., additive GP) in this sequential sampling context. Our algorithm Algorithm \ref{clustered GP} can be shown to converge when the exact partition is fixed, see Proposition  \ref{prop:fixed partition GP convergence} in Appendix E.

When $k=3$, the non-smoothness
at $x_{2}=0$ cannot be modeled very appropriately with either $k$-means or DGM as we observe. The
cGP with $k=1$ (i.e, GP) is over-smoothed, while the cGP with $k$-means ($k= 4$) does not capture a transition boundary near $x_{2}=0$ well. 
The cGP with $k$-means($k=2$) and DGM ($k=3$) captures the $x_{2}=0$ better, with optima similar to the true black-box function. %
The cGP with DGMs all have smaller $\Delta\arg\max$ compared to GP and reasonable improvement, while cGP models with $k$-means ($k=2,4$) DGM ($k=2,3,4$) have better $\Delta\max$ and $\Delta\arg\max$ compared to a simple GP on average as observed from Table \ref{tab:average_Fig31_and_Fig32}. This means that with the same limited sample budget, cGP models (with appropriate clustering) get closer to the true maximum value. 
We observe that a cGP model with DGM performs better in terms of identifying clusters. The number of clusters is selected by the DGM with a prespecified maximal number of clusters \citep{teh_dirichlet_2010}. DGM does not require us to know the number of clusters
$k$ exactly but the algorithm learns this number from sequentially sampled
data. As shown, we can see that DGM with the maximal number of components being 3 adjust the number of partitions better and performs better compared to $k$-means with $k=3$, as shown by one run in Figure \ref{fig:2d non-smooth black-box function compare}. 

One more observation we made based on the average statistics shown in Table \ref{tab:average_Fig31_and_Fig32} is that we can see that compared to $f_3$, $f_4$ tends to inflate the two statistics, no matter what surrogate model we use. This strengthens and is consistent with our intuition established in Figure \ref{fig:Mean-function-extracted} that when there exists non-smoothness, the Bayesian optimization based on surrogate models would encounter difficulties  %
due to the lack of fit near the non-smooth points.
In this non-smooth example, we only change the maximal  number of clusters allowed in DGM (e.g., in Figure \ref{fig:2d non-smooth black-box function compare}, we show the $k$ parameter and the number of clusters finally chosen as well). 

\begin{figure}[!t]
\begin{adjustbox}{center}

\begin{tabular}{ccc}
\toprule 
black-box $f_{3}\in[8/33,1]$ & GP ($k=1$) & $k$-means ($k=2$)\tabularnewline
\midrule
\midrule 
\includegraphics[width=5cm]{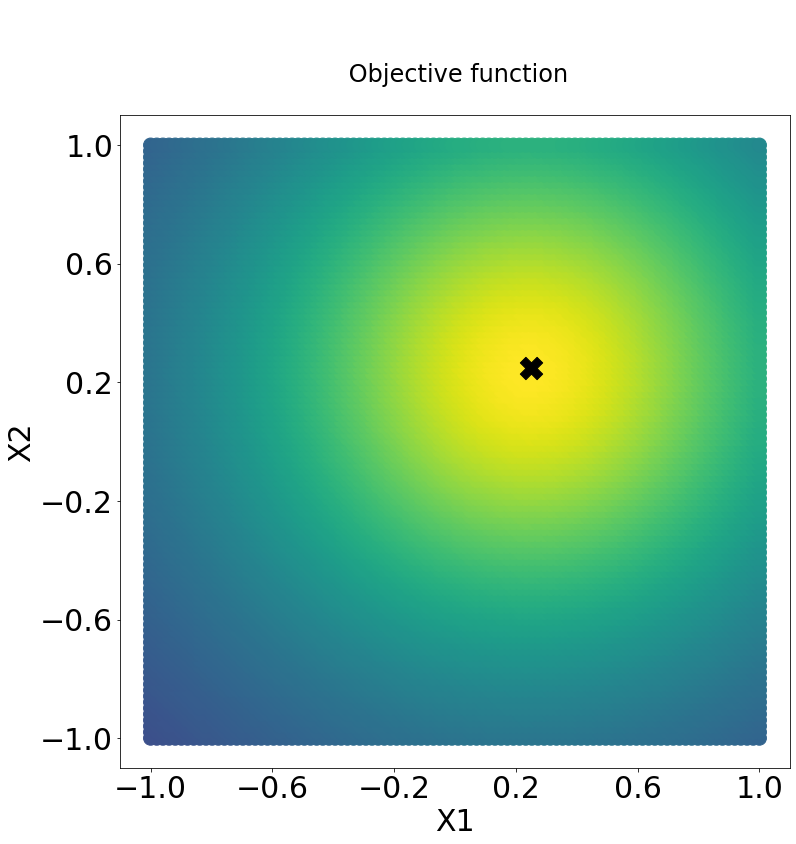} & \includegraphics[width=5cm]{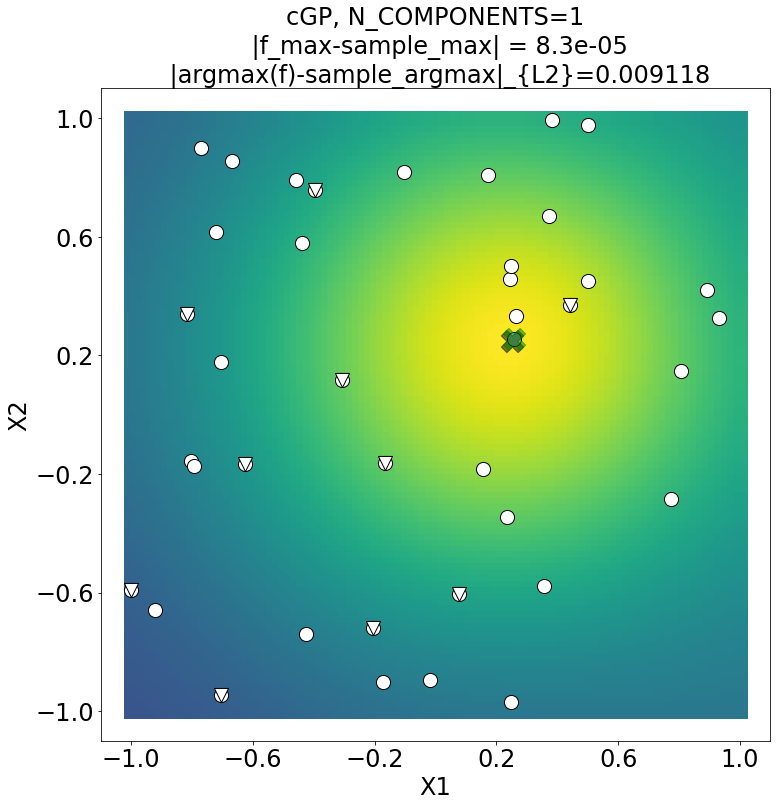} & \includegraphics[width=5cm]{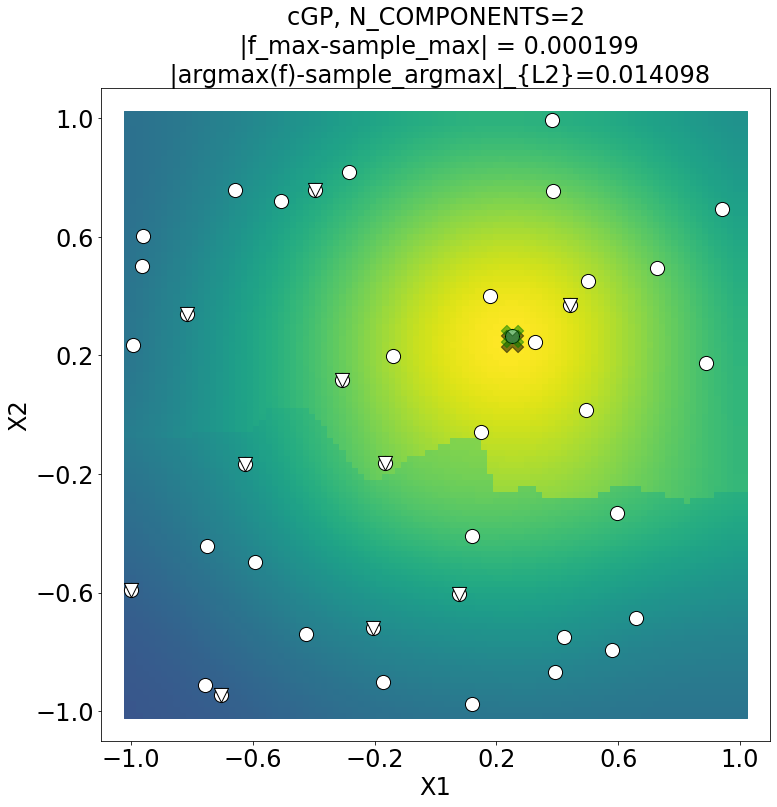}\tabularnewline
\midrule 
$k$-means ($k=3$) & $k$-means ($k=4$) & $k$-means ($k=5$)\tabularnewline
\midrule
\midrule 
\includegraphics[width=5cm]{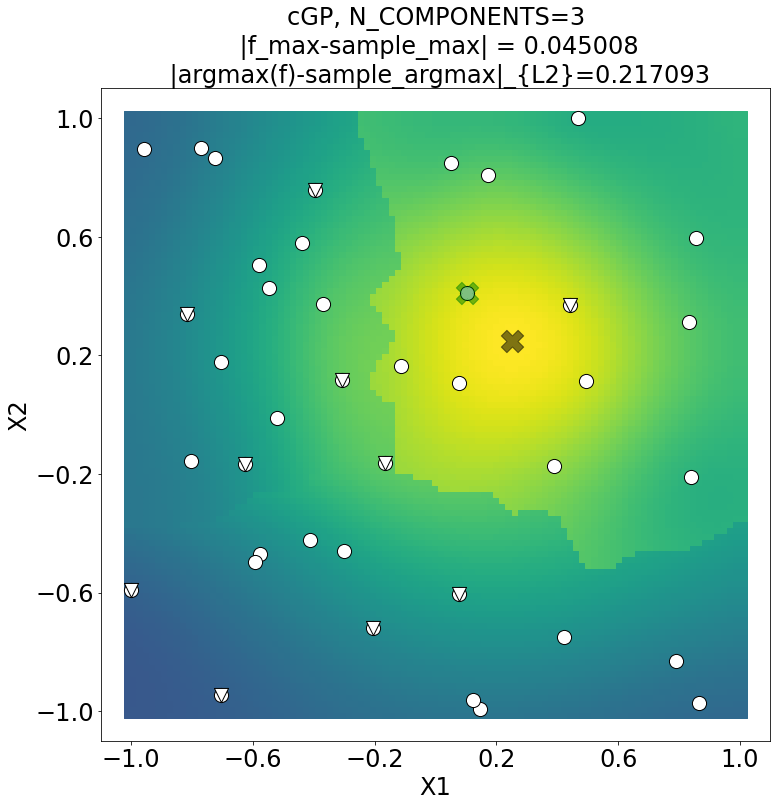} & \includegraphics[width=5cm]{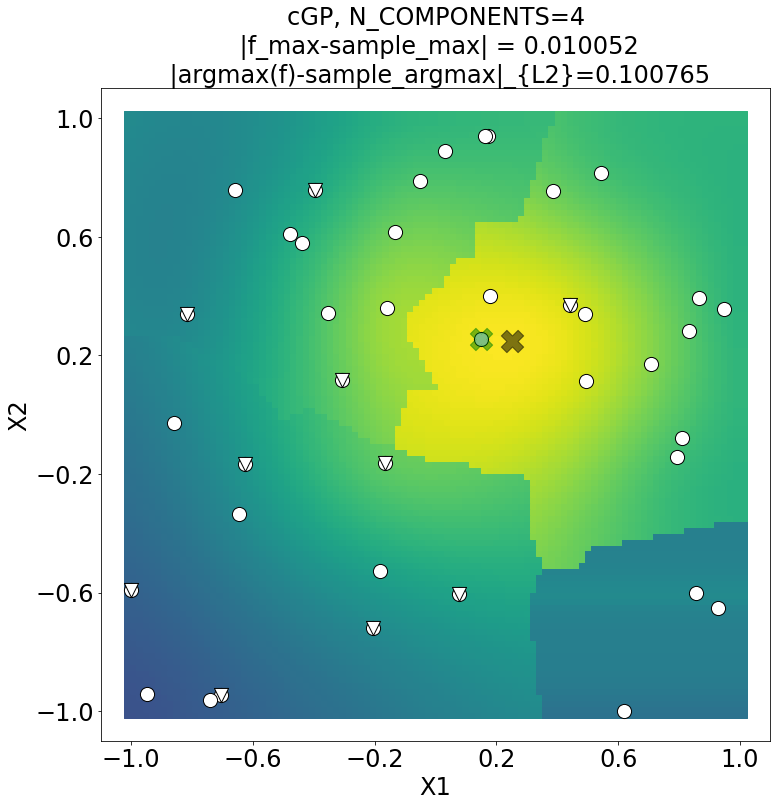} & \includegraphics[width=5cm]{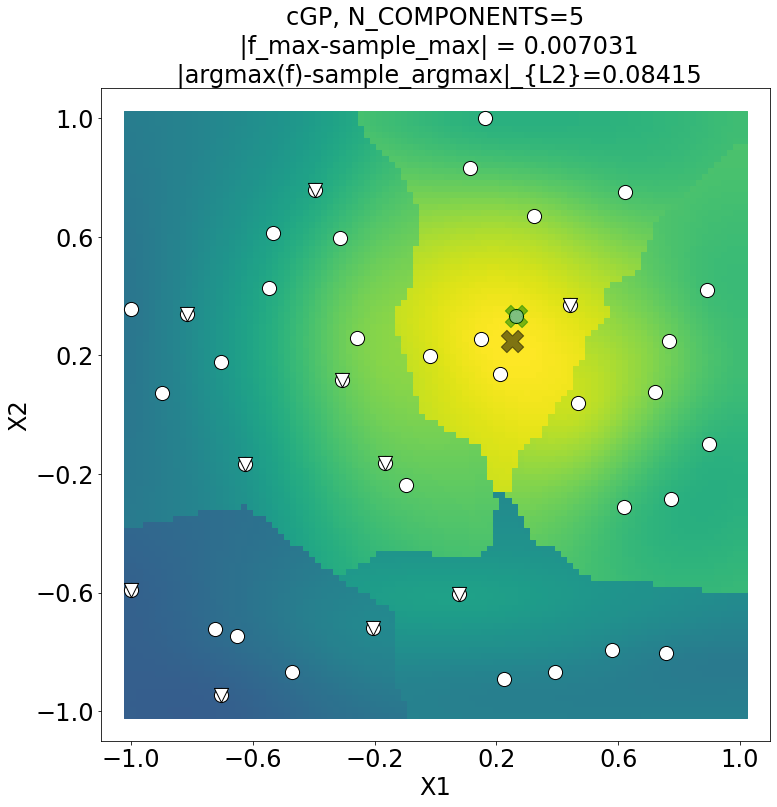}\tabularnewline
\bottomrule
\end{tabular}

\end{adjustbox}

\caption{\label{fig:2d smooth black-box function compare} Mean function and
partition scheme extracted from the GP surrogate model evaluated on
$100\times100$ equally spaced grids on $[-1,1]^{2}$ with weighted
expected improvement acquisition function and $n=10$ random pilot
samples (triangles), $n=30$ sequential samples (circles) from $f_3$. 
We use the same clustering methods with different assumed numbers of
clusters: $k$-means (with $k=1,2,3,4,5$). We show exact minimal point by a black cross; sample minimal point by a green cross.}
\end{figure}
\FloatBarrier

\begin{figure}[!t]
\vspace*{-1cm}
\begin{adjustbox}{center}

\begin{tabular}{ccc}
\toprule 
black-box $f_{4}\in[1/3,1]$ & cGP (partitioned) & GP ($k=1$)\tabularnewline
\midrule
\midrule 
\includegraphics[width=5cm]{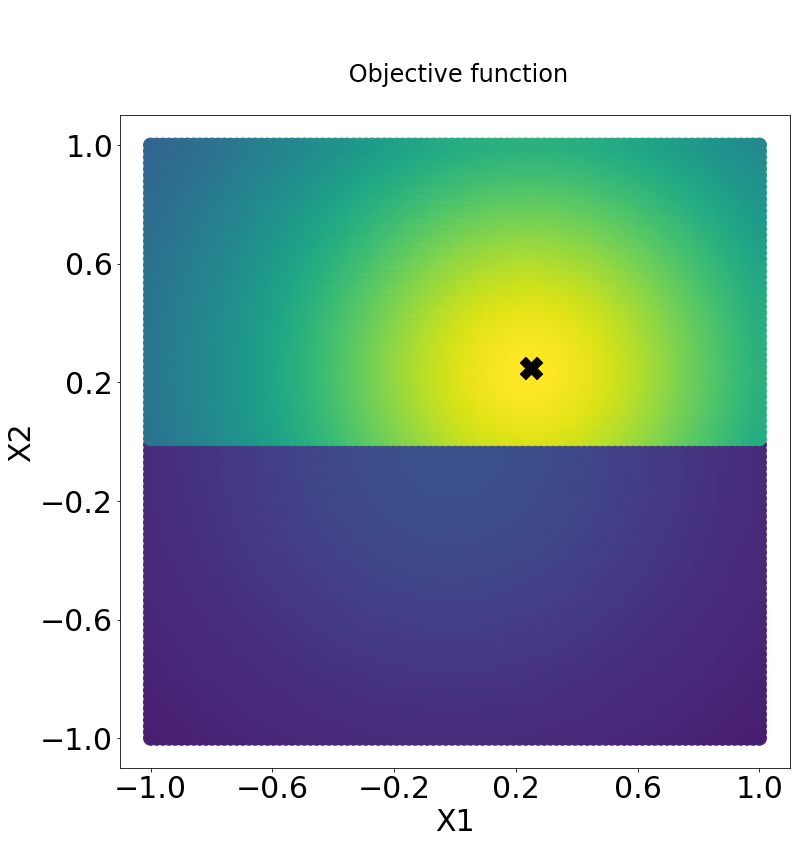} & \includegraphics[width=5cm]{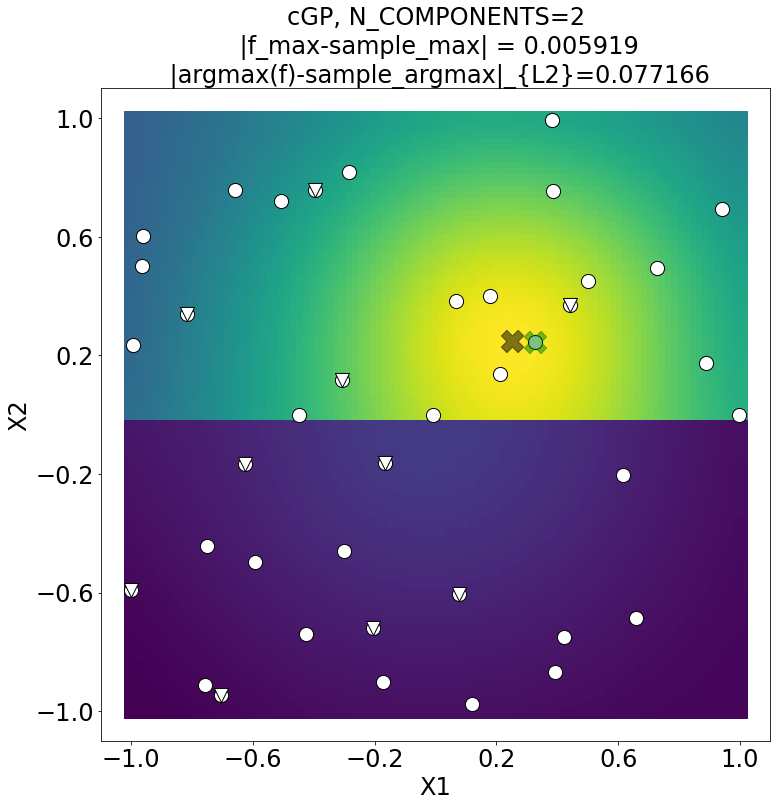} & \includegraphics[width=5cm]{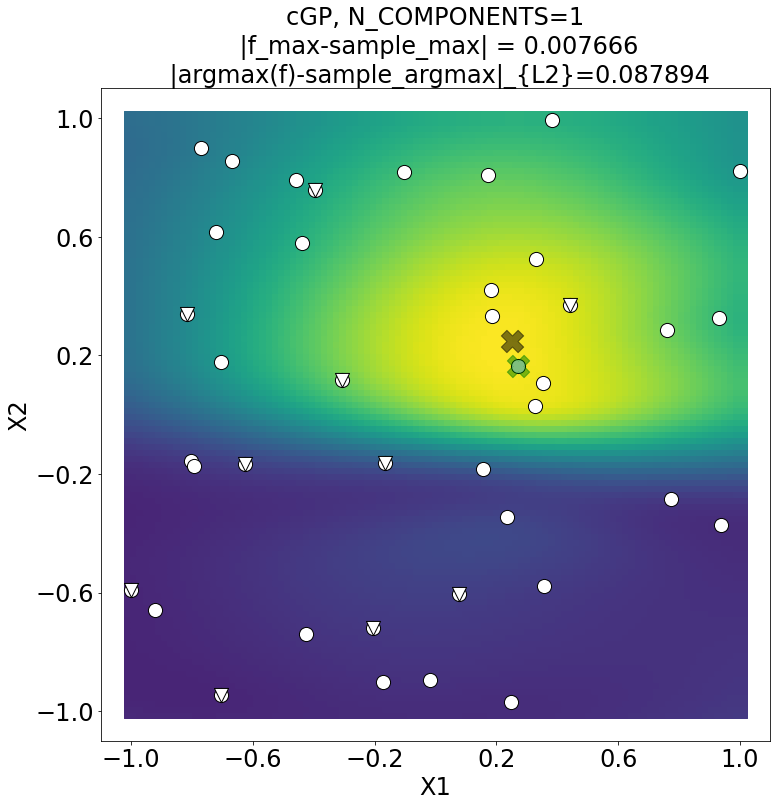}\tabularnewline
\midrule 
$k$-means ($k=2$) & $k$-means ($k=3$) & $k$-means ($k=4$)\tabularnewline
\midrule
\includegraphics[width=5cm]{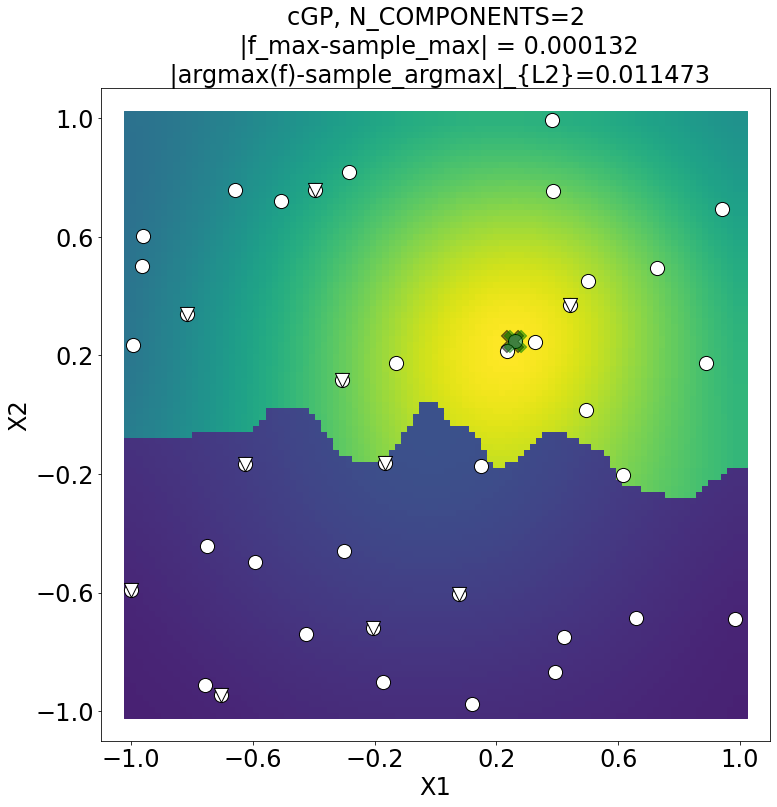} & \includegraphics[width=5cm]{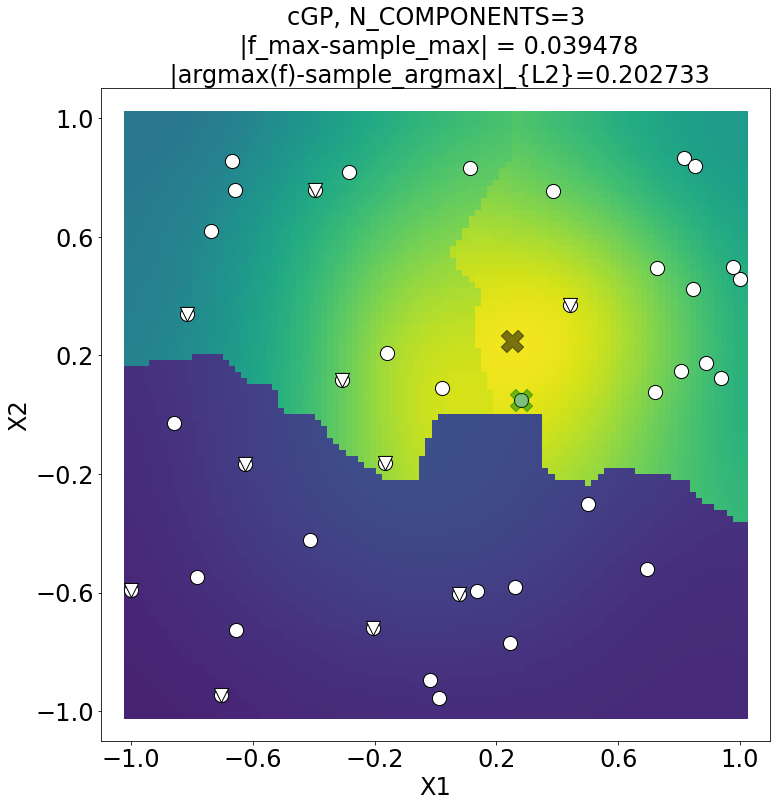} & \includegraphics[width=5cm]{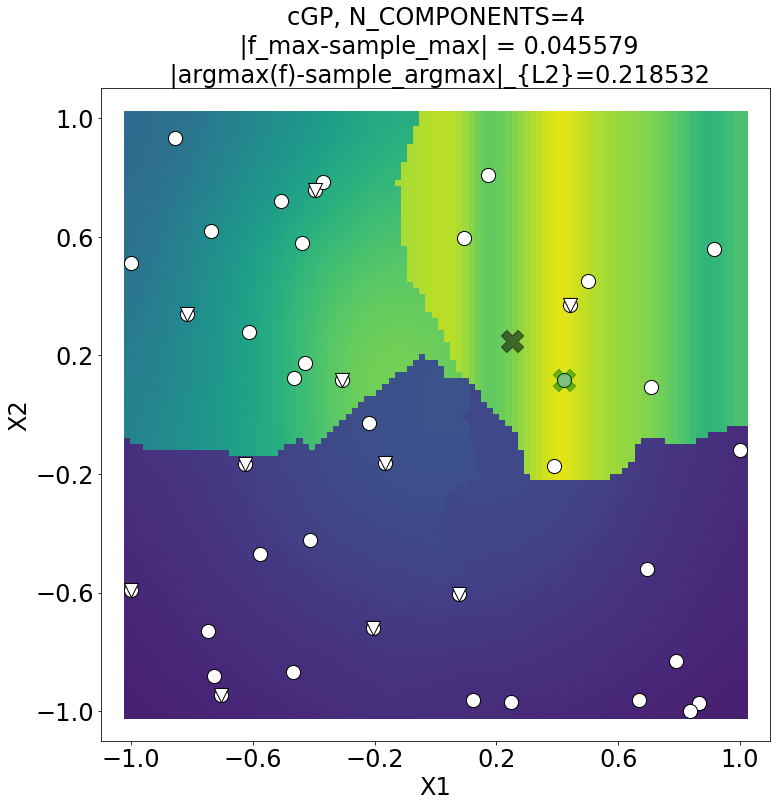}\tabularnewline
\midrule 
DGM ($k=2$, fits 2) & DGM ($k=3$, fits 2) & DGM ($k=4$, fits 4)\tabularnewline
\midrule 
\includegraphics[width=5cm]{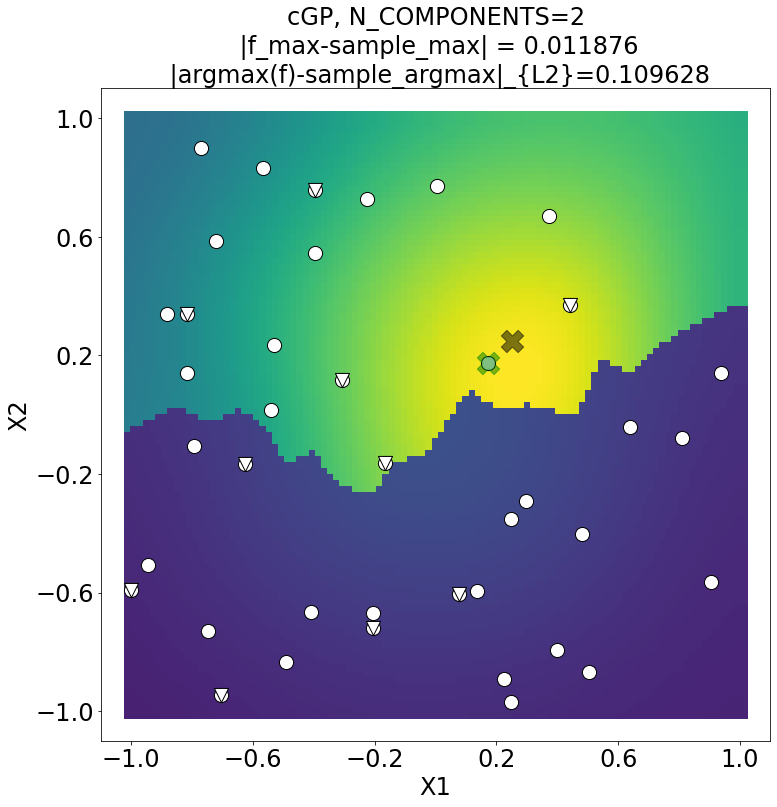} & \includegraphics[width=5cm]{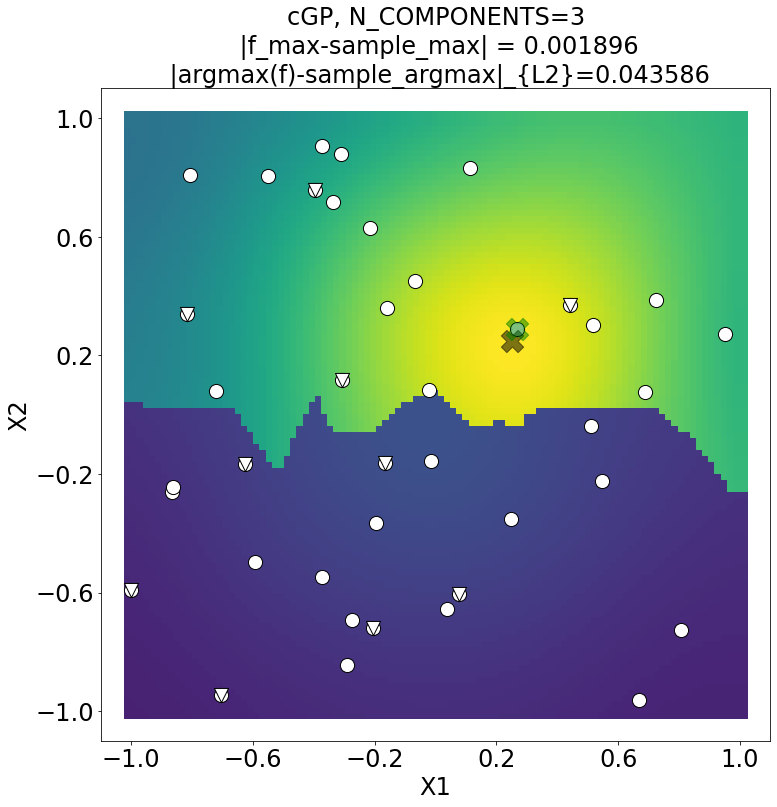} & \includegraphics[width=5cm]{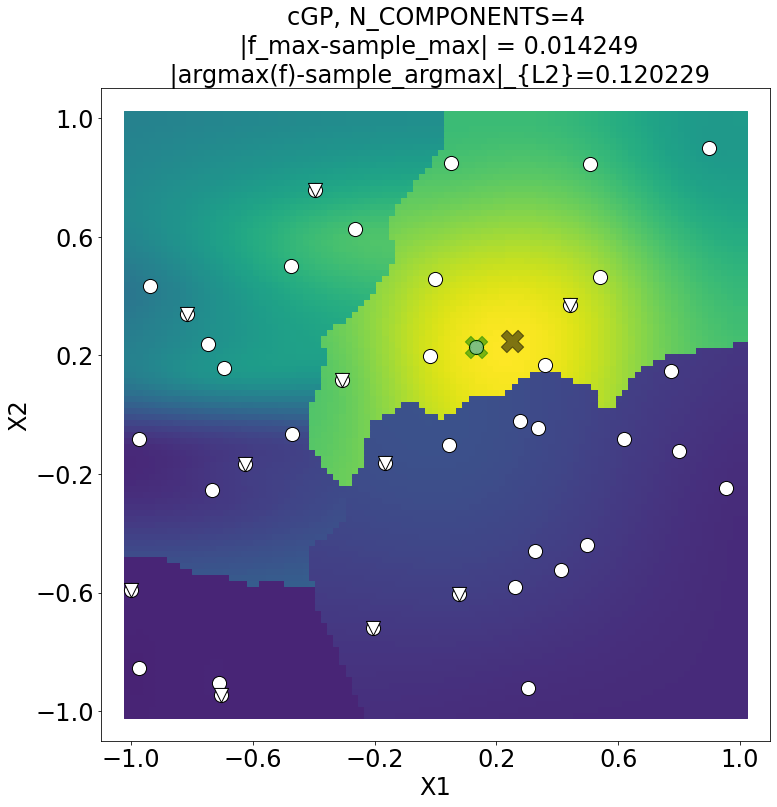}\tabularnewline
\bottomrule
\end{tabular}

\end{adjustbox}

\caption{\label{fig:2d non-smooth black-box function compare} Mean function
and partition scheme extracted from the GP surrogate model evaluated
on $100\times100$ equally spaced grids on $[-1,1]^{2}$ with weighted
expected improvement acquisition function and $n=10$ random pilot
samples (triangles), $n=30$ sequential samples (circles) from $f_4$,
We provide the cGP with oracle partition ($x_2=0$) in cGP (partitioned) panel. 
We use different kinds of clustering methods: $k$-means (with
$k=2,3,4$), Dirichlet Gaussian Mixture (DGM, the algorithm fits the surrogate with $2,3,4$ clusters when the maximal allowed number of components are 2,3,4, respectively.) We show exact minimal point by a black cross; sample minimal point by a green cross.%
}
\end{figure}
\FloatBarrier

Furthermore, the choice of clustering algorithm in the cluster-classify step matters. When $k=2$ the fit of cGP with $k$-means clustering
seems to capture the non-smoothness well while retaining the behavior that is similar to the underlying function.
However, in practice, we do not know how many clusters are there.
One approach is to treat $k$ as a hyper-parameter that is subject
to further tuning with model selection criteria or cross-validation
on real data. The other approach, which we take here, is to use Dirichlet Gaussian mixture. 
We show the fit of these latter two approaches in Figure \ref{fig:2d non-smooth black-box function compare}
with $k=2,3,4$. 
Figure \ref{fig:2d non-smooth black-box function compare} serves as a
qualitative verification of our method in a low-dimensional
domain ($d=2$). Usually, a reasonably large $k$ (e.g., 3,4) and DGM would be a practical choice for cluster-classify step whenever our sampling budget allows.

The illustrative examples above show how cGP behaves in terms of mean
function and error statistics. cGP behaves differently as we change
the number $k$ of components and clustering algorithms. Note that since we use expected improvement (EI) as our guiding acquisition function throughout the paper, it is expected that the $\|f-g\|_{\max}$ would accumulate from clusters, hence possibly worse than the GP. 

\begin{figure}
\begin{adjustbox}{center}

\textbf{\includegraphics[width=7cm]{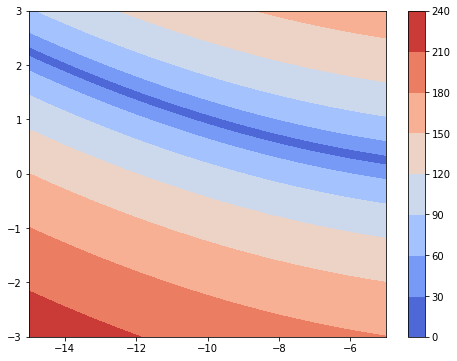}\includegraphics[width=7cm]{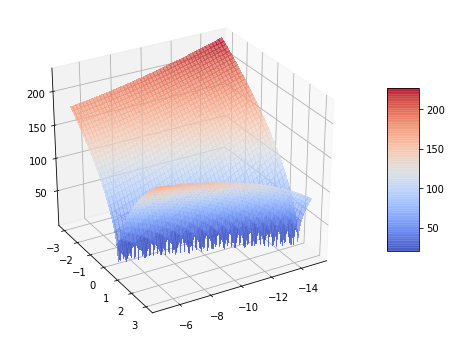}}

\end{adjustbox}

\caption{\label{fig:The-Bukin-N.6}The Bukin N.6 function is defined by
$f(x)=100\sqrt{|x_{2}-0.01x_{1}^{2}|}+0.01|x_{1}+10|$. The function
has unique global minimum $f(x_{\ast})=0$ at $x_{\ast}=(-10,1)$.
We optimize the function to find its minimum over the domain of $(-15,5)\times(-3,3)$. The plot is based on evaluation of $f(x)$ on a 100 by 100 grid.}
\end{figure}

Next, we want to see how the sequential
sample size $n$ and exploration rate $\tau$ could change the behavior
of the cGP model when compared to the GP model. As a third example,
we study a well-known benchmark function called Bukin N.6 function
(displayed in Figure \ref{fig:The-Bukin-N.6}) defined on a 2-dimensional
domain. 

We experiment with cGP surrogates with exploration rates $1$ (always sample
based on acquisition maximization), $0.8$ and $0.5$. When the exploration
rate is 0, the sequential samples are all randomly chosen; while when
the exploration rate is 1, the sequential sampling is fully based on
acquisition maximization. In some situations (like matmul in Sec.
\ref{subsec: Tuning}), adjusting the exploration rate would allow us
to explore the domain more efficiently. 

\begin{figure}
\begin{adjustbox}{center}

\begin{tabular}{ccc}
\toprule 
$k=2$ & $k=3$ & $k=4$\tabularnewline
\midrule 
\includegraphics[width=5cm,page=1]{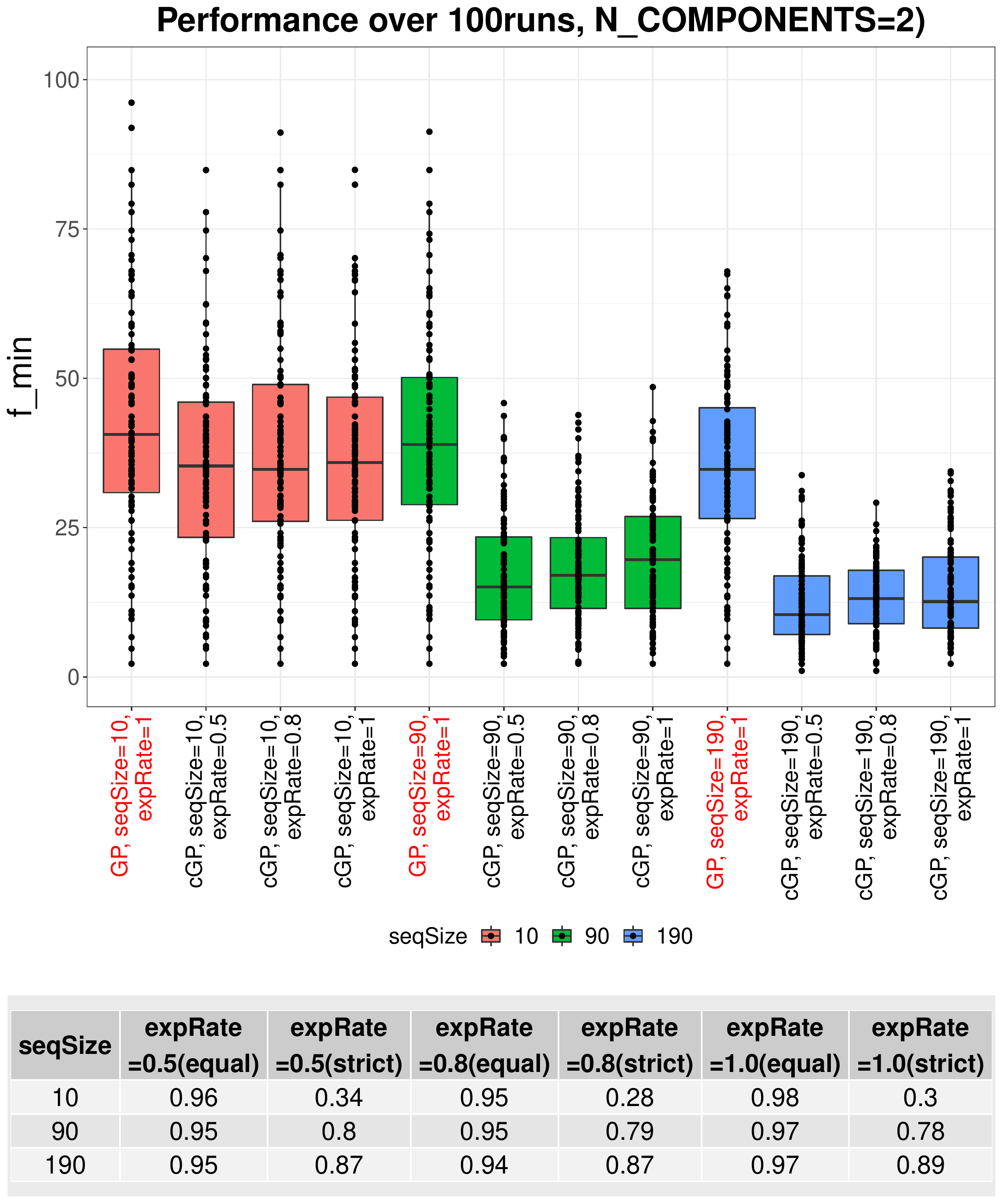} & 
\includegraphics[width=5cm,page=2]{Figures/BUKIN_N6_f_min_analysis_BOX} & 
\includegraphics[width=5cm,page=3]{Figures/BUKIN_N6_f_min_analysis_BOX}\tabularnewline
\midrule
\includegraphics[width=5cm]{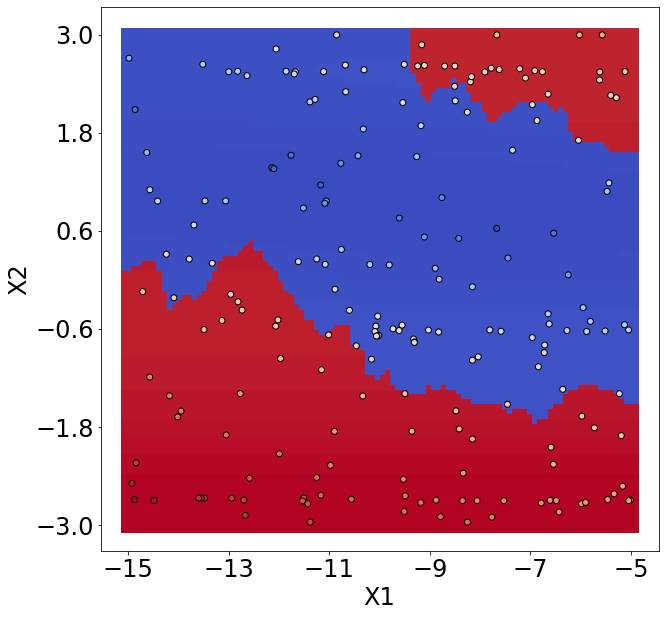} & \includegraphics[width=5cm]{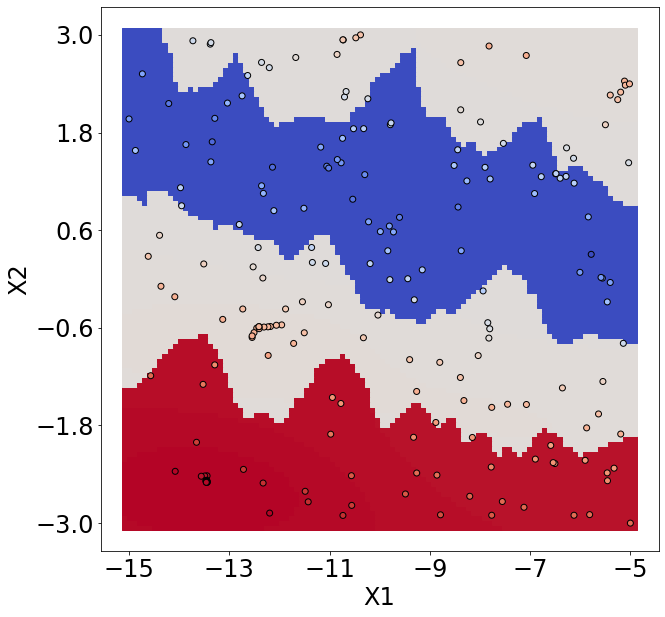} & \includegraphics[width=5cm]{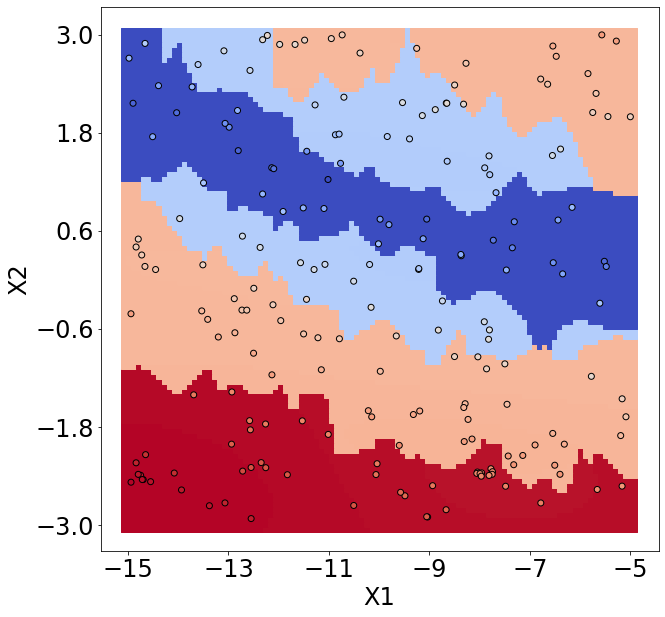}\tabularnewline
\bottomrule
\end{tabular}   

\end{adjustbox}

\caption{\label{fig:The-boxplot-of-Bukin N.6}The box plot of optima (minimum)
obtained from the cGP surrogate model with exploration rates (expRate) of 0.5, 0.8
and 1.0, 10 pilot samples and a sequential sample of size (seqSize) $n=10,90,190$
from Bukin N.6 function. Each dot on the box plot means one optimum
obtained from one random seed. We display the performance in terms
of distribution of optima from sequential samples from GP and cGP
surrogate models, with different numbers of components, exploration
rates and sequential sample sizes. \protect \\
In the percentage table, ``(equal)'' columns show the percentage
of cGP fitted surrogates with optima \textbf{equal or better} than
the baseline GP surrogate under different exploration rates in 100
different random seeds. ``(strict)'' columns shows the percentage
of cGP fitted surrogates with optima \textbf{strictly better} than
the baseline GP surrogate under different exploration rates in 100
different random seeds.\protect \\
We also provide corresponding partition schemes (different clustering
components are indicated by different colors) for one fit of cGP surrogate
model with 190 sequential samples. 
}
\end{figure}

For now, let us see how different
exploration rates would affect the performance of cGP surrogate models.
In this set of experiments, we first investigate the distribution
of optima found by different surrogate models. In the first row of
Figure \ref{fig:The-boxplot-of-Bukin N.6}, we repeat the optimization
for 100 different random seeds and we can see that when there are
moderate or sufficient sequential samples ($n=90,190$), cGP surrogates
produce $f_{\min}$'s that are lower (hence better) than the optima
given by the GP surrogate with the same sample size. When there are
only limited samples ($n=10$), cGP seems to have small improvements. In each of the box plots, we can see that this comparison
is systematic and not due to randomness. It is not hard to see in
the second row of Figure \ref{fig:The-boxplot-of-Bukin N.6} that,
partition schemes are actually capturing the level sets of Bukin N.6
function rather accurately and sample more near the optima. Quantitatively,
we provide the percentages that cGP surrogates (under different combination
of exploration rates) outperform their GP counterparts among 100 runs
with distinct random seeds. We not only provide the percentages that
cGP gives equal or strictly smaller $f_{\min}$, but also the percentage
that cGP gives strictly smaller $f_{\min}$. The percentage that cGP
gives equal or strictly smaller $f_{\min}$ should be over 50\% to
justify the use of cGP, and a larger percentage of strictly better
optima indicates the superiority of cGP.

This experiment shows a consistent and qualitative advantage of
cGP over GP surrogates. However, more importantly, it also shows that different
configurations of hyper-parameters in the cGP model can be adjusted to
yield better performance over the same function. The partition scheme
accompanying the cGP surrogate fit also reveals the nontrivial geometric
structure of the black-box function. Empirically, we find cGP
fits slightly faster compared to GP when $n=190$, since inverting
$k$ smaller covariance matrices costs less than 1 large covariance
matrix of GP. 
In general, if the black-box function has optima in
a relatively small region compared to the whole domain or the function
has non-smoothness (also sharp changes) near optima, then the cGP surrogate
may reach better optima compared to the GP surrogate, and run faster.
And cGP can identify different kinds of behavior rather faithfully through
the cluster-classify procedure. Even if the cGP fails to capture the different behaviors 
in the domain, we still observe that cGP has competitive performance
when compared to GP (which is consistent with Figures \ref{fig:2d smooth black-box function compare}
and \ref{fig:2d non-smooth black-box function compare}).

\subsubsection{Summary}
We explore the effect of hyper-parameters of the cGP surrogate model in
this section. With qualitative distributional summaries of experiments
like box plot and quantitative analysis introduced, we present evidence
that cGP has advantages over classical GP surrogates:
\begin{enumerate}
\item In the smooth function example (Figure \ref{fig:2d smooth black-box function compare}),
we observe that cGP with small $k$ performs similarly to GP.
When there is no non-smoothness in the black-box function, we can
use cGP in place of GP without much loss of performance. In fact,
GP is a special case of cGP with $k=1$.
\item In the non-smooth function example (Figure \ref{fig:2d non-smooth black-box function compare}),
we observe that cGP with different $k$'s can outperform GP when
$k$ is appropriately chosen. When there is non-smoothness in the
black-box function, we can use cGP to improve optimization results
and adjust hyper-parameters. cGP would also produce a partition scheme
that illustrates the non-smooth behavior in the black-box function.
\item In the Bukin N.6 function example (Figure \ref{fig:The-Bukin-N.6}),
we observe that cGP with different exploration rates and sample
sizes outperforms GP under the same conditions, but different configurations
of these hyper-parameters can affect the optimization  performance of cGP. In
this example, cGP produces a partition
scheme illustrating a non-trivial structure of the black-box function
and yields better optima.
\end{enumerate}
cGP has the potential of detecting the non-smoothness in the black-box
function and cGP usually performs at least as well as GP. We include
further synthetic study evidence in Appendix \ref{sec:Additional-Simulation-Studies}
to support this claim. In the next section, we investigate the
performance of the cGP surrogate model in real tuning problems. 

\FloatBarrier

\subsection{\label{subsec: Tuning}Tuning Problems}

As we explained in our cGP model specification above, there are quite a
few (hyper-)parameters 
we could select for the cGP surrogate model.

In a general application, the default choice would suffice. We run
some pilot experiments and choose the model parameters like the maximal
number of components and exploration rate sequentially. Although the
most rigorous way is to find a grid of all possible combinations of
these parameters and conduct an exhaustive search, empirical choices
would also help. For example, we could  guess how many components
are there in the black-box functions from the observed data.

The first example is a classic low-dimensional tuning problem, that
is to find an appropriate blocking to improve the performance of
a large matrix multiplication. We focus on the challenge of limited
sample sizes. The second and the third examples, on the other hand, have rather high-dimensional
parameter spaces, and hence the difficulty to fit a surrogate model
in a high-dimensional parameter space will be present. One of them uses the piston cycling function, the other involves the tuning of the sparse matrix factorization time in the SuperLU\_DIST package \citep{yamazaki_new_2012}.

\subsubsection{\label{sec:matmul}Low-dimensional tuning}

Let us revisit the matrix multiplication tuning problem we mentioned
in subsection \ref{subsec:Problem-and-Challenges}. The black-box
function is the computational speed function of the operation $A*B=C$
with blocking structure, where the $n \times n$ matrices $A,B,C$ have 
double precision entries,
and consist of $N\times N$ blocks each of size $b\times b$ such that $n=Nb$.
The parameter we attempt to tune is the shared block size $b$ and
the computational speed function $f(b)$. Therefore, we have six nested loops
(3 outer loops to perform $N^3$ multiplications of $b \times b$ matrices, and 3  inner loops to perform 1 $b \times b$ matrix multiplication).
Let $m$ be the amount of memory traffic (i.e., slow communication)
between main memory and cache, and $f$ be the number of
floating point operations. In this example, we have the following analytical performance model:
\begin{align*}
m= & N\cdot n^{2}+N\cdot n^{2}+2n^{2} & \text{ read each block of }A,B\text{ }N^{3}\text{ times},N^{3}\cdot b^{2}=N\cdot n^{2}\\
= & (2N+2)n^{2} & \text{ read and write each block of }C\text{ }1\text{ time},2N^{2}\cdot b^{2}=2n^{2}\\
f= & n^{3}+n^{3} & \text{number of additions,}N^{3}\cdot b^{3}=n^{3}\\
= & 2n^{3} & \text{and number of multiplications, \ensuremath{N^{3}\cdot b^{3}}=\ensuremath{n^{3}}}
\end{align*}
The quotient $q = f/m$ of these two quantities, known as computational
intensity, can be used to model the computational speed of the matrix
multiplication. It can be calculated as $q=f/m=\frac{2n^{3}}{(2N+2)n^{2}}\approx b\text{ as }n\rightarrow\infty,\text{s.t. }n=Nb$.
Therefore, based on this analytical performance model, we can optimize
(maximize) the function $q(b)$ by increasing the block size $b$ (denoted as
$\bm{x}$ below). However, this analytical model has an implicit assumption:
the 3 $b \times b$ blocks always fit into the cache. We want to
increase the
block size to reduce communication but also avoid overflowing the
cache, which would cause a sudden decrease in performance. \citet{hong_complexity_1981} proved that the minimum number of
words moved between cache of size $M$ and memory is
$\Omega (\frac{n^{3}}{\sqrt{M}})$ which corresponds to $q \approx b = O(\sqrt{M})$. The exact optimal value of $b$ depends on details of
how the loops are organized, compiler optimizations, and hardware architecture, which has led to a large literature on the topic,
including the development of special purpose autotuners. 
More details of optimization for matrix multiplication are explained
by \citet{bilmes1997optimizing}, \citet{whaley2001automated} and \citet{zee2016blis}.
Our goal is not to compete with these autotuners, which may include
custom code generation as well as tuning parameters, but to use 
matrix multiplication as a motivating example, where clear non-smooth and different regimes can be observed, for the clustered GP surrogate.

\begin{figure}[th]
\begin{center}%
\includegraphics[height=7cm]{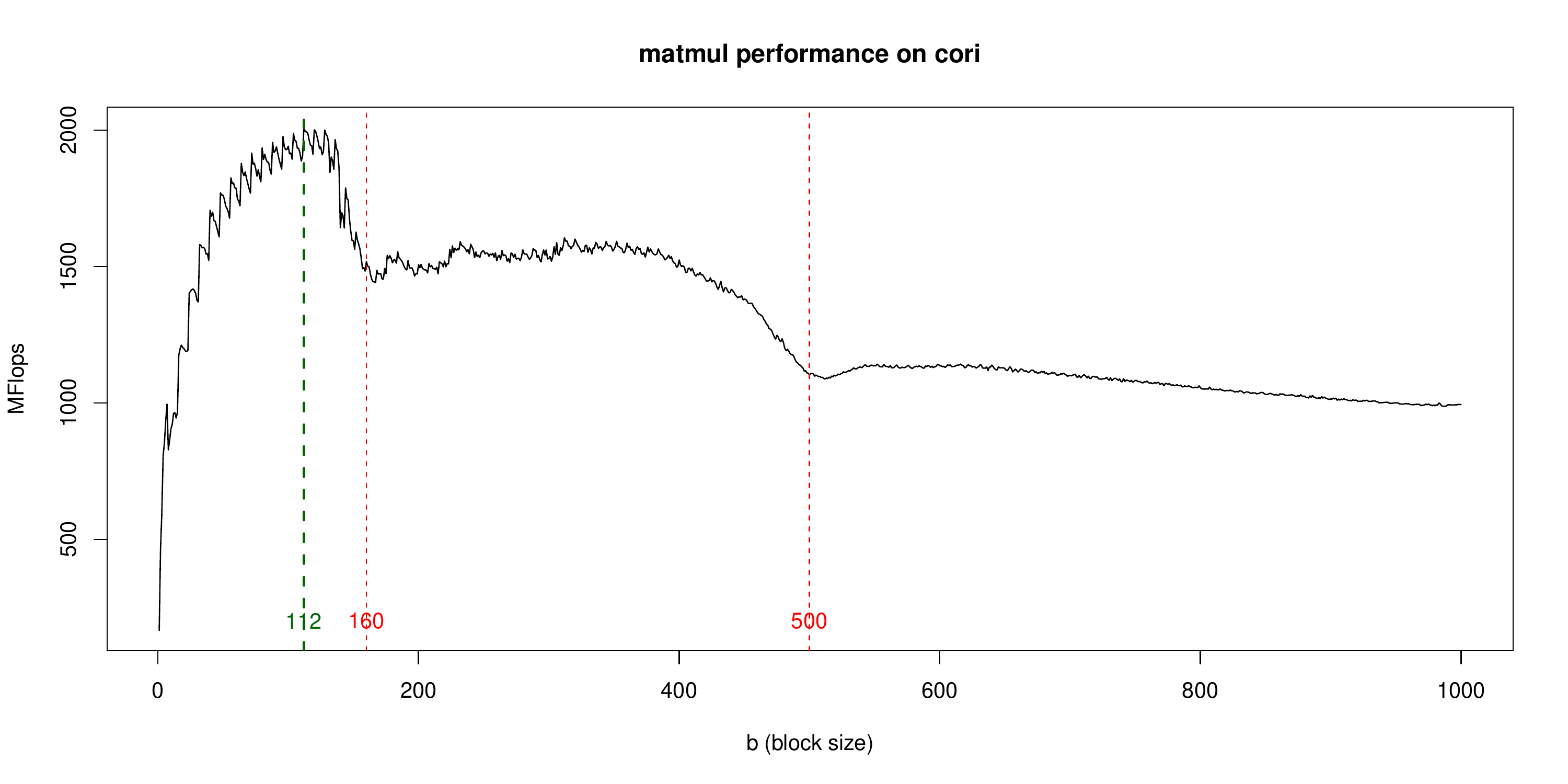}
\end{center} \caption{\label{fig: matmul-multiple8_block} The computational speed for
matrices of size $n=1000$. This is the true black-block function obtained from a Haswell node.
We also use a green dashed line to indicate the actual maximum 
$(\bm{x}_{\text{max}},f_{\max})=(112,2010.702)$; and red dashed lines for the regime cutoff $\bm{x}=b=160,500$ discussed in the text.}
\end{figure}

The machine on which we conduct this experiment is a
Cori Haswell computing node, with an L1 cache
size of 32KB, 256KB for L2 cache and 40960KB for L3 cache. All cache-line sizes are 64 bytes, i.e., 8 double precision words. Each node has an Intel Xeon Processor E5-2698 v3 2.3 GHz processor (with a theoretical peak of 2.81 PFlops) with a 128GB memory. 

\begin{figure}[ht!]
\begin{center}%
\begin{tabular}{cc}
\toprule 
(a) GP, 10 sequential samples. & (b) GP, 90 sequential samples.\tabularnewline
\midrule
\midrule 
\includegraphics[height=6cm]{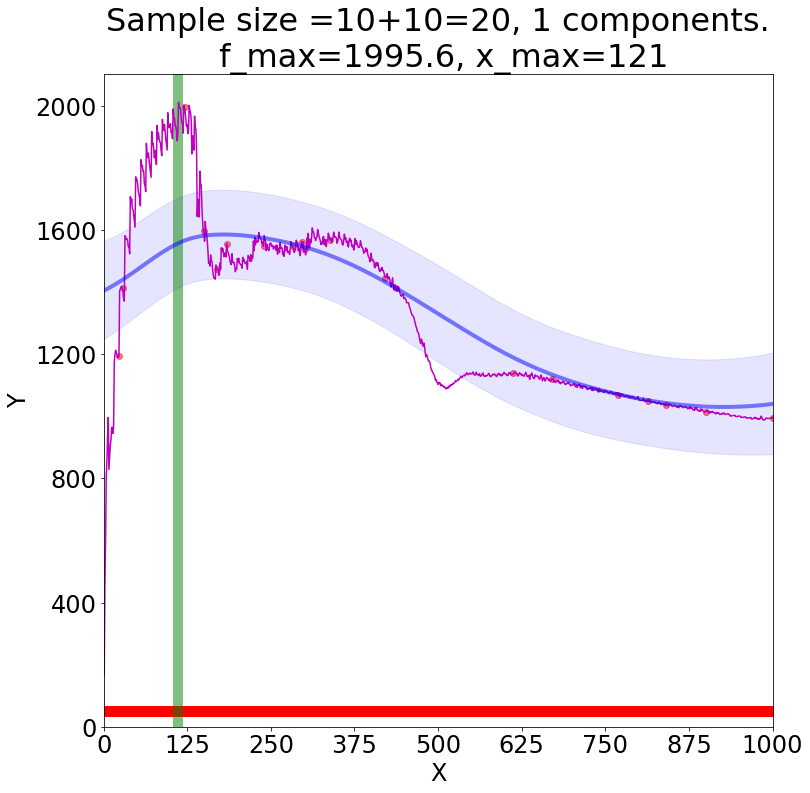} & \includegraphics[height=6cm]{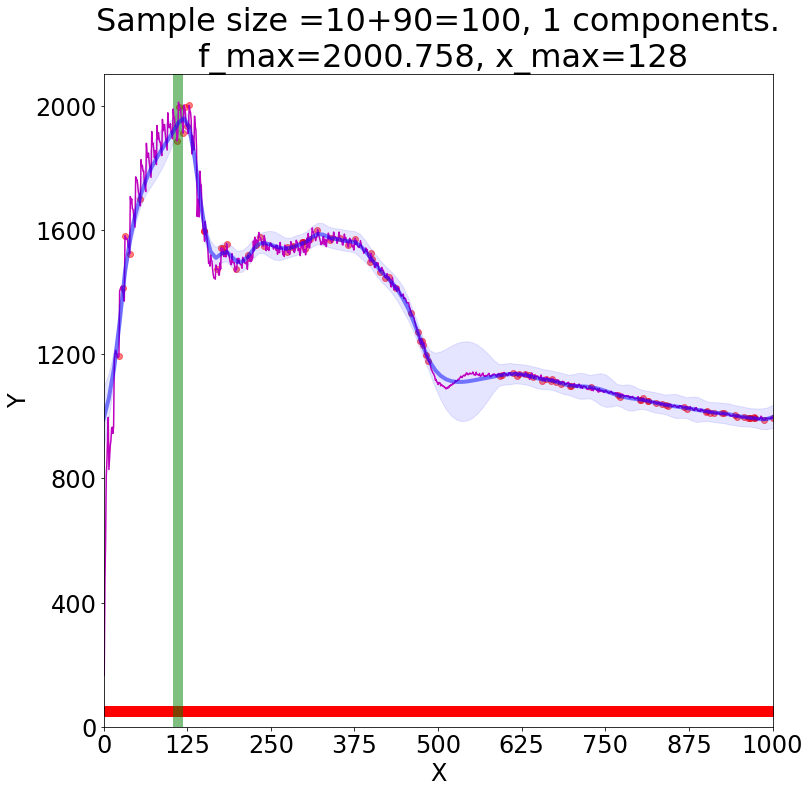}\tabularnewline
\midrule 
(c)  cGP, 10 sequential
samples. & (d) cGP, 90 sequential samples.\tabularnewline
\midrule
\midrule 
\includegraphics[height=6cm]{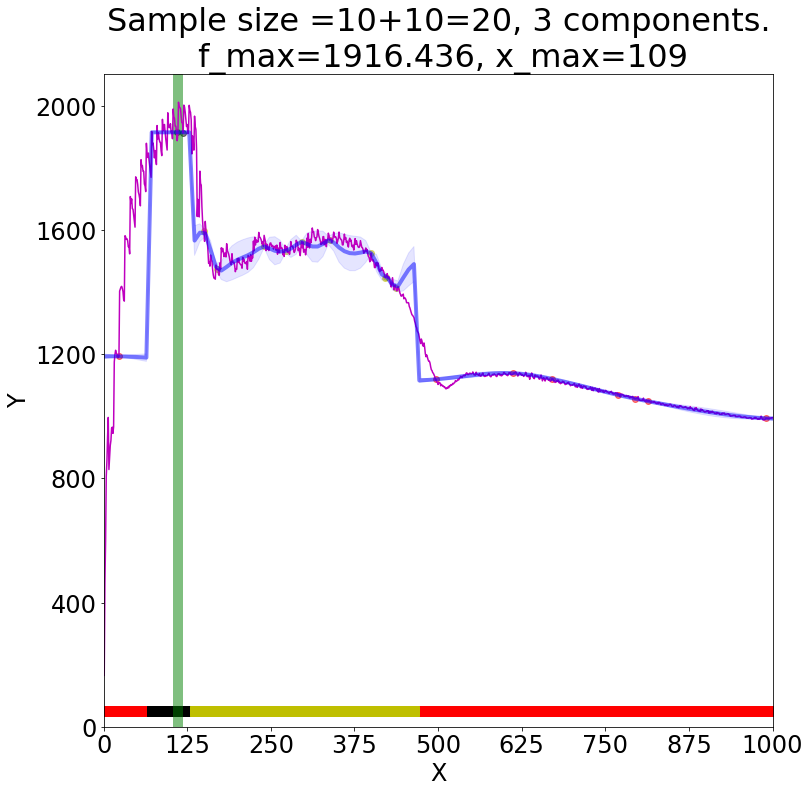} & \includegraphics[height=6cm]{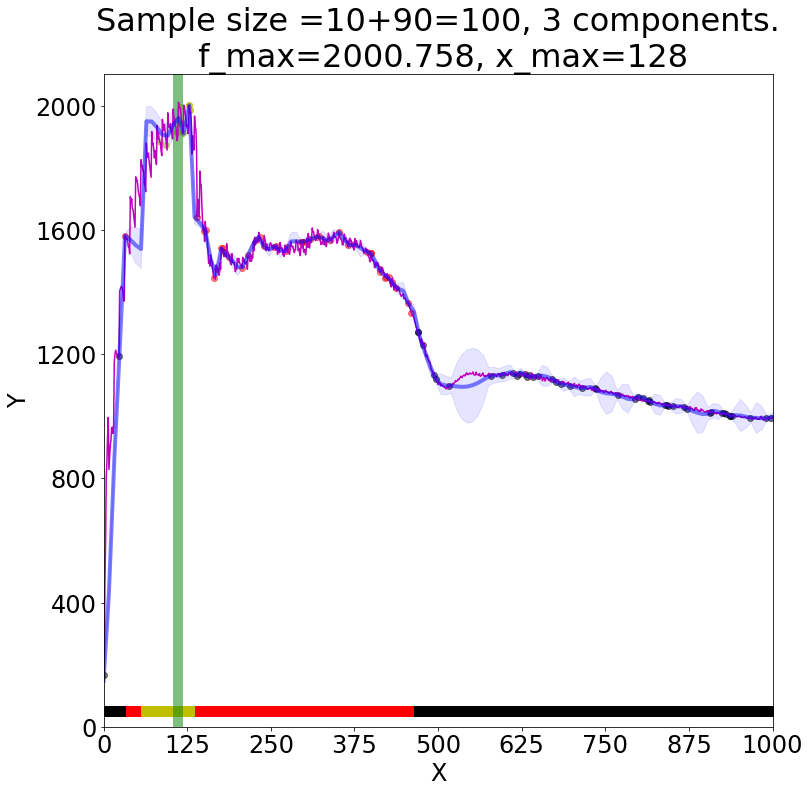}\tabularnewline
\bottomrule
\end{tabular}\end{center} \caption{\label{fig: matmul} The computational speed function and fitted surrogate models. The dark blue solid line and light blue shaded area are the mean and variance of fitted prediction models.  $(\bm{x}_{\text{max}},f_{\max})$ records the optimal block size and the actual optimal speed %
is illustrated by green vertical line on the figure. 
Bottom colored bars with different colors indicate to which
cluster certain portions of the domain belong (but color has no semantic meaning). 
We also use magenta solid lines to overlay the truth from Figure \ref{fig: matmul-multiple8_block}
for comparison purposes. (The exploration rates of these experiments are all 1.) 
}
\end{figure}

Large matrix multiplication is quite expensive, and if we evaluate
the computational speed of each block size $b$ from 1 to $n=1000$ (when the block
size is 1 or 1000,
there is no blocking). 
The L3 cache is large enough to fit all 3 $1000 \times 1000$ matrices $A$, $B$ and $C$.
Based on 
multiple runs, we observe that there is little noise in the data and do not
consider averaging it further here.
In this dataset, we can make preliminary observations that  there are several different kinds of behavior, apart from the noise:

\begin{itemize}
\item When $b<160$, the speed increases rapidly with some oscillations and then drops to a plateau level. The oscillations, with local peaks at $b$ equal to multiples of 8, 
are to be expected since the cache-line size is 8 words.
In practice, one might limit the search space to $b$ being multiples of 8,
resulting in a smaller search space and a smoother 
function to optimize, but 
here we consider all values of $b$ to more rigorously evaluate the cGP. 
The optimal configuration is attained at  $(\bm{x}_{\text{max}},f_{\max})=(112,2010.702)$, with the 2 second best
 configurations at $(120,2001.35)$ and $(128,2000.758)$.
 The choice of $b$ that minimizes memory traffic
 and so (potentially)
 maximizes speed, depends on details like the ordering of the loops,
 and what the cache chooses to keep inside or not. In particular, anywhere
 from just over 1 $b \times b$ to 3 $b \times b$ blocks may need to fit in 
 cache. In our case, since $A$, $B$ and $C$ all fit in L3 cache, the
 question is what $b$ minimizes the traffic between L2 and L3. Since the L2
 holds 32K words, this means it can hold at most 1 square block of size
 $b=181$, 2 of size $b=128$, and 3 of size $b=104$, 
 ignoring cache conflicts, etc. 
 So it is no surprise that the performance
 falls off rapidly past $b=128$, reaching a plateau around $b=160$.
For a regular sampling scheme, it requires some more sampling in this interval to attain this point, as seen from Figure \ref{fig: matmul} (a) and (d), while cGP benefits from a good partitioning.

\item When $160\leq b\leq 500$, the computational speed has a stable trend and rough behavior,  
with some fluctuations caused by memory alignment and code implementation.
Again, most of the small peaks occur at multiples of 8.
\item When $b\geq 500$, there is another significant drop (the trend can also be identified starting from $b=400$) until a roughly constant computational speed is attained.

\end{itemize}

We choose only 20 to 100 samples (including 10 pilot samples) when fitting surrogates so that our
surrogate model with sequential sampling can be completed within 20
minutes (versus 3-4 hours needed for an exhaustive search) on the same machine configuration. 
However, to reduce randomness,
we use the recorded dataset as the black-box function $f$ defined
on $[1,1000]\cap\mathbb{Z}$.
In Figure \ref{fig: matmul}, we show
the cGP with Dirichlet Bayesian Mixture clustering with maximal number
of components as $3$ (which is also the actual number
of components chosen), a $k$-nearest neighbor classifier
($k=3$), and a Mat\'ern 3/2 covariance kernel (with nugget). 
We start with the same
10 randomly chosen pilot samples and then cap the
total sample size at 20 and 100, respectively. 

We can get a first impression of the difference between GP and cGP surrogates by looking at Figure \ref{fig: matmul}. When the sample size is 20, we can see that there is a big difference
between the surrogate fits of GP and cGP, although their sample maxima are close.
This difference is due to
the fact that the clustering algorithm we choose gives us a correct cluster-based partition when there are as few samples as 20. When the sample size is 100, we observe that the cGP identifies the drop starting from $b=160$ and $b=500$ and attains maxima as good as $b=128$ (with $f(128)=2000.758$, close to the optimal $f(112)=2010.702$), and the same as GP. 
While its performance is similar to GP with larger sample sizes, we can see that cGP correctly identifies
the change of function behavior caused by the drop of computational speed near $b=160$ and
another behavior that changes rapidly when $b\geq500$ even with very limited samples. Within
each partition regime, the cGP provides a better fit in the  sense that its mean function captures the sharp slopes caused by cache size. The main point we try to make here is that cGP can provide additional partition information while still maintaining a similar performance as GP in searching for maximum. This observation is further strengthened by the summary of repeated experiments in Figure \ref{fig: matmul-LR-1_block_box}.

\begin{figure}[!ht]
\begin{adjustbox}{center}%
\begin{tabular}{ccc}
\toprule 
(a) & (b) & (c)\tabularnewline
\midrule 
Exploration rate = 1.0 & Exploration rate = 0.8 & Exploration rate = 0.5\tabularnewline
\midrule
\midrule 
\includegraphics[height=5.5cm]{Figures/fig_partition_d_10.png} & \includegraphics[height=5.5cm]{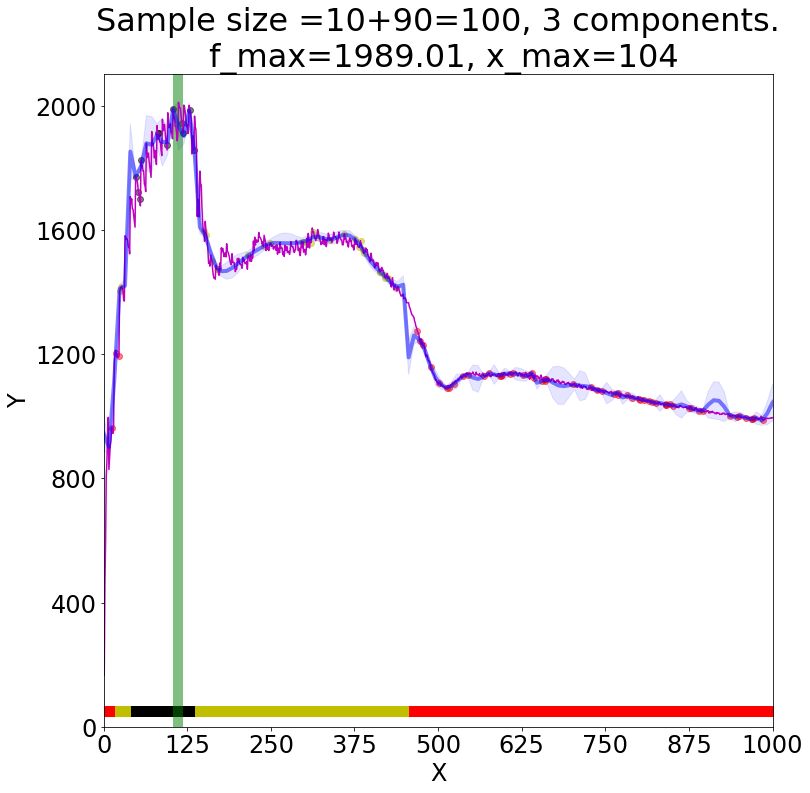} & \includegraphics[height=5.5cm]{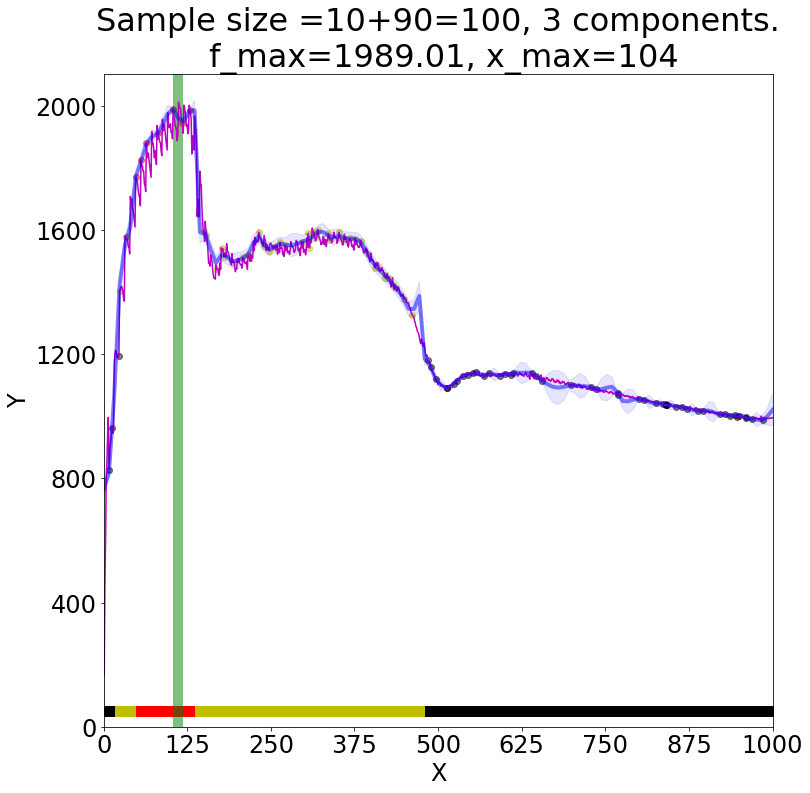}\tabularnewline
\bottomrule
\end{tabular}\end{adjustbox} \caption{\label{fig: matmul-LR_block_exp} The computational speed function and fitted
surrogate models. Bottom colored bars with different colors indicate
to which component a certain portion of the domain belongs (but color has no semantic meaning). 
We also use magenta solid lines to overlay the truth from Figure \ref{fig: matmul-multiple8_block}%
for comparison purposes. The fit from GP surrogate model with 60 sequential
samples and exploration rate (a) $1$. (b) $0.8$. (c) $0.5$.} %
\end{figure}

So far, we consider all sequential samples selected by acquisition maximization (exploration rate 1). In Figure \ref{fig: matmul-LR_block_exp}, we show how the cGP fit changes with
different exploration rates, which are defined in the flowchart (see
Figure \ref{fig:The-algorithm-flowchart}) and affects the random
sampling procedure. 
If the exploration rate is 0.6, then with a probability
0.6 we choose the next sequential sample with acquisition maximization;
and with probability 0.4, we randomly choose the next sample from the
input domain (i.e., a random block size $b$). 

It is not difficult to observe from Figure \ref{fig: matmul-LR_block_exp}
that, with a lower exploration rate, the partition components remain similar,
where for all exploration rates 
the drops at $b=160$ and $b=500$ are clearly indicated by the change of additive components.
However, with a lower exploration rate, we also observe that the sample
locations are more spread over the input domain.

\begin{figure}[ht!]
\begin{adjustbox}{center}%
\begin{tabular}{c}
\toprule 
\includegraphics[height=7cm]{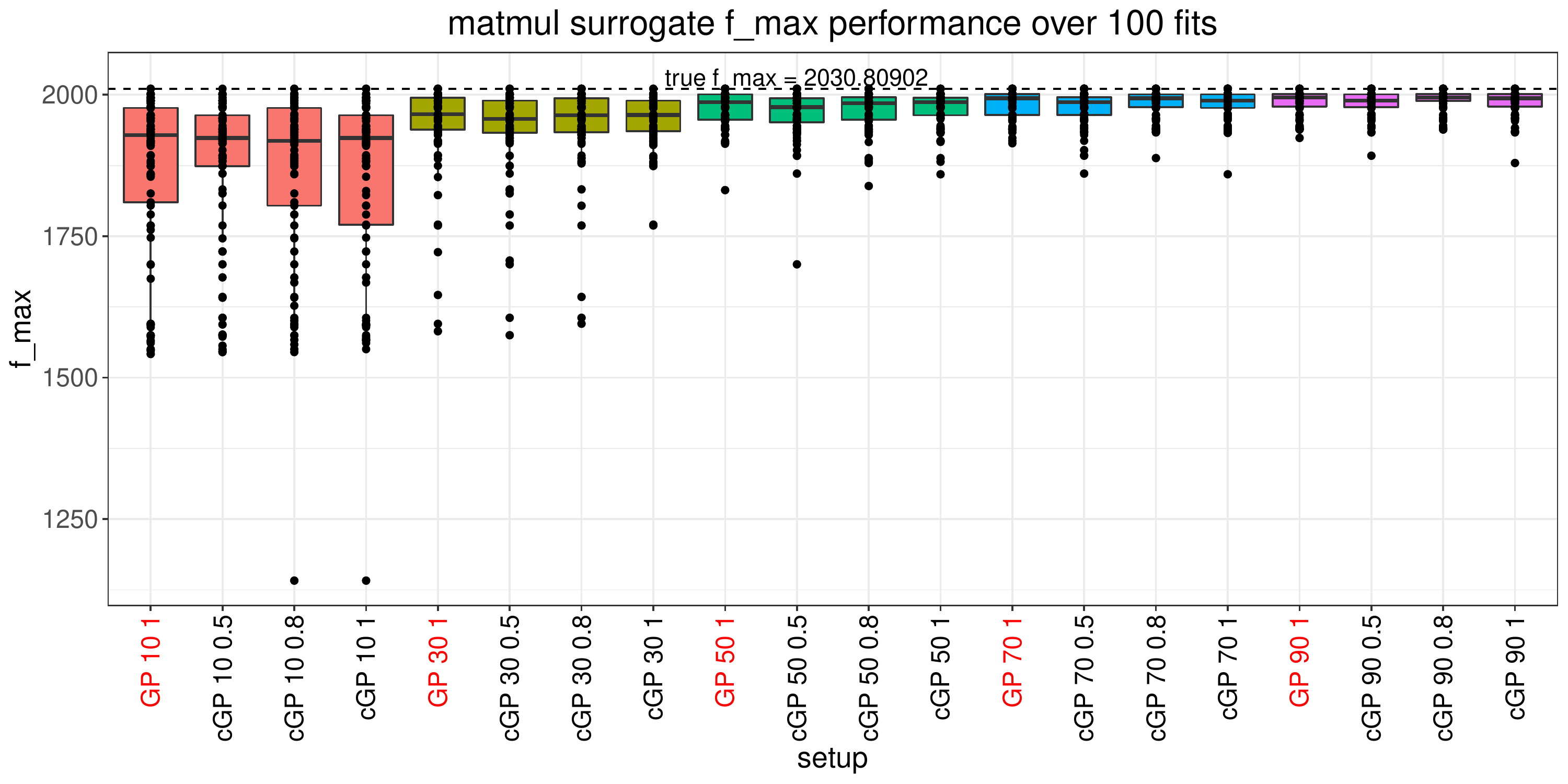} \tabularnewline
\includegraphics[height=7cm]{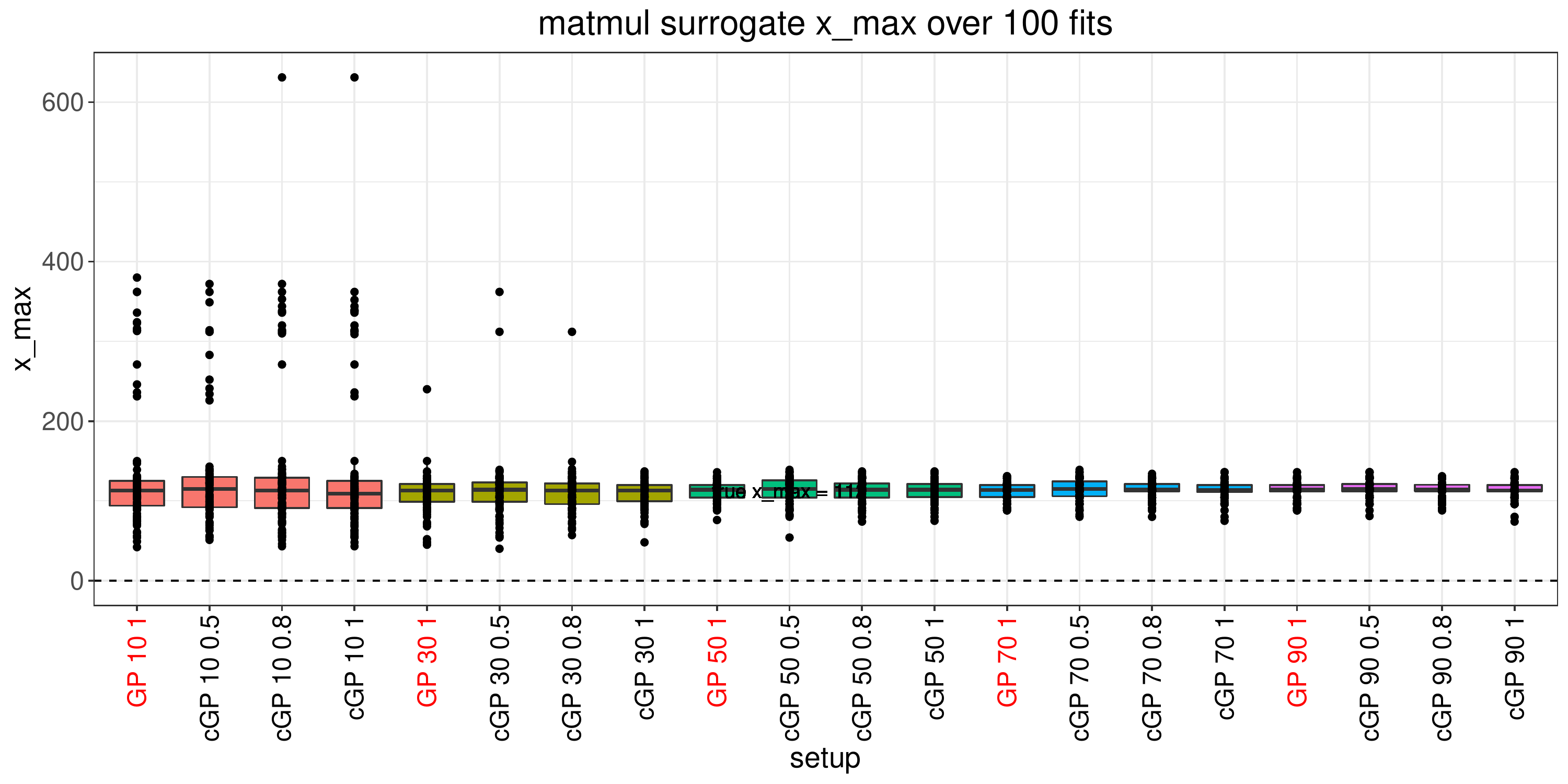}\tabularnewline
\bottomrule
\end{tabular}\end{adjustbox} \caption{\label{fig: matmul-LR-1_block_box} The box plot of $f_{\max}$ (top) and $\bm{x}_{\max}$ (bottom, log scale) elicited from sequential samples of 100 fitted surrogate
models, each dot in the box plot represents the optimal point in each
run (hence each box plot contains 100 points). The sequential sample
size varies from 10 to 190 with 10 pilot samples. The exploration
rates of the cGP model are chosen to be 0.5, 0.8 and 1 (x-axis format: model/sequential sample size/exploration rate). \protect \\
Results for simple GP surrogates are highlighted with red labels as
a baseline surrogate model; while cGP surrogates with different exploration
rates are highlighted with black labels.}
\end{figure}

In Table \ref{tab: matmul-LR-2block}, we also provide the percentage (among
100 runs with different shared random seeds) of runs that cGP outperforms
simple GP surrogate models, as a quantitative supplement to the illustration  in Figure \ref{fig: matmul-LR-1_block_box}, which 
supports the claim that our cGP surrogate models with the setups above (and different exploration
rates) would consistently behave similarly like a simple GP surrogate model, while providing the non-smooth information of the black-box function in terms of both partitions and number of components. As the sequential sample size increases from 10 to 90 
(see Table \ref{tab: matmul-LR-2block} (a)), we can see that both surrogate models
behave similarly in terms of reaching optima, but the cGP model reaches the true optimum $b=112$
of the underlying black-box function more often (see Table \ref{tab: matmul-LR-2block}
(b)). A medium exploration rate of 0.8 in the cGP model seems to improve the performance
in this specific data application. Eventually, the number of additive
components in the fitted cGP model stabilizes at around 2.9 (see Table
\ref{tab: matmul-LR-2block} (c)), which is consistent with our observation
that there are 3 partition regimes for the full recorded dataset. 
The average
number of components is not exactly 3 since the $[1,80]\cap\mathbb{Z}$ partition component of the domain is usually under-sampled in surrogates with small sequential sample sizes.

In this matrix multiplication tuning example, the cause of non-smooth
points (i.e., reduction of communication; overflow of fast cache)
and different kinds of behavior is clear and observed in the recorded dataset as
shown in Figure \ref{fig: matmul-multiple8_block}. cGP surrogates behave
similarly to the GP model when there are few samples; but when there
are enough samples, the cGP model clearly identifies different partition regimes much
better with a reasonable performance in optimization. It is also clear that
each additive component has a different normalization for the data
within this label, hence different degrees of  smoothness
are better elicited in the surrogate fit. The non-smooth change-points
between components are also captured in the cGP model. Therefore, the cGP model searches the optimum $f_{\max}$ and $\bm{x}_{\max}$ more
efficiently compared to a simple GP surrogate. In short, cGP provides non-smooth information about the black-box function without loss of optimization performance compared to GP in this application. 

In a different setting of matmul application, where we do not change our blocking strategy but vary the matrix size, we even witness the accuracy improvement by adopting the cGP as  our surrogate model, see Appendix \ref{sec:Additional-Results-of-matmul}.

\begin{table}[t]
\begin{adjustbox}{center}

\begin{tabular}{cc|ccc|cccc|ccc}
\toprule 
 &  & \multicolumn{3}{c|}{(a)} & \multicolumn{4}{c|}{(b)} & \multicolumn{3}{c}{(c)}\tabularnewline
\midrule
\midrule 
\multicolumn{2}{c|}{Exp. rate} & 1.0  & 0.8  & 0.5  & 1.0  & 0.8  & 0.5  & GP & 1.0  & 0.8  & 0.5\tabularnewline
\midrule 
\multirow{5}{*}{Sample size} & $n=10$  & 0.72  & 0.66  & 0.62 & 2  & 2  & 1  & 2 & 2.71  & 2.64  & 2.63\tabularnewline
\cmidrule{2-12} 
 & $n=30$  & 0.57  & 0.54  & 0.48 & 2  & 5 & 3 & 4 & 2.89 & 2.83 & 2.85\tabularnewline
\cmidrule{2-12} 
 & $n=50$  & 0.55  & 0.51  & 0.45  & 4  & 8  & 6 & 9 & 2.84 & 2.91 & 2.86\tabularnewline
\cmidrule{2-12}
 & $n=70$  & 0.55  & 0.57  & 0.49 & 14  & 14 & 11  & 11 & 2.88  & 2.90  & 2.91\tabularnewline
\cmidrule{2-12} 
 & $n=90$  & 0.54 & 0.63 & 0.54 & 15 & 19 & 16  & 13 & 2.94 & 2.94 & 2.97\tabularnewline
\bottomrule
\end{tabular}
\par
\end{adjustbox}
\caption{\label{tab: matmul-LR-2block} (a) The percentage of cGP fitted surrogates
with optima equal or better than the baseline GP surrogate under different
exploration rates. (b) The number of cGP (and GP) fitted surrogates
that attain the actual optimal matrix size $\bm{x}_{\max}=112$ under
different exploration rates in 100 different random seeds. (c) The
average number of additive components in the fitted cGP surrogate
models among 100 different random seeds. %
}
\end{table}

\FloatBarrier

\subsubsection{High-dimensional tuning}\label{sec:high-D tuning}

We acknowledge that there are different definitions for high
dimensionality. In the tuning context, we usually consider 3- to 32- dimensional
domains as high-dimensional tuning domains. This is partially due
to the fact that grid searches are no longer an effective strategy in
these dimensions. When we conduct a grid search on a one or two dimensional
domain, the grid size (i.e., the number of grid points) is usually
of the magnitude of $10^{3}$ to $10^{6}$ when we use $~1000$ grids
in each dimension. However, when the dimensionality is 3, the magnitude
grows to $10^{9}$, which takes a long time to enumerate.
And when the dimensionality exceeds 32, even a binary enumeration
vector would take more than $2^{32}$ bytes$=$32Gb to assign.
This already exceeds the usual cache size of current computational
machines, and makes the grid search infeasible.

In high-dimensional domains, the tuning problem can benefit from adopting
surrogate models \citep{gramacy_surrogates_2020}. Like the GP surrogate
generalizes into high dimensions, it is possible to generalize our
cGP model into a domain with higher dimensions. However, the limited
sample size would restrict our focus since we need more samples to
fit a higher dimensional surrogate model. In the existing literature,
synthetic and real examples with dimensionality higher than 32 are
seldom studied, with an exception by \citet{chen_joint_2012}.

\emph{\ref{sec:high-D tuning}.1 Emulated piston cycle time ($d=7$).}
The first high-dimensional
application we study is the piston cycle time model which was
proposed in \citet{kenett_modern_2013} for quality control in industrial
applications, and has been well-studied in \citet{owen_higher_2013}.
It is important to tune the variables (e.g., piston surface area,
initial gas volume, etc.) of a piston to manufacture products whose
minimum and maximum cycle time are within a certain range. This piston
model function is a continuous function describing the cycle time
of a piston, defined on a 7-dimensional domain with all of its tuning
variables being continuous. Although the model function is continuous,
empirically it does produce dramatic changes that make it almost
look ``non-smooth'', as we already observed in the Bukin N.6
function (Figure \ref{fig:The-Bukin-N.6}). In this specific application,
both maximum and minimum are of interest. Therefore,  we expect that
we need to fit a good surrogate to find all function optima. 

\begin{figure}[th]
\begin{adjustbox}{center}%
\begin{tabular}{cc}
\toprule 
(a) & (b)\tabularnewline
\midrule
\midrule 
\includegraphics[width=7cm,page=2]{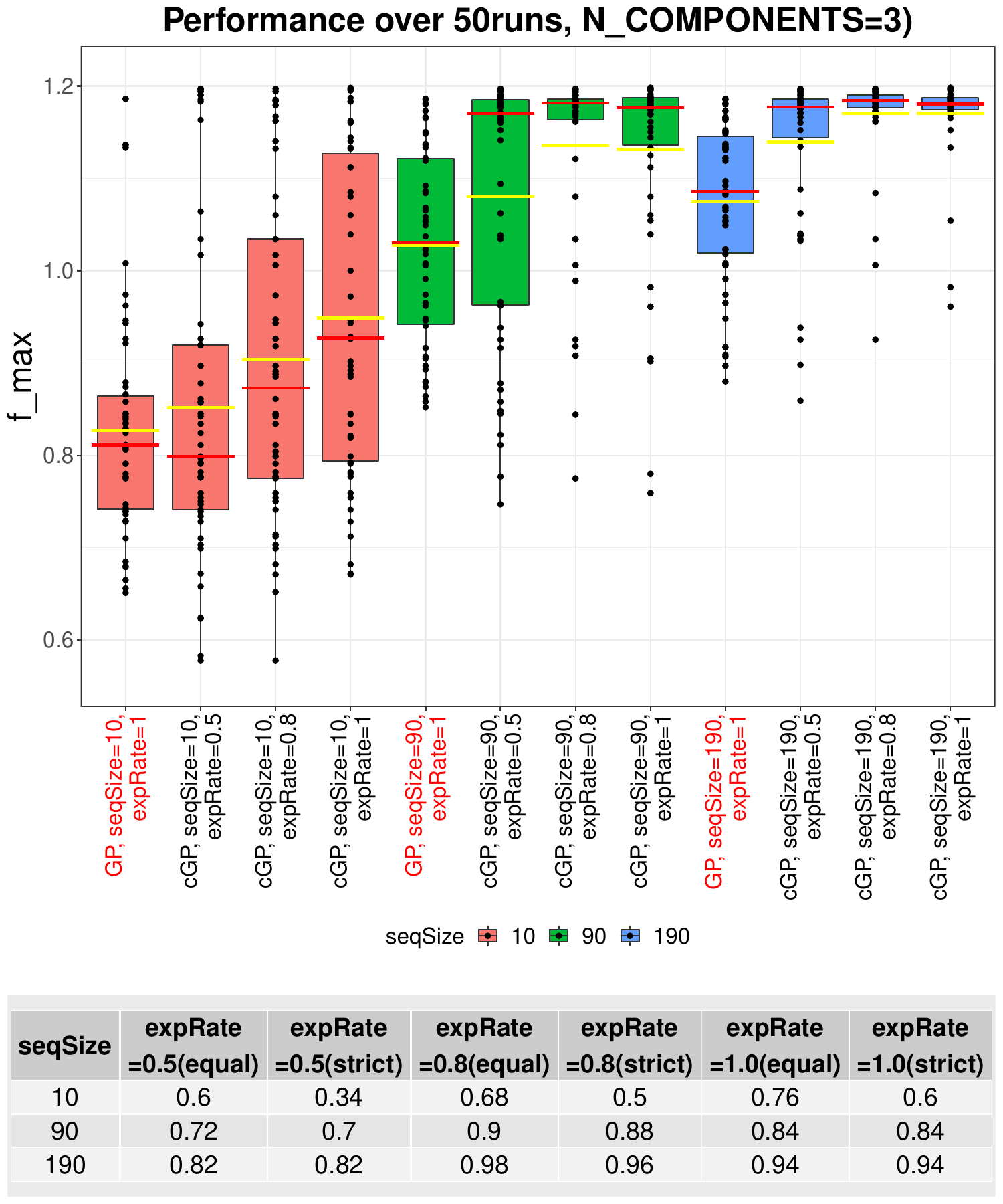} & \includegraphics[width=7cm,page=2]{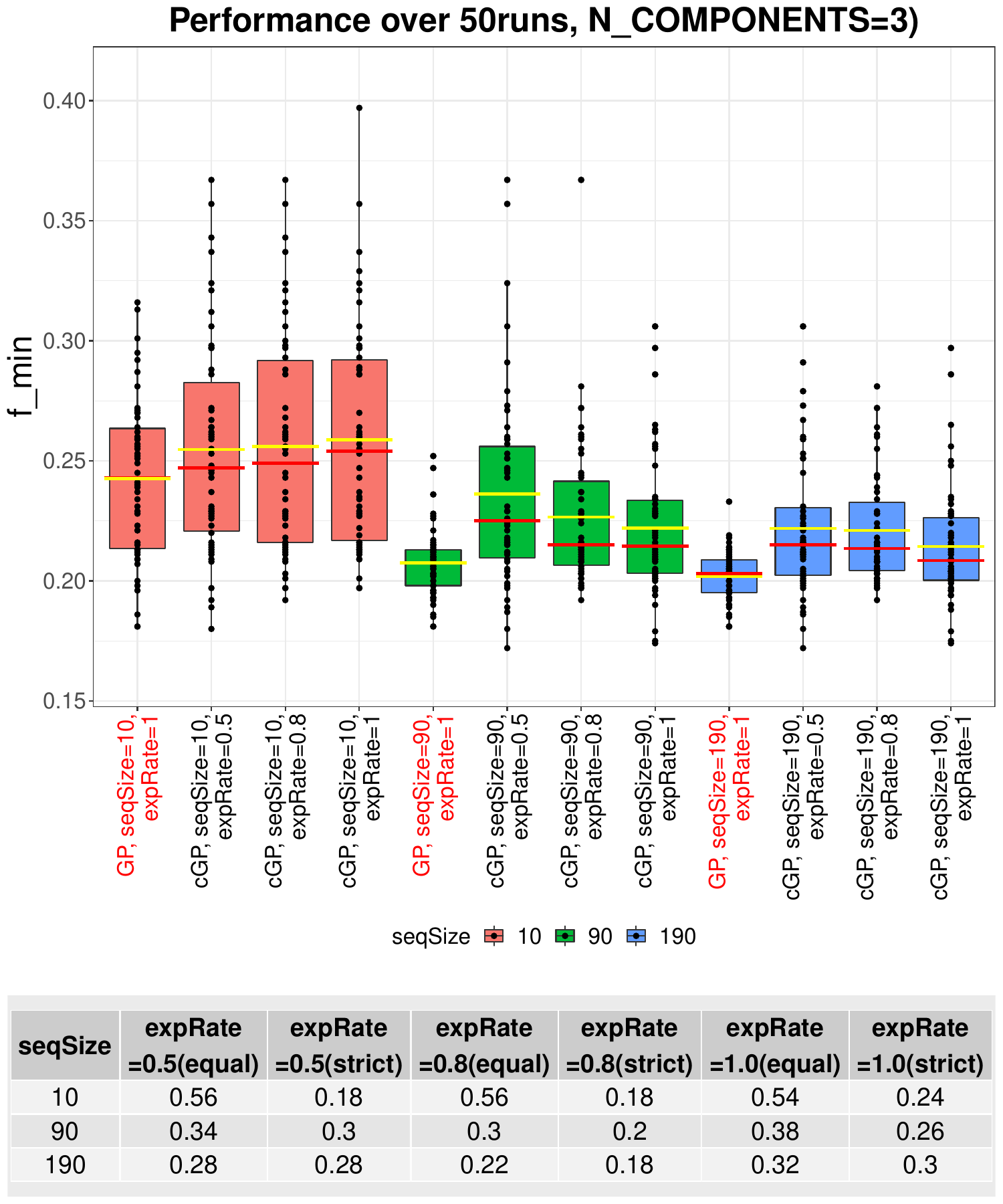}\tabularnewline
\bottomrule
\end{tabular}\end{adjustbox} \caption{\label{fig: piston-LR} (a) The box plot of $f_{\max}$ with y-scale
$(0.5,1.2)$ and (b) $f_{\min}$ with y-scale $(0.1,0.8)$ from sequential
samples of 50 fitted surrogate models of piston simulation function,
each dot in the box plot represents the optimal point in each run
(hence each box plot contains 50 points). The sequential sample size
varies from 10 to 190 with 10 pilot samples. Exploration rates
of the cGP model are chosen to be 0.5, 0.8 and 1. We also show the
sample mean (yellow horizontal bar) and median (red horizontal bar)
among 100 optima points.}
\end{figure}

In Figure \ref{fig: piston-LR}, we show the performance of GP and
cGP surrogate models. 
We use the DGM classifier and set the maximal number of components to
be 3 for cGP and compare against the GP with the same kernel. The
choice of the number of components does not affect the results much
for this application. However, we still calibrate the exploration
rate and increase the allowed sequential sample size in our exploration,
and show all boxplots on the same scale. We can see that cGP clearly
outperforms GP when we search for $f_{\max}$. Panel (a) in Figure
\ref{fig: piston-LR} shows that cGP systematically gives larger $f_{\max}$
and performs much better than GP surrogate when there is a lack of
sequential samples. When we search for $f_{\min}$, panel (b) in Figure
\ref{fig: piston-LR} shows that both GP and cGP produce similar
$f_{\min}$. However, cGP still explores the high-dimensional domain
more efficiently and gives the smallest $f_{\min}$ among GP surrogates.
Note that in the $f_{\max}$ situation, cGP produces a roughly 0.10
in function value improvement on average (difference between sample
means of $f_{\max}$); but in the $f_{\min}$ situation, cGP only
loses less than 0.02 function value on average. Considering the fact
that this function is defined on a 7-dimensional domain, we observed
that cGP can approximate the underlying black-box function quite well
with relatively few samples. 

\FloatBarrier
In the table at the bottom of 
Figure \ref{fig: piston-LR}
we provide a similar quantitative comparison for GP and cGP surrogates.
The fraction %
of out-performing batches shows that the cGP with a
high exploration rate is preferred when looking for $f_{\max}$. 
Although
the quantitative percentage does not favor cGP when searching for
$f_{\min}$,
by comparing the sample means and medians obtained from
GP and cGP, the claim that cGP does not give a much worse $f_{\min}$
is supported. This kind of trade-off is common in choosing different
surrogates. It is worth pointing out that cGP models can have a significant
gain like this even if a real ``non-smoothness'' does not occur.
It is not hard to 

\emph{\ref{sec:high-D tuning}.2 SuperLU\_DIST parallel tuning ($d=4$).} The second high-dimensional
application we are studying is the tuning problem arising in the distributed-memory parallel sparse LU factorization of SuperLU\_DIST \citep{yamazaki_new_2012}.
SuperLU\_DIST is a numerical software package developed for LU factorization
of large sparse matrices in parallel. There are several
mechanisms introduced to speed up the LU factorization for large matrices.
We use 8 Haswell nodes with 32 cores on each node, and 256 MPIs in
total. This tuning problem has 4 tunable parameters, LOOKAHEAD (number of lookahead panels used to overlap the communication with computation), nprow
(number of row processes out of 256 MPI processes), NSUP and NREL (parameters defining sizes of the so-called supernode representations). 
There are other categorical parameters COLPERM (different ways
of permuting the columns) etc., for which we take the default paramter values of SuperLU and do not tune in this example.

\begin{figure}
    \centering
    \includegraphics[width=.9\textwidth]{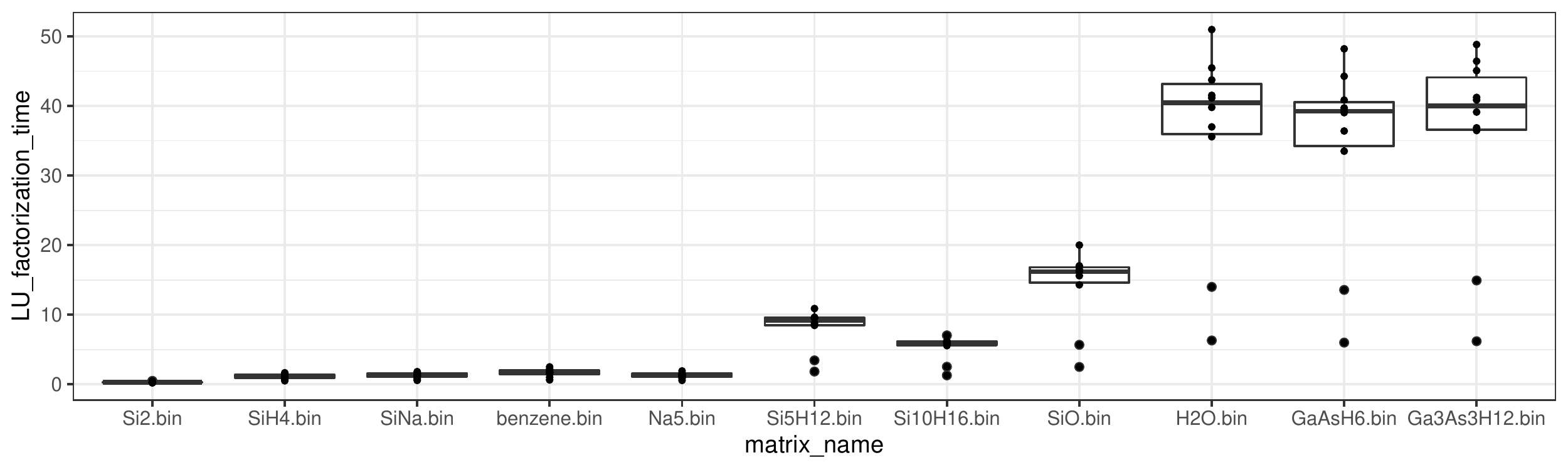}
    \caption{Box plots for LU factorization time when 4 parameters in the SuperLU application are randomly chosen with the same random number generator. To reduce randomness, we repeat 10 times for each parameter configuration and use the average value for each point in each column. }
    \label{fig: superLU-Box}
\end{figure}

All of these parameters are integers, and so we need to round them to the nearest integer when fitting the surrogate model
over a continuous domain. The black-box objective function $f$ in
this tuning problem is the running time of LU factorization for a given matrix; our examples arise in chemistry. In Table \ref{fig: superLU-table} and Figure \ref{fig: superLU-Box},
we display the approximate time range for the LU factorization when cGP explores the 4-parameter space with different parameter configurations,
the details of these matrices are available at \url{https://sparse.tamu.edu/PARSEC}.%

\FloatBarrier
\begin{table}[th]
\begin{adjustbox}{center}%
\begin{tabular}{cccccc}
\toprule 
\multicolumn{1}{|p{1.5cm}|}{\centering Size of raw data file}  & \multicolumn{1}{|p{2cm}|}{\centering Matrix dimension ($d\times d$ square matrices)} & \multicolumn{1}{|p{2cm}|}{\centering Number of nonzeros entries in matrix} & \multicolumn{1}{|p{1.5cm}|}{\centering Matrix name}  & \multicolumn{1}{|p{3cm}|}{\centering Approximate range of factorization times (seconds)} & \multicolumn{1}{|p{2cm}|}{\centering GP baseline factorization time (seconds)} \tabularnewline
\midrule
\midrule 
224Kb  & $769$ & 17801 & Si2  & $\approx$ 0.10-0.90 & 0.145 \tabularnewline
\midrule 
2.0Mb  & $5041$ & 171903 & SiH4  & $\approx$ 0.20-2.0 & 0.283 \tabularnewline
\midrule 
2.3Mb  & $5743$ & 198787 & SiNa  & $\approx$ 0.20-2.0 & 0.311 \tabularnewline
\midrule 
2.9Mb  & $8219$ & 242669 & benzene  & $\approx$ 0.30-4.0 & 0.362 \tabularnewline
\midrule 
3.6Mb  & $5832$ & 305630 & Na5  & $\approx$ 0.20-3.0 & 0.285 \tabularnewline
\midrule 
8.6Mb  & $19896$ & 738598 & Si5H12  & $\approx$ 1.0-14.0 & 1.136 \tabularnewline
\midrule 
11Mb  & $17077$ & 875923 & Si10H16  & $\approx$ 0.6-8.0 & 0.698\tabularnewline
\midrule 
16Mb  & $33401$ & 1317655 & SiO  & $\approx$ 2.0-22.0 & 1.992\tabularnewline
\midrule 
26Mb  & $67024$ & 2216736 & H2O  & $\approx$ 5.5-53.0 & 5.686\tabularnewline
\midrule 
39Mb  & $61349$ & 3381809 & GaAsH6  & $\approx$ 5.0-55.0 & 5.207\tabularnewline
\midrule
36Mb  & $61349$ & 5970947 & 	Ga3As3H12  & $\approx$ 5.0-65.0 & 5.382\tabularnewline
\bottomrule
\end{tabular}\end{adjustbox}  \\
\caption{\label{fig: superLU-table} The fixed matrices 
(stored in raw text files with the size indicated in bytes) considered in the SuperLU\_DIST
application and their typical factorization time (without tuning)
on 8 cori Haswell nodes, with 32 cores used on each node. We consider
those matrices smaller than 10Mb as ``small'' while the others ``large''. We also include baseline execution time when the application is tuned by a simple GP surrogate. }
\end{table}
\FloatBarrier
As above, we fix the random seed to ensure the pilot samples are the
same for each surrogate model. Even with the same random seed, the
running time has some noise due to the working condition of each node.
In our experiments, we use the DGM classifier with the maximal number of components $k$=2, 4 and expRate=0.5,
0.8, 1.0 for the cGP model and compare their performance against the simple
GP surrogate model. 
In this application, we want to find an optimal
configuration of the SuperLU\_DIST application so that the minimum running
time $f_{\min}$ becomes as small as possible. $f_{\max}$ is not
relevant in this application. 

\begin{figure}[th]
\begin{adjustbox}{center}\includegraphics[width=15cm,height=3.5cm,keepaspectratio,page=5]{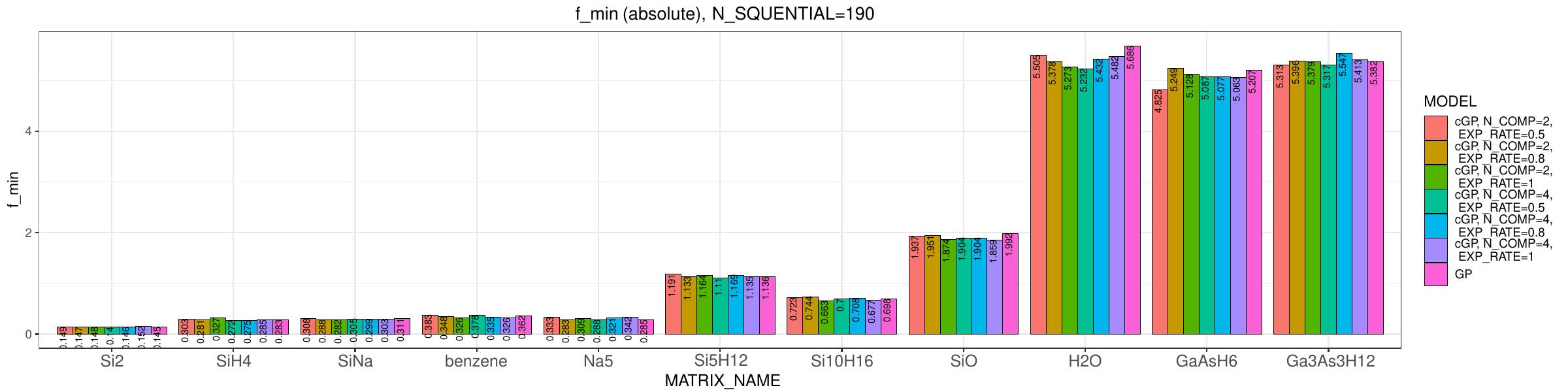}\end{adjustbox}\\
\begin{adjustbox}{center}\includegraphics[width=15cm,height=3.5cm,keepaspectratio,page=6]{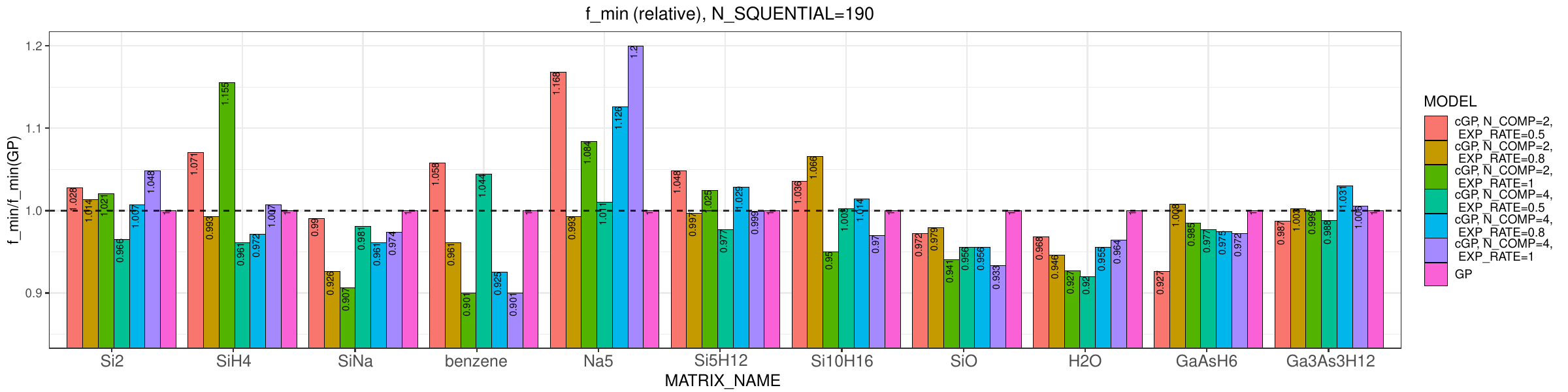}\end{adjustbox}\\
\caption{\label{fig: superLU-LR-3} In the top panel, 
we show the absolute
SuperLU\_DIST running time ($f_{\min}$) obtained by each specific surrogate
model with 190 sequential samples and 10 pilot samples
in one run. In the bottom panel, we show the relative ratio ($f_{\min}$
obtained by certain cGP models divided by the $f_{\min}$ obtained by simple
GP) of SuperLU\_DIST running time obtained by each specific surrogate model
against the one using simple GP surrogate model, ratios that are less than 1 means better performance. 
Further results can be found in Appendix \ref{sec:Additional-Results-of}.}
\end{figure}

In Figure \ref{fig: superLU-LR-3}, we show the performance of GP
and cGP surrogate models. 
In these applications, the choice of the
maximal number of components affects the result. For matrices smaller
than 10Mb, we can see that cGP can provide configurations of SuperLU\_DIST that are
faster than using the GP surrogate. cGP is usually not much worse than GP
except when in the Na5 matrix (but its absolute value is also close). Therefore, we can use
cGP in place of GP without much worry about getting a much worse $f_{\min}$.
For matrices larger than 10Mb, the noise in the running time is negligible,
and we observe that cGP is generally producing a better factorization
time with only a few exceptions. The improvement can be as much as
7\% 
(7.3\% for GaAsH6 matrix). Again, we expect that when cGP is giving
a worse result (e.g., Si10H16 matrix), it is within a 7\% margin
and
for a better choice of exploration rate (1.0, the number of components $k=2$), we can achieve a 5\%
improvement. Such an improvement is quite important when we often
need to conduct LU factorization multiple times. 

The geometries of the objective function in both piston and SuperLU\_DIST
applications are not known to us. And it is hard to explore the true
geometry as we did for the one-dimensional matmul application. However,
we show the power of cGP by simply choosing a few different cGP surrogate
models. In the piston application, we can see that $f_{\max}$ is
much improved while $f_{\min}$ only loses slightly. In the SuperLU\_DIST
application, we can see that $f_{\min}$ gets improved over GP results
very frequently, and again it would not lose by a large margin. 

\FloatBarrier

\section{\label{sec:Discussion}Discussion}

\subsection{Contribution }

In this paper, we start with a comprehensive literature review of
recent progress in Bayesian optimization. By a thorough review and
practical experience, we identify the non-smoothness problem in the
tuning context. For tuning problems, we usually have limited samples
and tune more than one or two parameters. The non-smoothness problem
is more challenging when we have limited samples and relatively high
dimensionality. This specific problem has not been treated nor explored
to the best of our knowledge, but arises in the tuning context.

To address this potential problem in the tuning context, we propose
a new clustering-based method that potentially improves surrogate
modeling for non-smooth black-box objective functions, which extends
to high-dimensional domains without much increase in complexity. The accompanying algorithm with acquisition re-weighting is a parsimonious but novel way to fit an additive GP model by automatically creating partitions, which can be utilized in the Bayesian optimization and more general online contexts.

The cGP surrogate model works well when the black-box function shows non-smoothness
or even changes sharply near the optima. When we have a clear knowledge
of the underlying geometry of the problem, this knowledge usually
helps us to choose the parameters of cGP surrogate models to attain
better results (e.g., Bunkin N.6 in Figure \ref{fig:The-boxplot-of-Bukin N.6}). However, even when we do not  know the objective functions for certain applications, cGP
may help us to identify the geometry (e.g., matmul in Section \ref{sec:matmul}). Even if we do
not care about the geometry of the objective function, we can usually
get improvement in our tuning task (e.g., piston, large matrices in
SuperLU\_DIST in Section \ref{sec:high-D tuning}). In the least favorable situation, using cGP would not lose
much in tuning tasks compared to GP (e.g., small matrices in SuperLU\_DIST). 

The cGP model usually has a computational advantage over GP since it fits multiple smaller GP components instead of one large GP. This advantage is more
obvious when we have a lot of samples to feed the surrogate model. When we have a bigger computational budget and many samples (as opposed to limited and sequential samples), a cGP is $k^2$ times (where $k$ is the number of clusters in the cGP model) faster than a GP with the same number of samples (assuming the clusters are roughly the same size).
In addition, cGP can be considered as a generalization of the GP model, when
we choose the maximal number of components to be 1, cGP reduces to
the GP model. In this sense, cGP can be a standalone modeling method that
incorporates the change-surface detection within a Bayesian framework
as well. Further extensions to the sparsified versions are possible. 

In terms of the partitioning induced by the cluster-classify step, we connect the sample points with the underlying black-box function in a novel way, which opens the door to more sophisticated designs to explore  properties of black-box functions using sequential samples.

\subsection{Future work}

Considering the high flexibility of the cGP surrogate model, we can
refine cGP in several natural ways. From a Bayesian perspective, it
would be natural to insert a prior mechanism to allow different kinds
of acquisition functions in cGP sequential sampling, potentially a
different weighting of clustering components with apriori information.
This may require us to modify the current cGP model into a fully Bayesian
framework which exceeds the scope of the current paper. The cGP model could also
be extended to multiple-output scenarios, where we may learn
the partition from multiple similar runs of the application.

Proposed as an additive model, we may want to allow different covariance
kernels for different components (or adaptive choice of kernels) induced
by the cluster-classify step in our algorithm. Combined with the current implementation,
we are able to parallelize the fitting of cGP, since the components
are conditionally independent. 

One practical issue is to determine the hyper-parameters used for the cGP model. For instance, we need to pick an appropriate number of components in the cGP model. One future direction is to develop a refined way of determining the number of additive components in our surrogate model. The typical cross-validation approach would not work here when there is non-smoothness, because we need to know the ``correct partition'' to create an appropriate training and testing set that are sampled from different regions in a balanced way.
Potentially, we need an algorithmic multi-level approach: at the coarse level we use a small number of samples to bound the number of components and pin down the number of components.

It is of theoretic interest how GP surrogate models work in conjunction
with non-stationary and/or compactly supported covariance kernels in the
tuning context. And it is also of theoretic interest whether  the clustering and classification consistency would be ensured when we have a lot
of sequential samples, although we have experimental results showing good large-sample behavior 
Or in other words, how many samples are needed to recover the ``true partition of the black-box function''? In the tuning context, an error bound for early stopping may be of more practical interest. Besides, replacing the simple GP model with more sophisticated models like deep GP \citep{damianou2013deep} would introduce another question: how can the trade-off between the number of model  parameters and the number of sequential samples be  balanced when we have limited samples for modeling?

One potentially more general approach is to choose different acquisition functions and exploration rates for different partition components adaptively. It is of interest to know how this relatively heuristic approach performs against the more sophisticated approach like optimization-based decomposition \citep{park2011domain}. And we also intend to develop new clustering criteria that fits into the cGP  framework. For instance, let each sample point $(\bm{x},y)$ be a vertex in a graph with edge weights equal to the reciprocal of a difference metric like $1/(\|\bm{x}_1-\bm{x}_2\|+\xi\cdot|y_1-y_2|)$.
Then we only keep those edges with large enough weights to form a graph-based clustering, so only ``nearby'' vertices representing ``no jumps'' (i.e., smoothness) are connected. %
The cGP model with the cluster-classify mechanism could leverage the (graph-based) convex clustering as a potential clustering algorithm \citep{sun2021convex}. 

Following the consideration of a different clustering method, a natural and important future direction for addressing non-smoothness in 
surrogate modeling is to consider higher order non-smoothness. In Definition \ref{def: (Non-smoothness)-Let-}, we are motivated by non-existence of the first-order gradients of black-box functions in an open neighborhood. 
We expect that higher order non-smoothness motivated by higher order gradients is more complicated and so will be efforts to model that kind of non-smoothness. And the incorporation of both geometric and topological information conveyed by the non-smooth features of the black-box function may assist our modeling further \citep{luo2019combining}. 

Another main question we have not touched in the current paper is the handling of categorical (or binary) variables in the tuning context. More often than not, high-performance computing applications bring categorical variables without explicit ordering (e.g., different kinds of algorithms used for a specific optimization problem). When there are few categories involved, we build separate surrogate models for each of these categories. However, it is still of interest how we could model non-smoothness in a tuning problem where many categorical variables with multiple categories involved. In addition, we can also view the choice of clustering and classification algorithms as a categorical variable.

To summarize, we have identified a rarely explored practical problem,
proposed and implemented a novel algorithm that generalizes the classical
GP surrogate model. This cGP model has the ability to handle non-smoothness
and complex geometry of the objective black-box function with limited online samples. We provide
evidence how this model outperforms the simple GP surrogate model for high and low dimensional functions,
and discuss its flexibility and possible extensions.

\subsection{Ethics Compliance}
\begin{itemize}
\item Code and data availability\\
We stored our code at \url{https://github.com/hrluo/cGP} for reproducibility
and production purposes. 
For the SuperLU\_DIST application, we use the embedded code stored at \url{https://github.com/gptune/GPTune} as part of GPTune project and the SuperLU software available at \url{https://github.com/xiaoyeli/superlu\_dist}. 
\item Acknowledgment/Funding source\\
We thank Giulia Guidi for providing us the C++ code for the matmul application and helpful comments.

This research was supported by the Exascale Computing Project (17-SC-20-SC), a collaborative effort of the U.S. Department of Energy Office of Science and the National Nuclear Security Administration.
We used resources of the National Energy Research Scientific Computing Center (NERSC), a U.S. Department of Energy Office of Science User Facility operated under Contract No. DE-AC02-05CH11231.
\end{itemize}
\newpage
\appendix

\section{\label{sec:cGP-Algorithm}cGP Algorithm}

\LinesNumberedHidden
\setcounter{algoline}{0}

\begin{algorithm}[h!]
\smaller{
\vspace*{-.1cm}
\setstretch{0.05}
\caption{\label{clustered GP} Clustered Gaussian process (cGP) surrogate model algorithm (the algorithm that carries out the idea
described in Sec. \ref{sec:Methodology})} 
\KwData{$X_{n_0,d}$ (data matrix consisting of pilot samples in $H^d$)} 
\KwIn{$X_{new}$ (data matrix of predictive locations in $H^d$), $k$ (optional, the number of clusters), $\tau$, (the exploration rate), $maxSampleSize$ (the maximal number of samples that can be drawn from  $f$)}
\KwResult{
list $g$ of GP surrogate model components corresponding to clusters}
Prediction of clustered GP surrogate model at predictive locations $X_{new}$ %

\SetKwFunction{GP}{GP} \GP($X_{n,d},Y_n$) is a Gaussian process surrogate model fitted with the data matrix $X_{n,d}$ and response $Y_n$. \\
\SetKwFunction{EI}{EI} \EI($x$,$g$) is an 
acquisition function based on the surrogate model $g$, evaluated at $x\in H^d$. \\
\SetKwFunction{CLUSTER}{CLUSTER} \CLUSTER($X_{n,d}$,$Y_n$,$k$) is a (e.g., $k$-means) clustering method performed on $X_{n,d},Y_n$ that returns the labels of a data matrix $X_{n,d}$ as an $n\times 1$ vector.\\
\SetKwFunction{CLASSIFY}{CLASSIFY} \CLASSIFY($X_{n,d}$,$(X^{'}_{n',d},L^{'}_{n',1})$) is a (e.g., $k$-nearest neighbors) classification method that is trained by a set of labeled data $(X^{'}_{n',d},L^{'}_{n',1})$; and returns the labels of a data matrix $X_{n,d}$ as an $n\times 1$ vector.\\
\SetKwFunction{uni}{uni} \uni($v$) is a function that returns unique values in the vector $v$.\\

\Numberline \Begin { 
\Numberline $X$ = $X_{n_{0},d}$ \\
\While{$i<maxSampleSize$}{ 

\Numberline Generate a random number $u\in[0,1]$ \\
\uIf{$u \leq \tau$}{
\Numberline $g$ = $[\cdots]$; $x0$ = $[\cdots]$; $ei0$ = $[\cdots]$ \\

\Numberline $dataLabel$ = \CLUSTER($X$,$k$); $clusLabel$ = \uni($dataLabel$) \\
\tcc{fit the clustered GP surrogate model with $X$ = $X_{n_{0},d}$}

\tcc{In practice, we also eliminate those clusters with too small sizes to avoid mis-fits.}
\For{j $\text{\bfseries{in}}$ $clusLabel$}{
\Numberline 	$X_j$ = $X[dataLabel == j,:]$ \\
\Numberline     $Y_j$ = $Y[dataLabel == j]$\\
\Numberline     $g[j]$ = \GP($X_j$,$Y_j$) \\
\Numberline 	$x0[j]$ = $\arg\max_{x\in H^d}$\EI($x$,$g[j]$); $ei0[j]$ = \EI($x0[j]$,$g[j]$) \\
\tcc{The \CLASSIFY($\bm{x}$,$(X,dataLabel)$ is implicitly used in this step.}
} 
\Numberline $j0$ = $\arg\max_{j\in clusLabel}ei0[j]/|dataLabel == j|$  \\
\tcc{weighted acquisition function by cluster size}   %
\Numberline Append $x0[j0]$ and its function values $f(x0[j0])$ to the $X$ and $Y$
}

\Else{
\Numberline Append random location $x1$ and its function values $f(x1)$ to the $X$ and $Y$ 
}
}
\tcc{predict based on the fitted clustered GP surrogate model}
\Numberline $dataLabel$ = \CLUSTER($X$,$Y$,$k$) \\
\For{$\bm{x} \text{ \bfseries{in} } X_{new}$}{
\Numberline 	Find the label \CLASSIFY($\bm{x}$,$(X,dataLabel)$) and let $j_{\bm{x}}$ = the $dataLabel$ of $\bm{x}$ \\ 
\Numberline 	Predict the value of surrogate function at $\bm{x}$ using GP model $g[j_{\bm{x}}]$ 
}
}
}
\end{algorithm}
\FloatBarrier
\newpage{}

\section{\label{sec:Additional-Simulation-Studies}Additional Synthetic Studies}

In this set of experiments, we pick a set of 2-dimensional benchmark
functions. For each of these functions, we know the exact minimum
from its expression. However, these analytical solutions are designed
to be difficult to find by popular optimization methods. We would
briefly describe the function at the beginning of each section and
display its minima. Among all these well-known benchmark functions,
practitioners only concern about minima. We do not have to consider
maxima, as suggested in most benchmarks. In addition, there
is a typical domain for these benchmark functions, over which people
would like to search for minima.

For each run, we assume that we always have 10 randomly chosen pilot
samples (it turns out that the optimal design would not systematically
help in synthetic function cases). In addition, we choose three different
sample sizes: 10 (limited sample size), 90 (normal sample size) and
190 (abundant sample size). The major factor we tune for cGP model
is the number of components ($k=2,3,4$), which is the maximal number of components
allowed in the cGP model. We use DGM cluster algorithm to decide clustering
labels; and k-NN $(k=3)$ to decide the classification labels. For each
random seed, we have the same set of 10 pilot samples, and fit GP
model, cGP models (with N\_SEQUENTIAL 10, 90, 190). For each sample
size, we also adjust the expRate of the cGP model as 0.5, 0.8
and 1.0. Therefore, we would have $3\times3=9$ different cGP models
against one GP model for a fixed sample size.

The first comparison is the violin plot, a generalization of the box
plot. The violin plot shows 10\%, 25\%, 50\%, 75\%, 90\% quantiles
of the estimated density of 100 minima obtained by each surrogate
model among 100 runs, with black solid lines. We also show the sample
mean (yellow horizontal bar) and median (red horizontal bar) among
100 points. In the violin plot, we can see the estimated density
of surrogate minima as the sequential sample size increases. In addition,
since minima are displayed as a distribution of dots in the plot,
we can avoid the delusion of a good single run and see the systematic
performance of a specific surrogate model against another model. When
we see a significantly lower clump, we can conclude that the cGP model
is distributionally better than the GP model. This shows a significant
trend and eliminates the randomness in single runs as much as possible.

The second comparison is the percentage table, as we have seen in
Figure \ref{fig:The-boxplot-of-Bukin N.6}. The table includes two
types of percentage. The ``(equal)'' percentage means the percentage
of batches that the corresponding cGP models give an equal \textbf{or}
smaller minimum among all sequential samples drawn. The ``(strict)''
percentage means the percentage of batches that the corresponding
cGP model gives strictly smaller minima among all sequential samples
drawn. When we see (equal) greater than 50\% and (strict) greater
than 1-(equal)\%, we can conclude that there is a significant advantage
of using cGP (with an appropriate setting). This means that at least
half of times cGP outperforms or equal to GP; and cGP performs better
more often than GP.
\newpage
\begin{enumerate}
\item The Bukin N.6 function has an expression $f(x)=100\sqrt{|x_{2}-0.01x_{1}^{2}|}+0.01|x_{1}+10|$.
The function has four global minima $f(x_{\ast})=0$ at $x_{\ast}=(-10,1)$.
We optimize over $[-15,5]\times[-3,3]$.\\
\begin{adjustbox}{center}\\
\includegraphics[width=5cm,height=8cm,keepaspectratio,page=1]{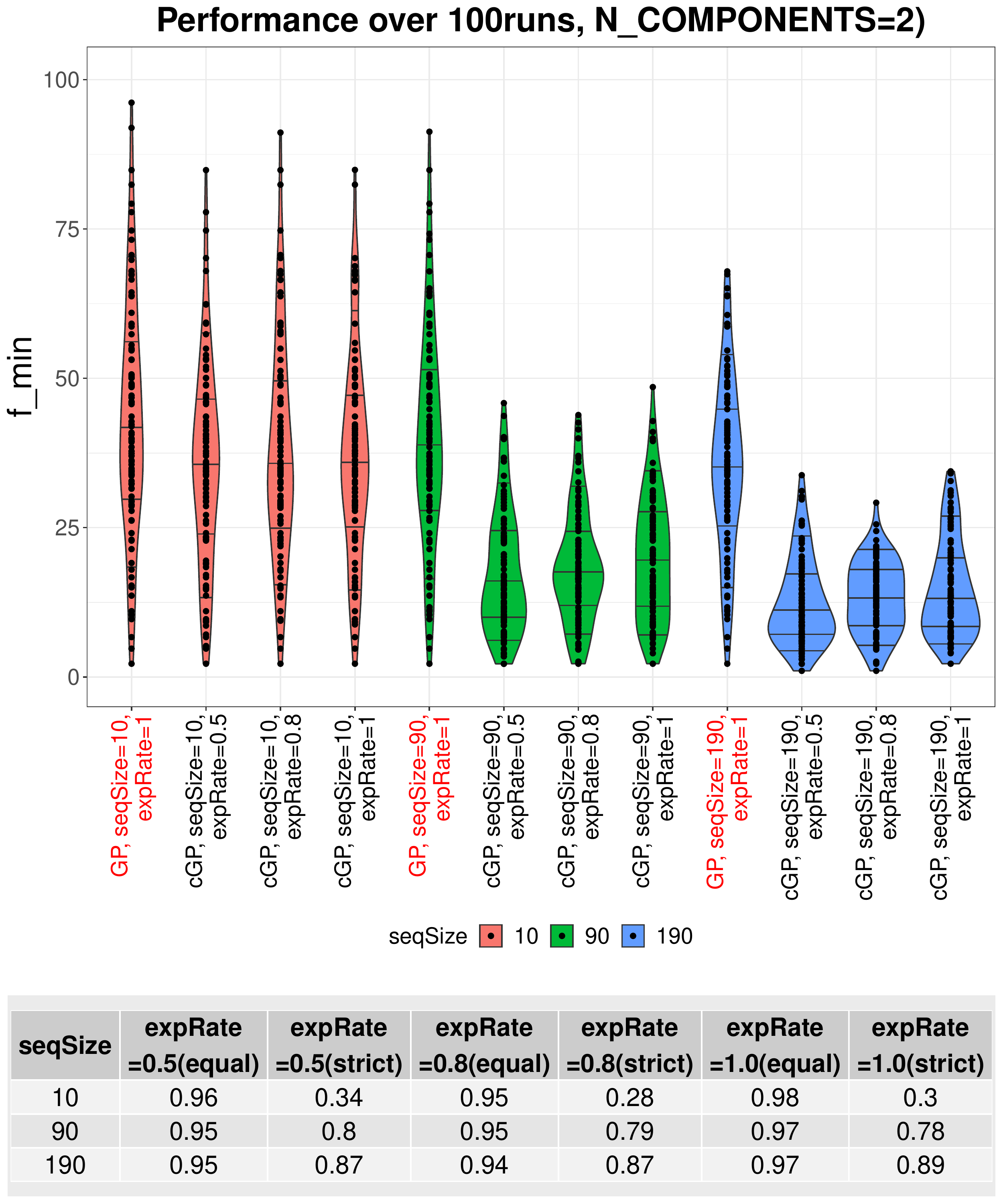}\includegraphics[width=5cm,height=8cm,keepaspectratio,page=2]{Figures/BUKIN_N6_f_min_analysis}\includegraphics[width=5cm,height=8cm,keepaspectratio,page=3]{Figures/BUKIN_N6_f_min_analysis}\end{adjustbox}
\item The Easom function has an expression $f(x)=-\cos(x_{1})\cos(x_{2})\exp(-(x_{1}-\pi)^{2}-(x_{2}-\pi)^{2})$.
The function has four global minima $f(x_{\ast})=-1$ at $x_{\ast}=(\pi,\pi)$.
We optimize over $[-10,10]\times[-10,10]$.\\
\begin{adjustbox}{center}\\
\includegraphics[width=5cm,height=8cm,keepaspectratio,page=1]{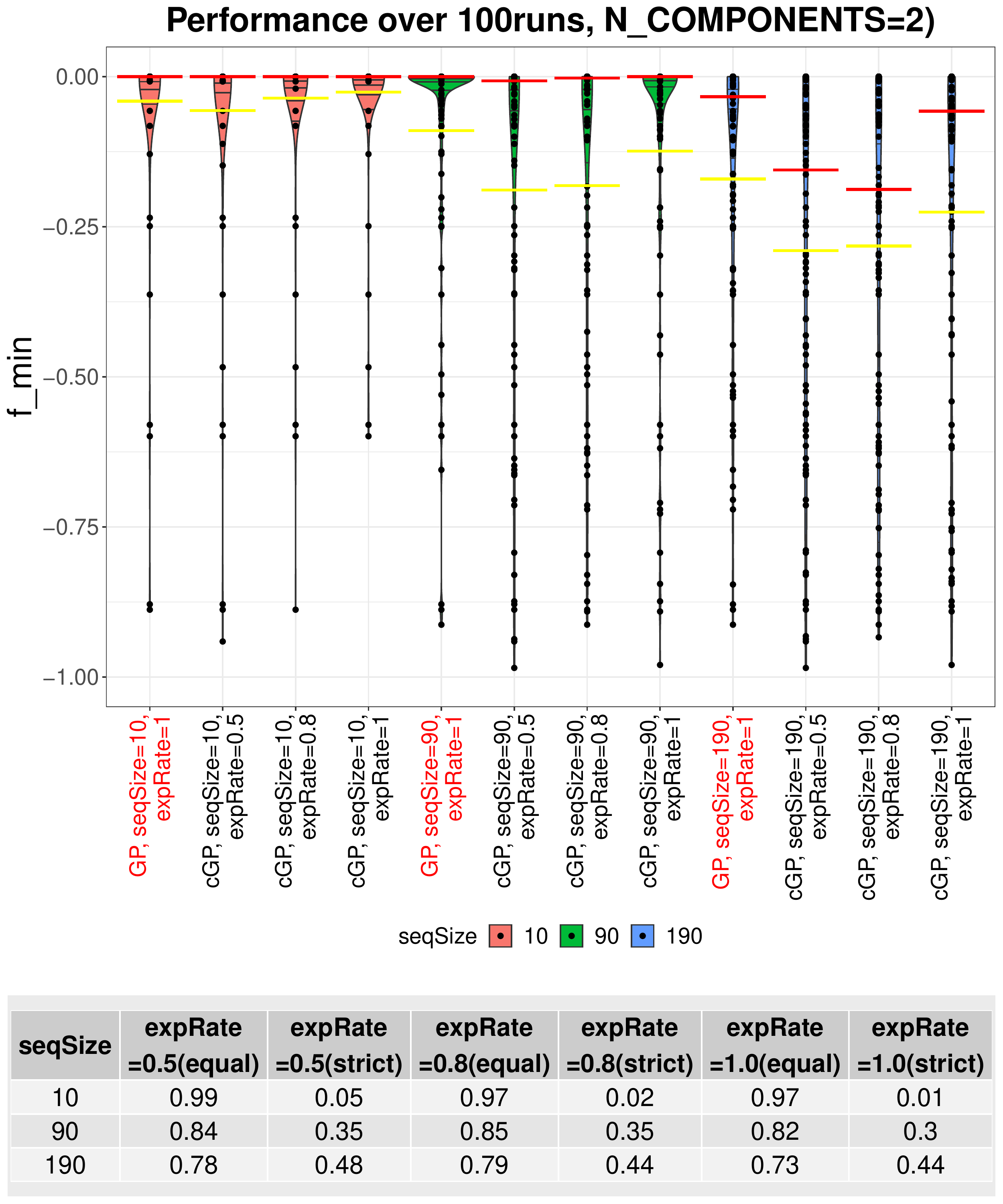}\includegraphics[width=5cm,height=8cm,keepaspectratio,page=2]{Figures/EASOM_FUNCTION_f_min_analysis}\includegraphics[width=5cm,height=8cm,keepaspectratio,page=3]{Figures/EASOM_FUNCTION_f_min_analysis}\end{adjustbox}
\item The Michalewicz function has an expression $f(x)=-\sin(x_{1})\sin^{20}\left(\frac{x_{1}^{2}}{\pi}\right)-\sin(x_{2})\sin^{20}\left(\frac{2x_{2}^{2}}{\pi}\right)$.
The function has four global minima $f(x_{\ast})=-1.8013$ at $x_{\ast}=(2.20,1.57)$.
We optimize over $[0,4]\times[0,4]$.\\
\begin{adjustbox}{center}\\
\includegraphics[width=5cm,height=8cm,keepaspectratio,page=1]{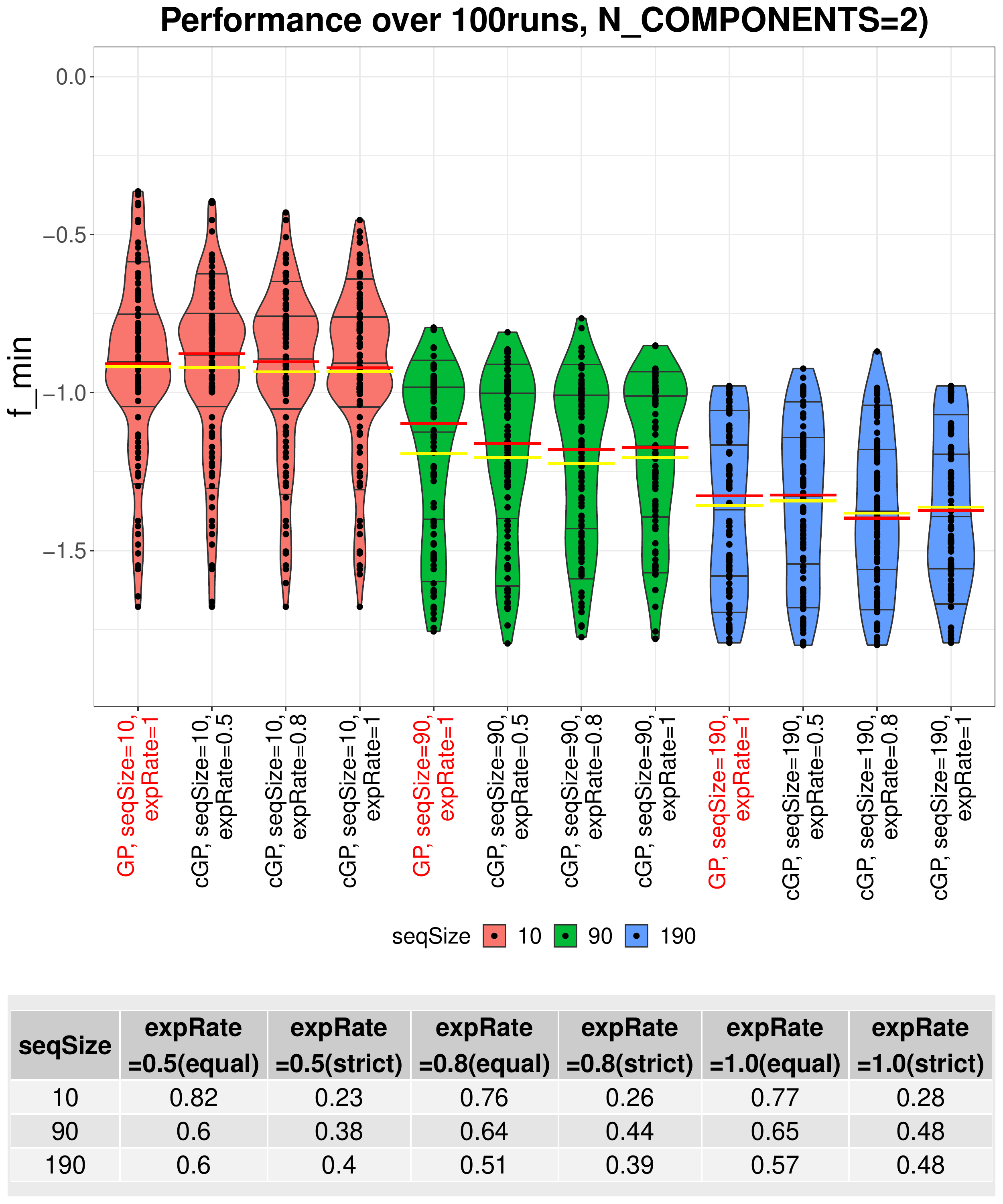}\includegraphics[width=5cm,height=8cm,keepaspectratio,page=2]{Figures/MICHALEWICZ_FUNCTION_f_min_analysis}\includegraphics[width=5cm,height=8cm,keepaspectratio,page=3]{Figures/MICHALEWICZ_FUNCTION_f_min_analysis}\end{adjustbox}
\item The Schaffer N.2 function has an expression $f(x)=0.5+\frac{sin^{2}(x_{1}^{2}-x_{2}^{2})-0.5}{\left[1+0.001\cdot(x_{1}^{2}+x_{2}^{2})\right]^{2}}$.
The function has four global minima $f(x_{\ast})=0$ at $x_{\ast}=(0,0)$.
We optimize over $[-2,2]\times[-2,2]$.\\
\begin{adjustbox}{center}\\
\includegraphics[width=5cm,height=8cm,keepaspectratio,page=1]{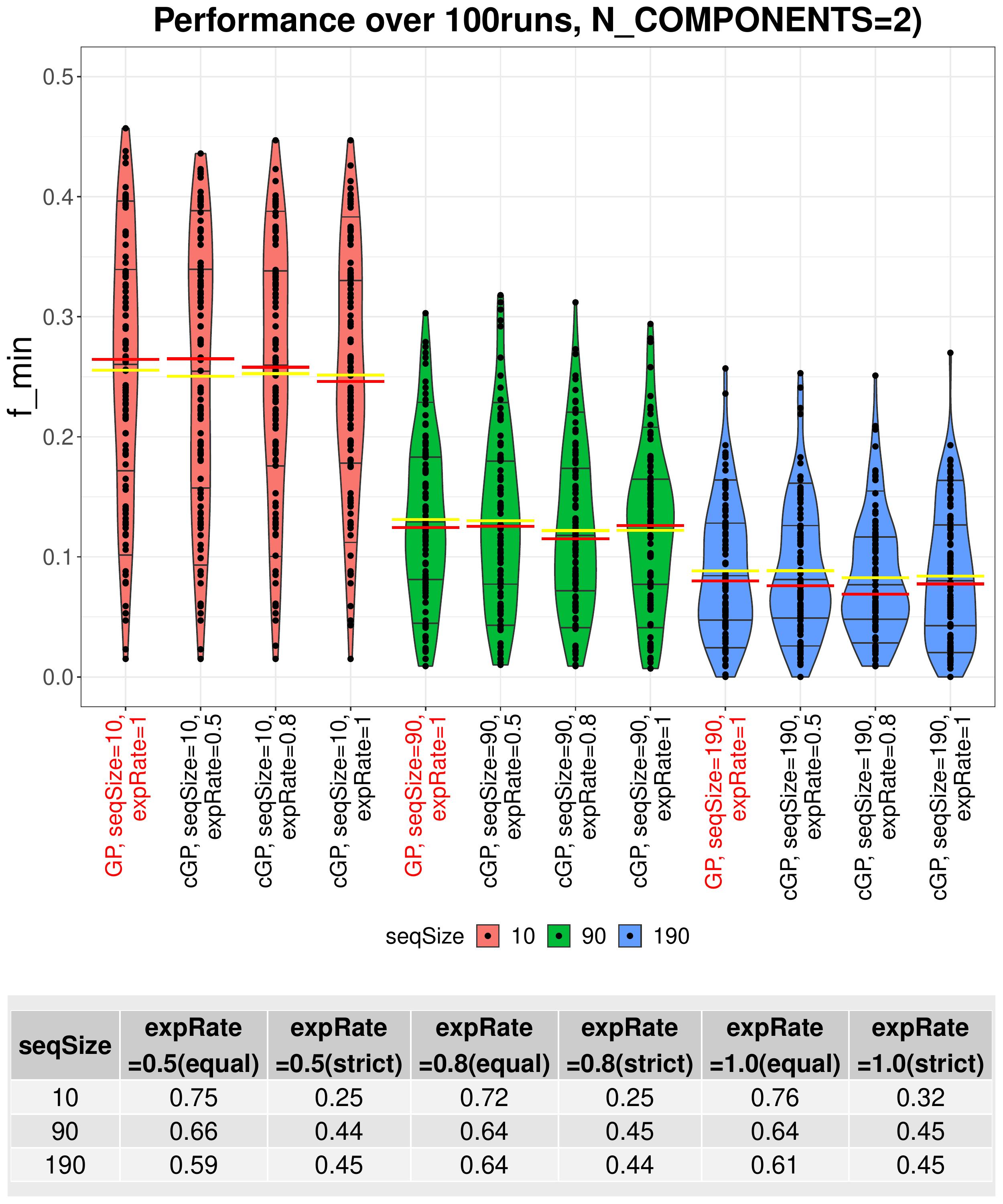}\includegraphics[width=5cm,height=8cm,keepaspectratio,page=2]{Figures/SCHAFFER_N2_f_min_analysis}\includegraphics[width=5cm,height=8cm,keepaspectratio,page=3]{Figures/SCHAFFER_N2_f_min_analysis}\end{adjustbox}
\item The Holder Table function has an expression 
\begin{align}
f(x)=-\left|\sin(x_{1})\cos(x_{2})\exp\left(\left|1-\frac{\sqrt{x_{1}^{2}+x_{2}^{2}}}{\pi}\right|\right)\right|.
\end{align}
The function has four global minima $f(x_{\ast})=-19.2085$ at $x_{\ast}=(\pm8.05502,\pm9.66459)$.
We optimize over $[-10,10]\times[-10,10]$.\\
\begin{adjustbox}{center}\\
\includegraphics[width=5cm,height=8cm,keepaspectratio,page=1]{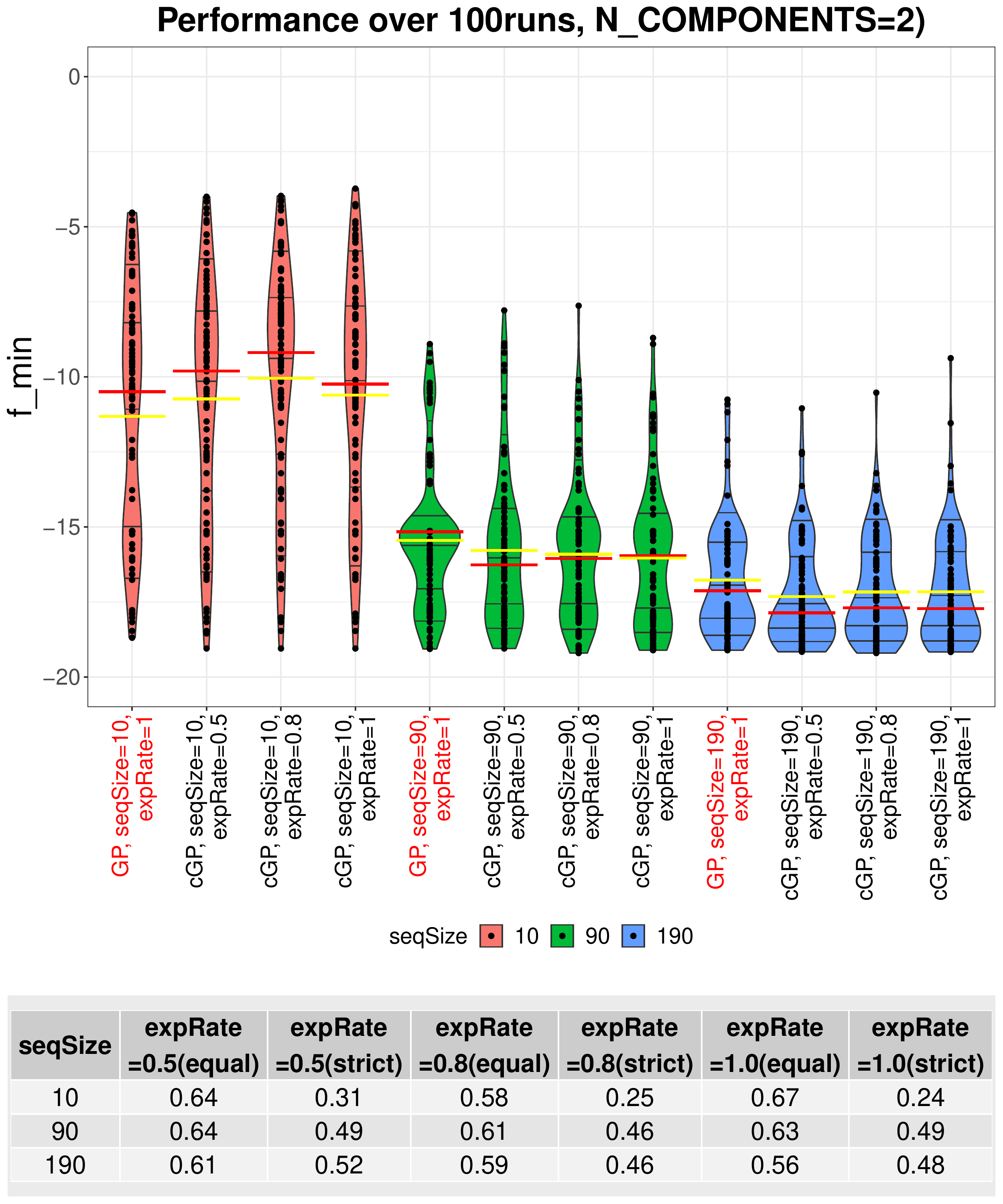}\includegraphics[width=5cm,height=8cm,keepaspectratio,page=2]{Figures/HolderTable_f_min_analysis}\includegraphics[width=5cm,height=8cm,keepaspectratio,page=3]{Figures/HolderTable_f_min_analysis}\end{adjustbox}
\item The Cross-in-Tray function has an expression 
\begin{align}
f(x)=-0.0001\left(\left|\sin(x_{1})\sin(x_{2})\exp\left(\left|100-\frac{\sqrt{x_{1}^{2}+x_{2}^{2}}}{\pi}\right|\right)\right|+1\right)^{0.1}.
\end{align}
The function has four global minima $f(x_{\ast})=-2.06261218$ at
$x_{\ast}=(\pm1.3494,\pm1.3494)$. We optimize over $[-10,10]\times[-10,10]$.\\
\begin{adjustbox}{center}\\
\includegraphics[width=5cm,height=8cm,keepaspectratio,page=1]{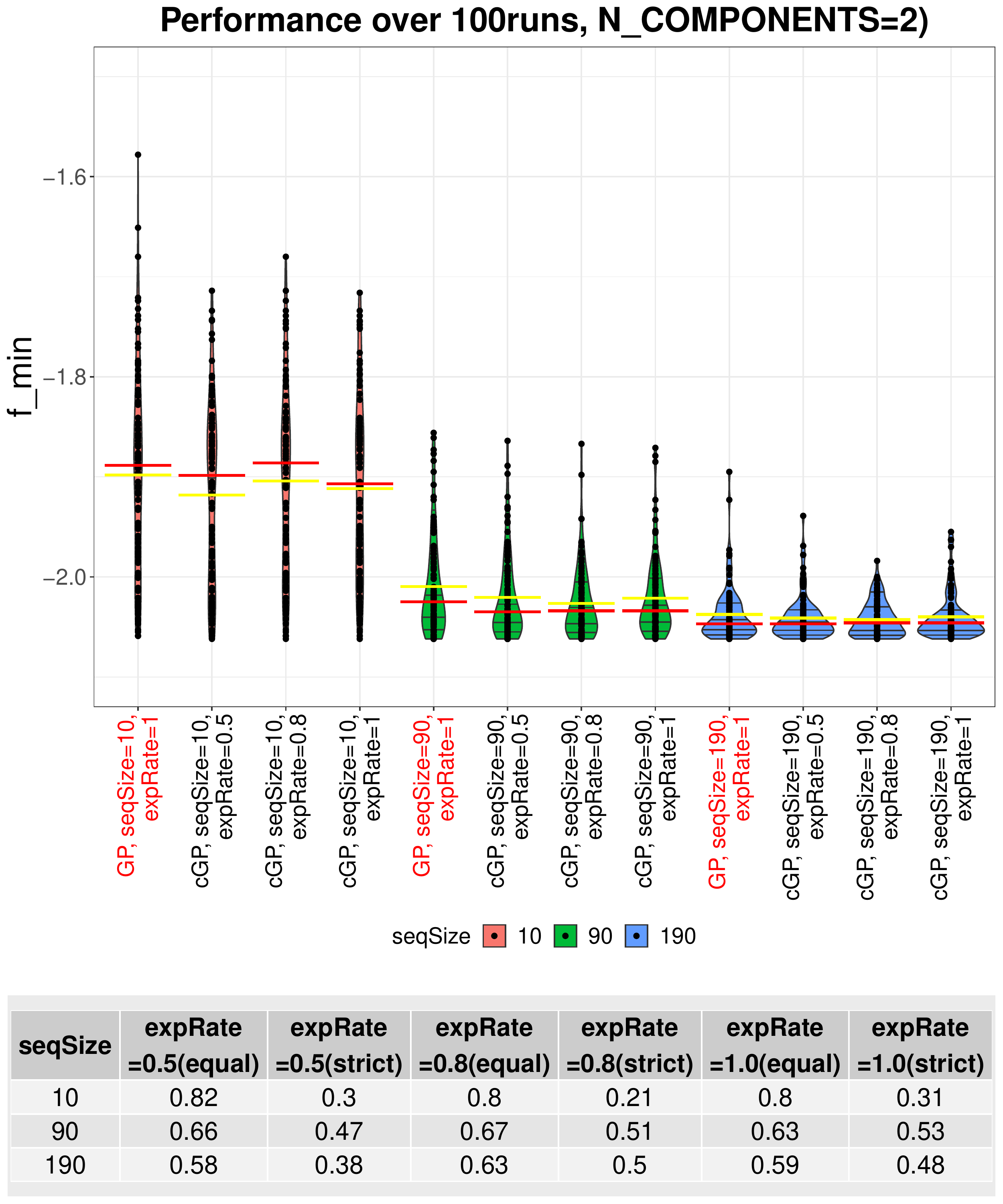}\includegraphics[width=5cm,height=8cm,keepaspectratio,page=2]{Figures/CIT_f_min_analysis}\includegraphics[width=5cm,height=8cm,keepaspectratio,page=3]{Figures/CIT_f_min_analysis}\end{adjustbox}
\end{enumerate}
Based on the results of the synthetic studies we obtained from these benchmark
functions, the following empirical observations are made:
\begin{enumerate}
\item If the black-box function has optima hidden in a ``slit-like''
(low-dimensional and also near dis-continuous) region, then cGP produces
better and faster results, along with a nice partition scheme (Bukin
N6 as in Figure \ref{fig:The-boxplot-of-Bukin N.6} and Easom). 
\item In some situations where the black-box function is smooth, cGP could
still show competitive performance using several smaller GP components.
It could also be tuned to show better tail behavior in the sense that
the obtained minima are closer to the truth more often (Michalewicz).
\item If the black-box function has a noisy and rough surface, then cGP
can be configured to perform as good as GP, usually faster, and come
up with clustering regimes. (Schaffer N2) 
\item If the black-box function has optima in a relatively small region
compared to the whole domain, then cGP may not produce very informative
regimes (Easom and Holder Table) but it would reach better optima;
and runs faster.
\item If the true function has significant dis-continuities and relatively
strong signal (matmul with averaging), then cGP can identify different
regimes rather faithfully. The common feature of those function where
cGP outperforms GP is that the function changes sharply (hence analogous
to a non-smoothness) near the minimum. 
\end{enumerate}
\section{\label{sec:Additional-Results-of-matmul}Additional Results of Experiments
in Section \ref{sec:matmul}}
Besides the problem setting of changing block size in the matmul application, we can also fix and optimize our blocking strategy but test the optimized blocking strategy with different matrix sizes. In this experiment, when the matrix size exceeds the memory cache size, the computational speed would experience an immediate drop, as we would observe in Figure \ref{fig: matmul-multiple8} below. The black-box function $f$ in this application is still  computational speed, but the tuning variable $\bm{x}$ is the matrix size.

All  
of their cache-line
sizes are 64 bytes, equivalent to 8 double precision floating point numbers. Based on this consideration,
we can sample the block sizes $b$ that are multiples of 8  in this example.%
\begin{figure}[th]
\begin{center}%
\begin{tabular}{cc}
\toprule 
(a) & (b)\tabularnewline
\midrule
\midrule 
\includegraphics[height=7cm]{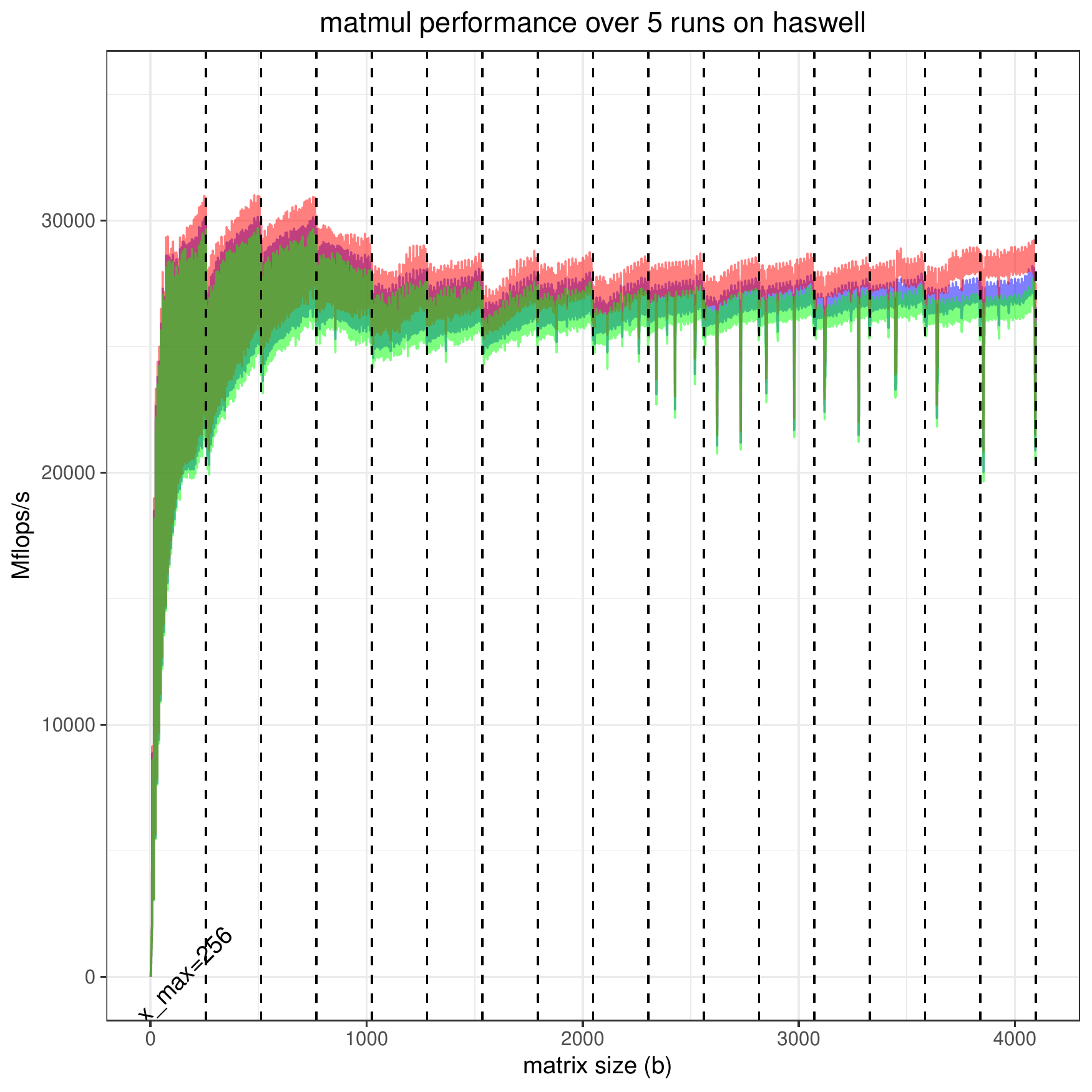} & \includegraphics[height=7cm]{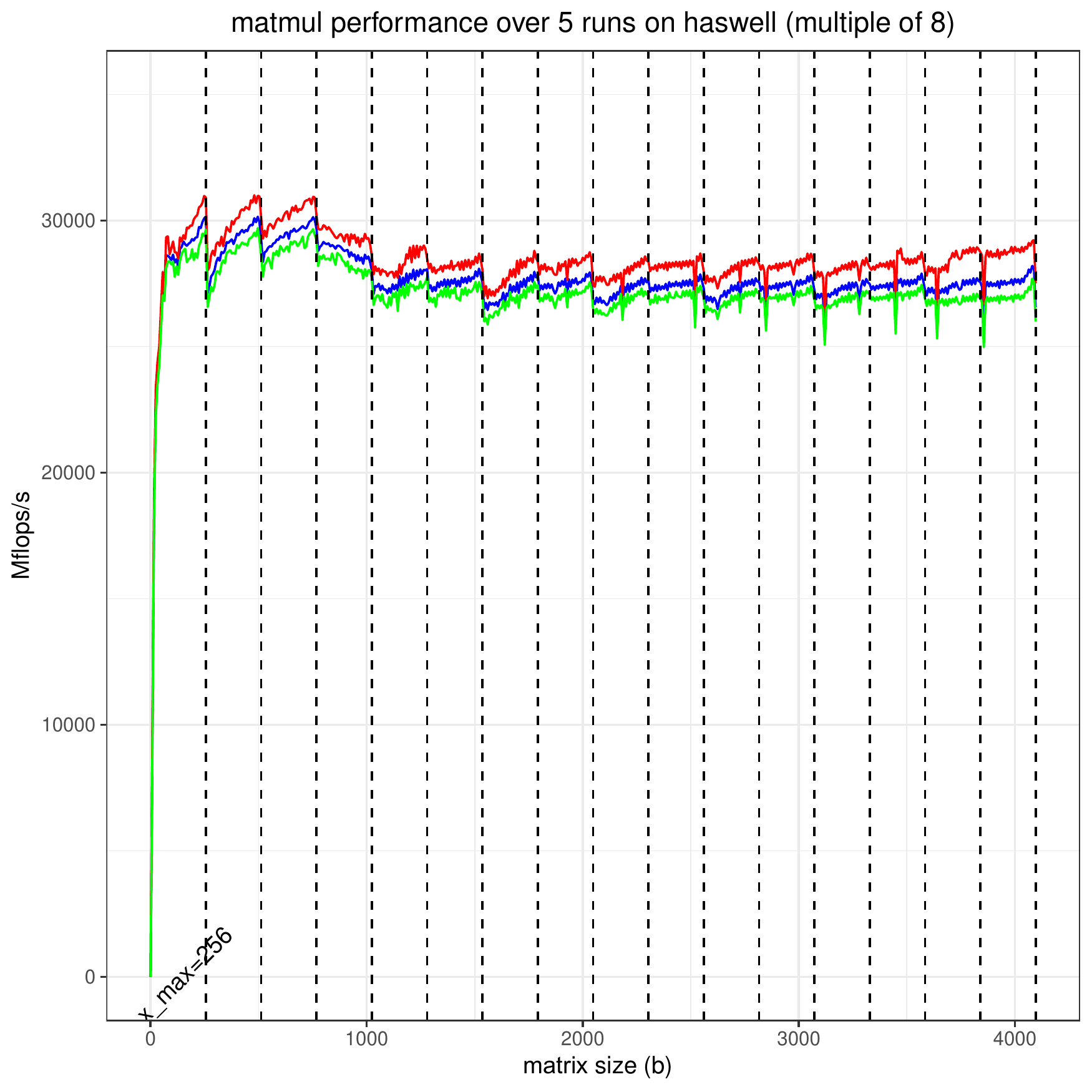}\tabularnewline
\midrule 
$\bm{x}_{\text{max}}=256$ & $\bm{x}_{\text{max}}=256$\tabularnewline
\bottomrule
\end{tabular}\end{center} \caption{\label{fig: matmul-multiple8} The computational speed for
matrices of sizes varying from $16$ to $4096$. The red line shows the maximum among 5
runs; the green line shows the minimum among 5 runs; the blue line
shows the average among 5 runs. (a) The matrix size varies from 16 to
4096, obtained from 5 different runs. (b) The matrix  sizes are multiples
of 8, varying from 16 %
to 4096. We also use dashed lines to indicate
the matrix sizes that are multiples of 256 where we expect a drop by machine  architecture.}
\end{figure}

\begin{figure}[ht!]
\begin{adjustbox}{center}%
\begin{tabular}{c}
\toprule 
\includegraphics[height=7cm]{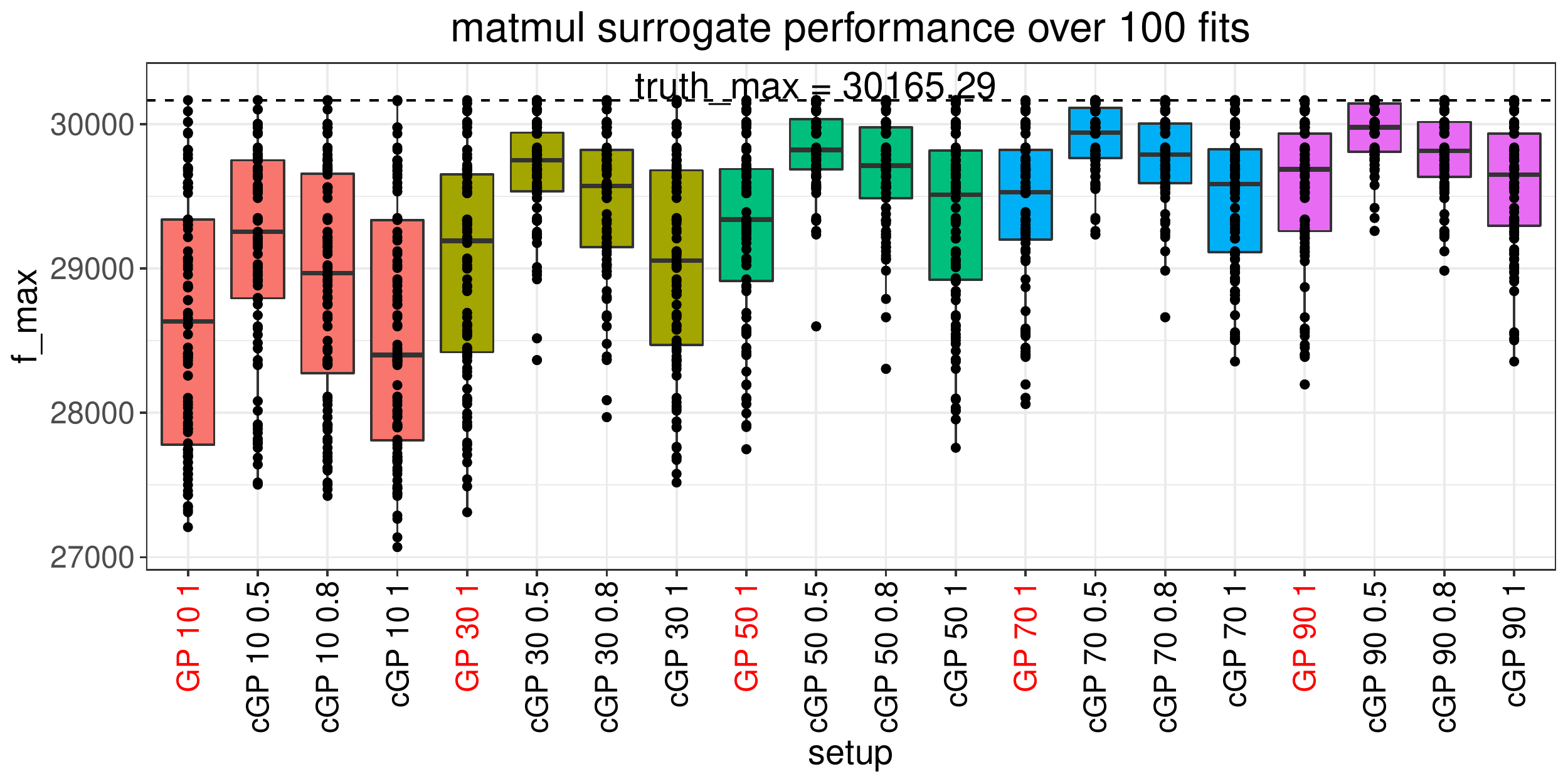} \tabularnewline
\includegraphics[height=7cm]{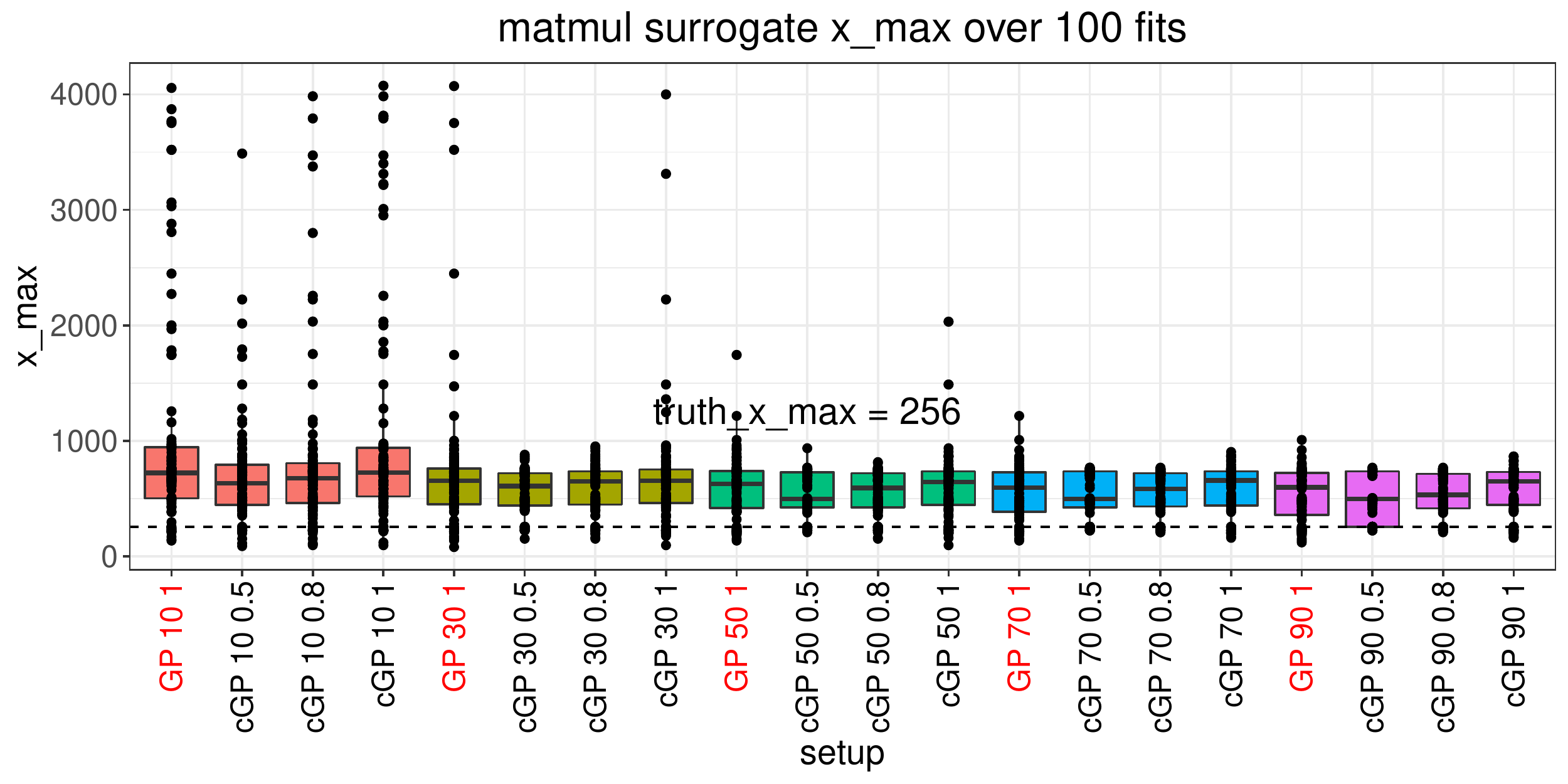}\tabularnewline
\bottomrule
\end{tabular}\end{adjustbox} \caption{\label{fig: matmul-LR-1} The box plot of $f_{\max}$ (top) and $\bm{x}_{\max}$ (bottom) elicited from sequential samples of 100 fitted surrogate
models, each dot in the box plot represents the optimal point in each
run (hence each box plot contains 100 points). The sequential sample
size varies from 10 to 190 with 10 pilot samples. The exploration
rates of the cGP model are chosen to be 0.5, 0.8 and 1. \protect \\
Results for simple GP surrogates are highlighted with red labels as
a baseline surrogate model; while cGP surrogates with different exploration
rates are highlighted with black labels.}
\end{figure}

To eliminate the randomness caused by a single run, we repeat 
cGP (and GP) for 100 different random seeds with 10 pilot random samples for
each run. 
In Figure \ref{fig: matmul-LR-1}, we compare the performance
of a simple GP surrogate model searching over any integer $[16,4096]\cap\mathbb{Z}$
against our cGP model with the exact number of components 3 (with $k$-means clustering) and $k=3$
in the k-NN classification step (See Figure \ref{fig:The-cluster-classify-scheme}).
Our cGP model searches over integers that are multiples of 8 over $[16,4096]$. Qualitatively,
it is not hard to see that cGP performs better than the GP model,
under almost all sample sizes.

In this matrix multiplication example with varying matrix sizes, the cause of non-smooth
points (i.e., reduction of communication; overflow of fast cache)
and different kinds of behavior are clear and observed in the recorded dataset as
shown in Figure \ref{fig: matmul-multiple8}. Our optimized strategy would attain the best performance on the matrix with matrix size $\bm{x}_\max=256$. In the summary Figure \ref{fig: matmul-LR-1}, cGP surrogates behave
similarly to the GP model when there are few samples; but when there
are enough samples, the cGP model clearly identifies different partition regimes much
better with a reasonable accuracy improvement. Therefore, the cGP model searches the optimum $f_{\max}$ and $\bm{x}_{\max}$ more
efficiently compared to a simple GP surrogate, with more evidence shown in Table \ref{tab: matmul-LR-2}.

\begin{table}[H]
\begin{center}%
\begin{tabular}{cc|ccc|ccc|ccc}
\toprule 
 &  & \multicolumn{3}{c}{(a)} & \multicolumn{3}{c}{(b)} & \multicolumn{3}{c}{(c)}\tabularnewline
\midrule
\midrule 
\multicolumn{2}{c|}{Exp. rate} & 1.0 & 0.8 & 0.5 & 1.0 & 0.8 & 0.5 & 1.0 & 0.8 & 0.5\tabularnewline
\midrule 
\multirow{5}{*}{Sample size} & $n=10$ & 0.57 & 0.69 & 0.78 & 2 & 2 & 2 & 1.79 & 1.85 & 1.83\tabularnewline
\cmidrule{2-11} 
 & $n=30$ & 0.65 & 0.74 & 0.82 & 3 & 3 & 5 & 2.08 & 2.10 & 2.20\tabularnewline
\cmidrule{2-11} 
 & $n=50$ & 0.67 & 0.74 & 0.85 & 4 & 5 & 9 & 2.15 & 2.20 & 2.27\tabularnewline
\cmidrule{2-11} 
 & $n=70$ & 0.62 & 0.74 & 0.82 & 4 & 6 & 11 & 2.24 & 2.31 & 2.38\tabularnewline
\cmidrule{2-11} 
 & $n=90$ & 0.61 & 0.70 & 0.80 & 4 & 7 & 12 & 2.30 & 2.37 & 2.38\tabularnewline
\bottomrule
\end{tabular}\end{center} \caption{\label{tab: matmul-LR-2} (a) The percentage of cGP fitted surrogates
with optima equal or better than the baseline GP surrogate under different
exploration rates. (b) The number of cGP fitted surrogates that attain
the actual optimal matrix size $\bm{x}_{\max}=256$ under different exploration
rates. A simple GP surrogate model would only reach $\bm{x}_{\max}=256$\textbf{
once} in 100 different random seeds. (c) The average number of additive
components in the fitted cGP surrogate models among 100 different
random seeds.
}
\end{table}

To better understand the performance behavior, we provide cache-miss profiling results of the matrix multiplication for varying matrix sizes.
We used Linux's \texttt{Perf} to collect the cache miss rates.
The profiling results show that the cache miss rate at each cache level has a jump when exceeding a certain matrix size.

We show the peak performance and cache traffic for only one execution of the matmul application. The execution is done on the same node type with the Haswell architecture as mentioned in Sec. \ref{sec:matmul}. 
We 
obtain the data read/load traffic of L1 (Data cache, 32KB), L2 (Unified cache, 256KB) and L3 (Unified cache, 40960KB; also known as LLC, last layer cache) cache.

\begin{figure}[h!]
\begin{adjustbox}{center}\includegraphics[width=15cm,height=6.5cm,keepaspectratio,page=1]{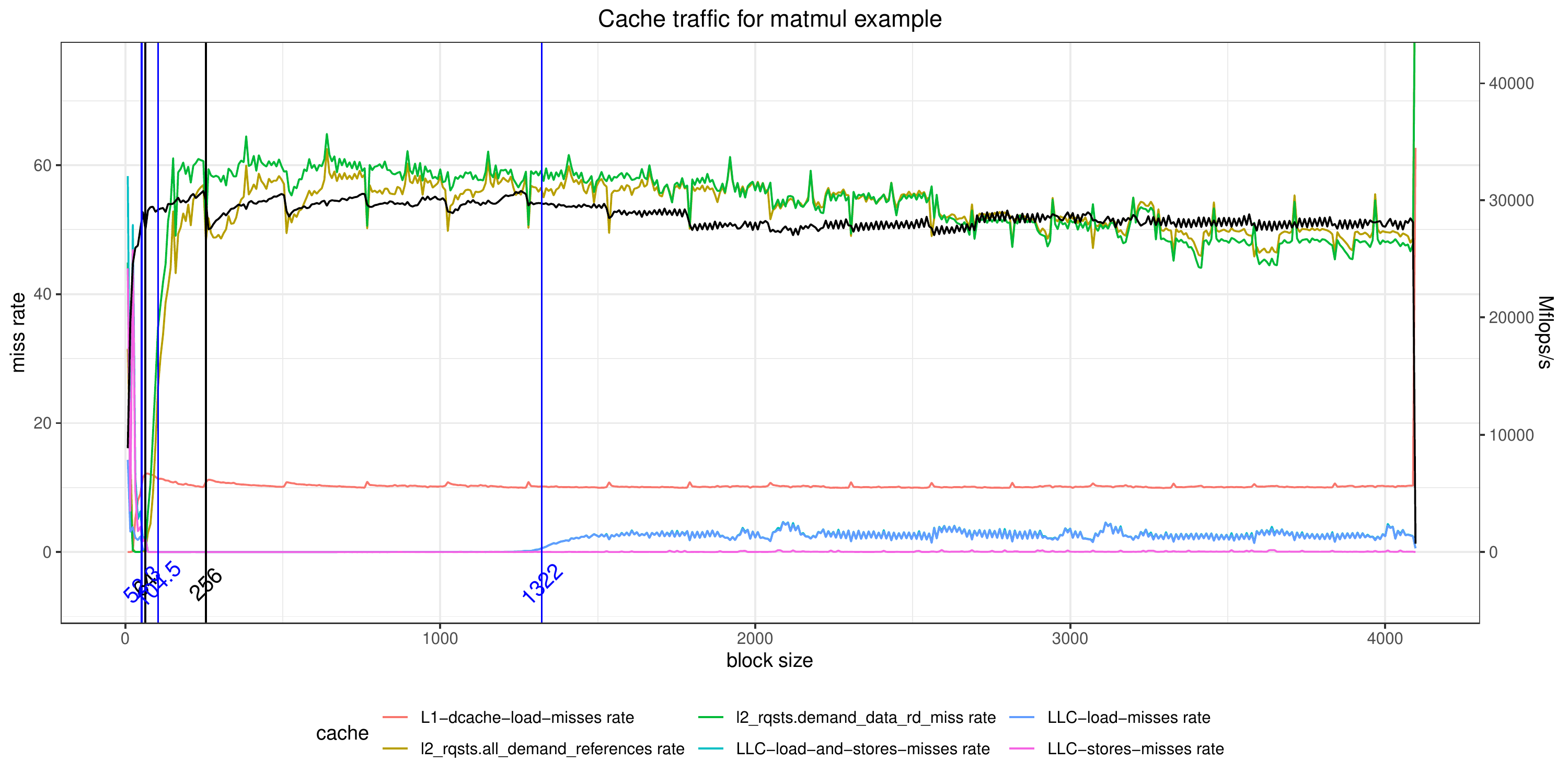}\end{adjustbox}\\
\begin{adjustbox}{center}\includegraphics[width=15cm,height=6.5cm,keepaspectratio,page=2]{Figures/dgemm}\end{adjustbox}\\
\caption{\label{fig:matmul-cache-1} 
Cache traffic and computational performance for one batch of the  matmul application. The top panel shows the traffic when the block size changes among multiples of 8 from 10 to 4096; the bottom panel shows a zoom-in version of the top panel focusing on the traffic for block size less than 512. 
The x-axis is the block size in the matmul application. The left y-axis is the percentage of miss rate for each type of cache. The right y-axis is the computational speed of the matmul application.}
\end{figure}

Theoretical calculations concerning only whether matrices fit into storage show that, ideally, the matrix size that makes the matrices exceed the L1 ($\approx$52), L2 ($\approx$105) and L3 ($\approx$1322) caches (using  $\sqrt{\frac{1}{3}\text{cache bytes}/8}$ for accessing three double precision square matrices in the matmul application.) These threshold matrix sizes are illustrated by blue vertical lines in the Figure \ref{fig:matmul-cache-1}.

The statistic we monitor is the \emph{cache miss rate}\footnote{Its definition is slightly different for L1, L2 and L3 cache, but generally it is the number of cache misses among all cache read/write during a certain procedure.}.
For L1 cache, we monitor the data load miss rate; for L2 cache, we monitor the data demand miss rate and all demand miss rate; for L3 cache, we monitor the data load, store miss rates and their sum. 
These hardware measurements lend support to our claim that the black-box function $f$ does have some non-smoothness and partially explain why the partition obtained by the cGP model is beneficial. It also  partially validates the findings of surrogate models in the tuning context. By incorporating the non-smoothness in the tuning context, our model also induces informative partitions consistent with non-smoothness observed in the profiling.

\FloatBarrier

\section{\label{sec:Additional-Results-of}Additional Results of Experiments
in Figure \ref{fig: superLU-LR-3}}
The following figures are organized in the same  formats:
In the top panel, we show the absolute
SuperLU\_DIST running time ($f_{\min}$) obtained by each specific surrogate
model with different number of sequential samples and 10 pilot samples
in one run. In the bottom panel, we show the relative ratio ($f_{\min}$
obtained by cGP models divided by the $f_{\min}$ obtained by simple
GP) of SuperLU\_DIST running time obtained by each specific surrogate model
against the one using the simple GP surrogate model, ratios that are less than 1 means better performance.
\begin{figure}[H]
\begin{adjustbox}{center}\includegraphics[width=15cm,height=3.5cm,keepaspectratio,page=1]{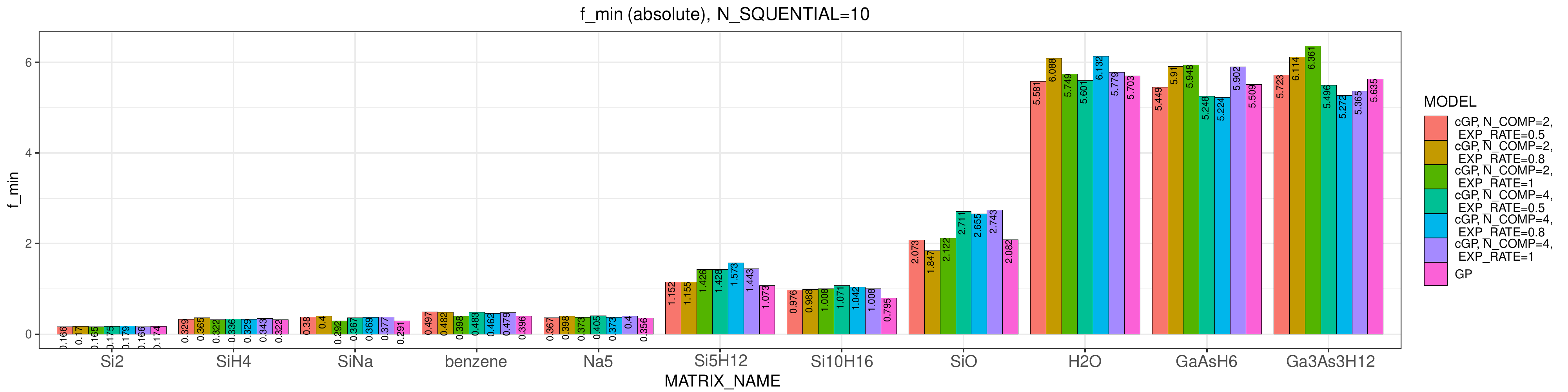}\end{adjustbox}\\
\begin{adjustbox}{center}\includegraphics[width=15cm,height=3.5cm,keepaspectratio,page=2]{Figures/mix_compare_updated}\end{adjustbox}\\
\caption{\label{fig: superLU-LR-1} SuperLU\_DIST results with 10 pilot and 10 sequential samples.}
\end{figure}

\begin{figure}[H]
\begin{adjustbox}{center}\includegraphics[width=15cm,height=3.5cm,keepaspectratio,page=3]{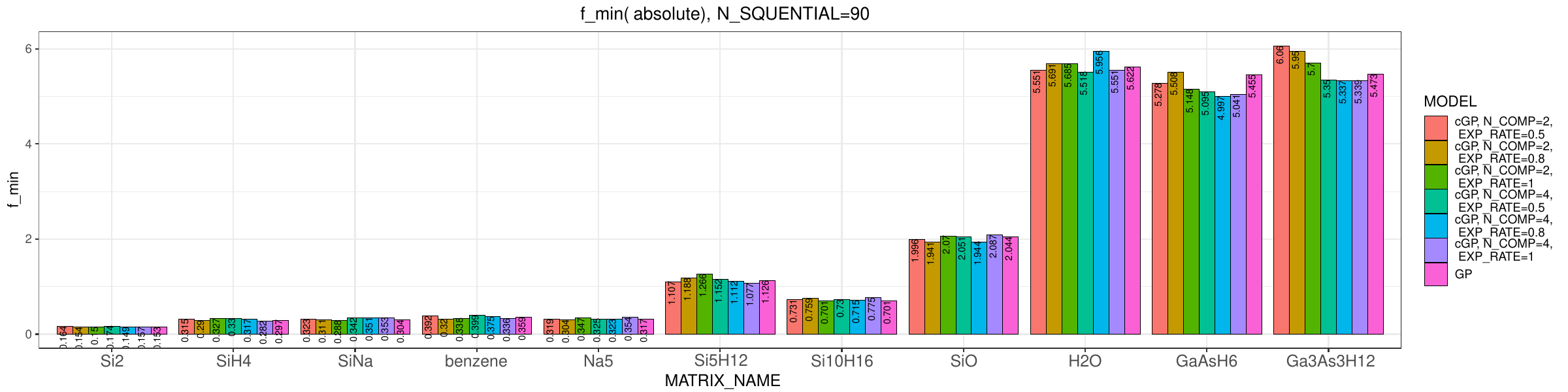}\end{adjustbox}\\
\begin{adjustbox}{center}\includegraphics[width=15cm,height=3.5cm,keepaspectratio,page=4]{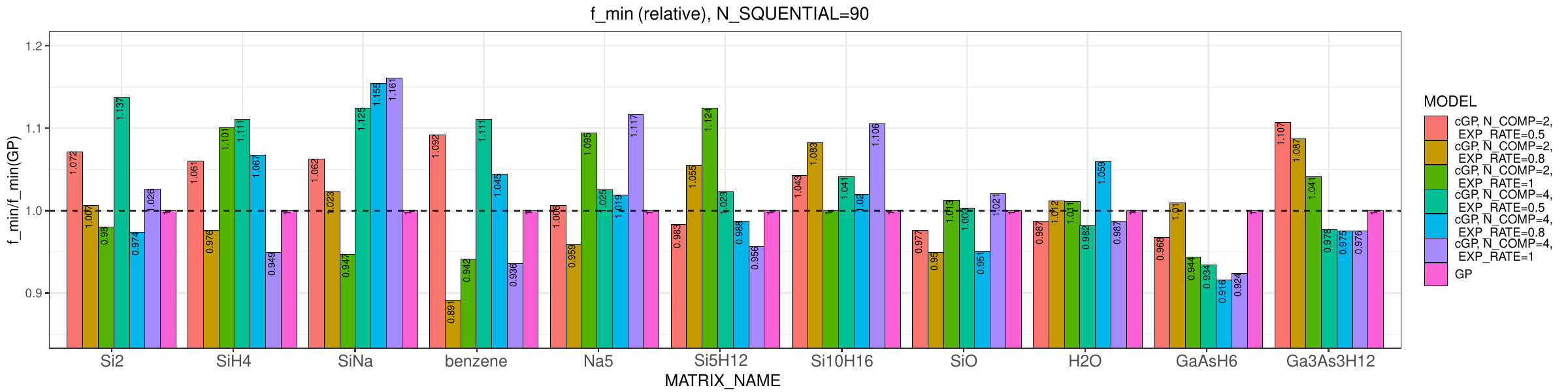}\end{adjustbox}\\
\caption{\label{fig: superLU-LR-2} SuperLU\_DIST results with 10 pilot and 90 sequential samples.}
\end{figure}
\FloatBarrier
\section{Theoretic Results \label{sec:theory discussion}}
Let us start with a simpler situation by supposing the partitions
of components are fixed from the beginning of sequential sampling.
\begin{quote}
\textbf{Assumption A.} The number of components $k<\infty$ is known
and cannot change. %
\end{quote}

To ensure the convergence of the surrogate model, we need to assume that the
covariance kernel $K$ satisfies:

\begin{quote}
\textbf{Assumption B.} The Fourier transform $\hat{K}(\bm{\xi})\coloneqq\int_{\mathbb{R}^{d}}2^{-2\pi i\left\langle \bm{x},\bm{\xi}\right\rangle }K(\bm{x})d\bm{x}$
exists, is isotropic and non-increasing and satisfy either $\hat{K}(\bm{\xi})=\Theta\left(\|\bm{\xi}\|^{-2\nu-d}\right)$
for some $\nu>0$ or $\hat{K}(\bm{\xi})=O\left(\|\bm{\xi}\|^{-2\nu-d}\right)$
for $\forall\nu>0$ (and denote $\nu=\infty$). We call this $\nu$
the \emph{smoothness parameter} of the covariance kernel $K$. (Assumption B is exactly the Assumptions 
1,2 and 3 in \citet{bull_convergence_2011})
\end{quote}
\begin{quote}
\textbf{Assumption C.} The covariance kernel $K$ is $\mathcal{C}^{\left\lceil 2\nu\right\rceil }$
and the $\left\lceil 2\nu\right\rceil $-th order Taylor approximation
$P_{K}$ satisfy $\left|K(\bm{x})-P_{K}(\bm{x})\right|=O\left(\|\bm{x}\|^{2\nu}\left(-\log\|\bm{x}\|\right)^{2\alpha}\right)$
as $\bm{x}\rightarrow0$ for some $\alpha\geq0$. (Assumption 4 in
\citet{bull_convergence_2011})
\end{quote}
Both Matern family and exponential kernel satisfy Assumption 2 and
3 above. When the partitions are fixed, the crucial condition of the
convergence of GP surrogate lies in the regularity of the component
domains and the sample size. The regularity of the component domain can
be described by 
\begin{definition}

(Lipschitz domain, \citet{wynne_convergence_2020}) An open set $\mathcal{X}_{i}\subset\mathbb{R}^{d}$
is called a (special) Lipschitz domain if there exists a rotation
of $\mathcal{X}_{i}$, denoted by $\bar{\mathcal{X}}_{i}$, and a
function $\psi:\mathbb{R}^{d-1}\rightarrow\mathbb{R}$ which satisfies
the following:

(1) $\mathcal{X}_{i}=\{(x,y)\in\mathbb{R}^{d},y>\psi(x)\}$

(2) $\psi$ is a Lipschitz function such that $\left|\psi(x)-\psi(x')\right|\leq M\|x-x'\|_{2}$
for any $x,x'\in\mathbb{R}^{d-1}$ and some $M>0$.
\end{definition}
For instance, the binary divisions adopted by \citet{chipman_bart_2010}
consist of Lipschitz domains with piecewise constant functions $\psi=c_{j}$.
\begin{proposition}
\label{prop:fixed partition GP convergence} Suppose Assumptions A
hold and that the domain $\mathcal{X}$ can be decomposed into Lipschitz
domains $\mathcal{X}_{1}\cup\cdots\cup\mathcal{X}_{k}=\mathcal{X}$
and each $\mathcal{X}_{j}$ is compact with an  nonempty interior. In
the reproducing kernel Hilbert space (RKHS) $\mathcal{H}_{\theta}(\mathcal{X})$ defined by the covariance
kernel $K_{\theta}$ on $\mathcal{X}$, let us suppose that black-box
functions $f\mid_{\mathcal{X}_{j}}\in\mathcal{H}_{\theta_{U}}(\mathcal{X}_{j}),j=1,\cdots,k$
with $\theta\leq\theta_{U}<\infty$ satisfy Assumption B and C. 
Then
there exists an integer $n_{0}>k$ such that the interpolant optimum \citep{bull_convergence_2011} 
$\bm{x}_{n}^{*}$ %
satisfy
\begin{equation}
\sup_{\begin{array}{c}
\|f\|_{\mathcal{H}_{\theta^{U}}(\mathcal{X}_{j})}\leq R\\
j=1,\cdots,k
\end{array}}\mathbb{E}_{f}\left|f(\bm{x}_{n}^{*})-\min_{\mathcal{X}}f\right|=O\left(\left(\frac{n}{\log n}\right)^{-\nu/d}\left(\log n\right)^{\alpha}\right)\label{eq:main result-statement 1}
\end{equation}
for some $R>0$ and all sufficiently large sample size $n>n_{0}$.
If $\nu=\infty$ then the statement holds for all $\nu<\infty$.
\end{proposition}

\emph{Proof.}
The idea of the proof is that the overall sequence $\bm{x}_{1},\bm{x}_{2},\cdots,\bm{x}_{n}$ of sequential sampling 
can be organized into sub-sequences for each component, which grow
simultaneously due to our algorithm. Within each component we have
an asymptotic scheme, the regularity of the component domains allow
us to apply results in \citet{bull_convergence_2011} obtained from the RKHS techniques for each component.

The sequential samples $\bm{x}_{1},\bm{x}_{2},\cdots,\bm{x}_{n}$
are chosen from one of all $k$ independent components in the additive
GP model. Therefore, if we use the super script to denote the component
a sample point belongs to, we can reorganize the sample sequence $\bm{x}_{1},\bm{x}_{2},\cdots,\bm{x}_{n}$
into $k$ sub-sequences where the sizes of sub-sequences satisfy $n_{1}+\cdots+n_{k}=n$:
\begin{align*}
\bm{x}_{1}^{(1)},\bm{x}_{2}^{(1)},\cdots,\bm{x}_{n_{1}}^{(1)} & ;\quad\bm{x}_{1}^{(2)},\bm{x}_{2}^{(2)},\cdots,\bm{x}_{n_{2}}^{(2)}\quad\cdots & \bm{x}_{1}^{(k)},\bm{x}_{2}^{(k)},\cdots,\bm{x}_{n_{k}}^{(k)}.
\end{align*}

Now we also need to show that all $n_{1},n_{2},\cdots,n_{k}>1$ and tend to infinity, for
otherwise, there would be components containing no sequential sample
points at all. However, this cannot happen, due to the fact that overall
sequential sampling strategy is to pick the maximizer of EI acquisition
function $\text{EI}_{j}(\bm{x})$ in each component weighted by component
sample size $n_{j}$. If $n_{1}=1$ 
while $n_{2},\cdots,n_{k}>1$ 
then we can pick $n_{0}$ such that 
\[
\frac{\max_{j=2,\cdots,k}\max_{\bm{x}\in\mathcal{X}_{j}}\text{EI}_{j}(\bm{x})}{\min(n_{2},\cdots,n_{k})+n_{0}/(k-1)}\leq\max_{\bm{x}\in\mathcal{X}_{1}}\text{EI}_{1}(\bm{x})/1\]
The numerator is the maximal acquisition function among all $\text{EI}_{j}(\bm{x}),j=1,2,\cdots,k$,
which is non-increasing by the definition of EI function. The denominator
$n_{0}/(k-1)$ can be intuitively understood as each sequential sample
avoiding $\mathcal{X}_{1}$ would add $1/(k-1)$ sample to each of
components $2,\cdots,k$.

Alternatively, assume that there exists some $j_0$ such that $n_{j_0}\not{\rightarrow}\infty$ and bounded by $n_{j_0}\leq N_0<\infty$ as $n\rightarrow\infty$. When $n$ is large enough, the following holds since the RHS is a constant and the LHS has its denominator tending to infinity while its numerator is bounded.
\[
\frac{\max_{j\in A}\max_{\bm{x}\in\mathcal{X}_{j}}\text{EI}_{j}(\bm{x})}{\min_{j \in A}n_{j}}\leq \frac{\max_{\bm{x}\in\mathcal{X}_{j_0}}\text{EI}_{j_0}(\bm{x})}{N_0}\]

Therefore, we can increase the sample size
$n>n_{0}$, then all weighted acquisition functions from components
$2,\cdots,k$ would have maxima less than the maximum of $\text{EI}_{1}(\bm{x})$.
Therefore, the $(n+1)$-th sequential sample must be selected as the
maximizer of $\text{EI}_{1}(\bm{x})$ in $\mathcal{X}_{1}$. Following
similar arguments on $n_{j}-\min_{j}n_{j}$
, we can see that $\min_{j=1,\cdots,k}n_{j}\rightarrow\infty$
as $n\rightarrow\infty$.

When all $n_{1},n_{2},\cdots,n_{k}>0$, by our algorithm, for the
sub-sequence $$\bm{x}_{1}^{(j)},\bm{x}_{2}^{(j)},\cdots,\bm{x}_{n_{j}}^{(j)},$$
$$\bm{x}_{J}^{(j)}\coloneqq\arg\max_{\bm{x}\in\mathcal{X}_{j}}\text{EI}_{j}(\bm{x})/J,$$
for $j=1,2,\cdots,k$. It is straightforward to see that $\bm{x}_{J}^{(j)}$
is also the $\arg\max_{\bm{x}\in\mathcal{X}_{j}}\text{EI}_{j}(\bm{x})$.
Due to the fact that additive components are independent, each sub-sequence
constitutes a sequential sampling within the component domain $\mathcal{X}_{j}$
with the same strategy. We use the assumption on $f\mid_{\mathcal{X}_{j}}$
and apply Theorem 5 of \citet{bull_convergence_2011} to each of these
$k$ independent components ($j=1,\cdots,k$) that 
\[
\sup_{\|f\mid_{\mathcal{X}_{j}}\|_{\mathcal{H}_{\theta^{U}}(\mathcal{X}_{j})}\leq R_{j}}\mathbb{E}_{f}\left|f\mid_{\mathcal{X}_{j}}(\bm{x}_{n_{j}}^{(j)*})-\min_{\mathcal{X}_{j}}f\mid_{\mathcal{X}_{j}}\right|=O\left(\left(\frac{n_{j}}{\log n_{j}}\right)^{-\nu/d}\left(\log n_{j}\right)^{\alpha_{j}}\right),
\]
 and we take the maximum (since $k<\infty$) on the LHS to yield the
stated result (\ref{eq:main result-statement 1}) with $\alpha=\max_{j}\alpha_{j}>0$
and $R=\min_{j}R_{j}>0$.
$\square$

Our result is slightly weaker than Theorem 5 in \citet{bull_convergence_2011}
in the sense that the same claim of error in terms of the supremum
norm holds for every sample size $n>0$. Our results can be applied
to quite general additive GP surrogate models with mutually independent
components. One step further, \citet{wynne_convergence_2020} proved
that even the $\nu$ are mis-specified, the convergence may still
hold with a worse rate. 

This discussion reveals the effect and importance
of re-weighting acquisition functions in the cGP algorithm. Assumptions
B and C focus on the smoothness for $f$ within each component. This
convergence would not hold if EI is not weighted, regardless the distribution
assumption of $\bm{x}_{i}'s$.

However, we shall point out that our Algorithm \ref{clustered GP} would update the partition induced by the cluster-classify step for each sampling step. Therefore, this proposition does not cover exactly cGP, but we provide a justification for weighting acquisition functions when we use an additive model in a Bayesian optimization setting. Besides, we do observe empirically that the partition usually becomes stable after sufficiently many samples are collected, so the simplification is not groundless. 

That being said, we still need to rigorously prove clustering and classification consistency (in sense that the decision boundaries coincide with non-smooth borders with an appropriately chosen $\xi$). This remains an unsolved theoretic question in the current paper, but usually a reasonable choice of $\xi$ or even $\xi=1$ can lead to empirically satisfying optimization results. 

\FloatBarrier

\vskip 0.2in 
\newpage
\bibliography{Nonsmooth}

\end{document}